\pdfoutput=1

\documentclass[11pt]{article}

\usepackage{acl}
\usepackage{amsmath}
\usepackage{bussproofs}
\usepackage{tcolorbox}
\usepackage{caption}
\usepackage{subcaption}
\usepackage{booktabs}
\usepackage{graphicx}
\usepackage{multirow}
\usepackage{extsizes}
\usepackage{mathptmx}
\usepackage{cuted}
\usepackage{flushend}
\usepackage{multicol}
\usepackage{anyfontsize}
\usepackage{rotating}
\usepackage{array,makecell,multirow}
\usepackage{wrapfig}
\usepackage{setspace}
\usepackage{microtype}
\usepackage{hyperref}
\usepackage{graphicx}
\usepackage{fontawesome}
\usepackage{amsmath}
\usepackage{amssymb}
\usepackage{comment}
\usepackage{algorithm}
\usepackage[noend]{algpseudocode}
\usepackage[algo2e,ruled,linesnumbered]{algorithm2e}
\usepackage{lipsum}
\usepackage[export]{adjustbox} 
\usepackage{varwidth}
\usepackage{enumitem}
\setlist{leftmargin=1mm}
\usepackage{pifont}
\usepackage{booktabs}
\usepackage{multirow}
\usepackage{subcaption}
\usepackage[capitalise]{cleveref}
\usepackage{graphicx}
\usepackage{colortbl}
\usepackage{tikz}
\usetikzlibrary{shapes.geometric, arrows}
\usetikzlibrary{decorations.markings}
\usepackage{soul}
\usepackage{wrapfig,graphicx,lipsum}
\usepackage[utf8]{inputenc}
\usepackage{extsizes}
\usepackage{mathptmx}
\usepackage{cuted}
\usepackage{flushend}
\usepackage{float}
\usepackage{multicol}
\usepackage{changepage,threeparttable}
\usepackage{anyfontsize}
\usepackage{setspace}

\usepackage{times}
\usepackage{latexsym}
\usepackage[T1]{fontenc}

\usepackage[utf8]{inputenc}

\usepackage{microtype}

\usepackage{inconsolata}

\usepackage{graphicx}
\usepackage{amsfonts}

\usepackage{booktabs}
\usepackage{tabularray}

\DefTblrTemplate{firsthead,middlehead,lasthead}{default}{
}
\DefTblrTemplate{firstfoot}{default}{
  \UseTblrTemplate{contfoot}{default}
  \UseTblrTemplate{caption}{default}
}
\DefTblrTemplate{middlefoot}{default}{
  \UseTblrTemplate{contfoot}{default}
  \UseTblrTemplate{capcont}{default}
}
\DefTblrTemplate{lastfoot}{default}{
  \UseTblrTemplate{note}{default}
  \UseTblrTemplate{remark}{default}
  \UseTblrTemplate{capcont}{default}
}

\newcolumntype{P}[1]{>{\centering\arraybackslash}p{#1}}

\usepackage{microtype}

\usepackage{inconsolata}

\usepackage[edges]{forest}
\definecolor{hidden-draw}{RGB}{20,68,106}
\definecolor{hidden-pink}{RGB}{255,245,247}
\definecolor{red}{RGB}{255,0,0}
\definecolor{hidden-draw}{RGB}{0,0,0}
\definecolor{hidden-pink}{RGB}{255,182,193}
\usepackage{tikz}

\usepackage{longtable}
\usepackage{tablefootnote}
\usepackage{xcolor}         

\definecolor{first}{RGB}{210,255,140}
\definecolor{second}{RGB}{136, 162, 190}
\definecolor{third}{RGB}{129, 222, 228}
\definecolor{fourth}{RGB}{132, 84, 246}
\definecolor{fifth}{RGB}{250, 223, 112}
\definecolor{sixth}{RGB}{203, 193, 172}
\definecolor{seventh}{RGB}{88, 112, 246}
\definecolor{eighth}{RGB}{245, 192, 106}
\definecolor{nine}{RGB}{171, 162, 111}
\definecolor{ten}{RGB}{217, 217, 217}

\definecolor{paired-light-blue}{RGB}{198, 219, 239}
\definecolor{paired-dark-blue}{RGB}{49, 130, 188}
\definecolor{paired-light-orange}{RGB}{251, 208, 162}
\definecolor{paired-dark-orange}{RGB}{230, 85, 12}
\definecolor{paired-light-green}{RGB}{199, 233, 193}
\definecolor{paired-dark-green}{RGB}{49, 163, 83}
\definecolor{paired-light-purple}{RGB}{218, 218, 235}
\definecolor{paired-dark-purple}{RGB}{117, 107, 176}
\definecolor{paired-light-gray}{RGB}{217, 217, 217}
\definecolor{paired-dark-gray}{RGB}{99, 99, 99}
\definecolor{paired-light-pink}{RGB}{222, 158, 214}
\definecolor{paired-dark-pink}{RGB}{123, 65, 115}
\definecolor{paired-light-red}{RGB}{231, 150, 156}
\definecolor{paired-dark-red}{RGB}{131, 60, 56}
\definecolor{paired-light-yellow}{RGB}{231, 204, 149}
\definecolor{paired-dark-yellow}{RGB}{141, 109, 49}
\definecolor{Teal}{RGB}{0, 50, 50}
\definecolor{White}{RGB}{250, 250, 250}

\definecolor{bg1}{HTML}{FF9966}
\definecolor{bg2}{HTML}{CCE5FF}
\definecolor{bg3}{HTML}{FFCC99}
\definecolor{bg4}{HTML}{FFC107}
\definecolor{bg5}{HTML}{FFCCCC}
\definecolor{bg6}{HTML}{D5E8D4}
\definecolor{bg7}{HTML}{eeeeee}
\definecolor{bg8}{HTML}{cdeb8b}
\definecolor{bg9}{HTML}{dae8fc}
\definecolor{bg10}{HTML}{a2e6eb}

\definecolor{bg31}{HTML}{FFCDD2} 

\definecolor{bg32}{HTML}{F8BBD0}

\definecolor{bg33}{HTML}{E1BEE7} 

\definecolor{bg34}{HTML}{D7CCC8} 

\definecolor{bg35}{HTML}{B2DFDB} 

\definecolor{bg36}{HTML}{A5D6A7} 

\definecolor{bg37}{HTML}{FFF9C4} 

\definecolor{bg38}{HTML}{FFECB3} 

\definecolor{bg111}{HTML}{CB6843}

\definecolor{bg112}{HTML}{D77C5C}

\definecolor{bg113}{HTML}{E28E6E}
\definecolor{bg114}{HTML}{E89F7D}
\definecolor{bg115}{HTML}{EDAE8A}
\definecolor{bg116}{HTML}{F0BA95}
\definecolor{bg117}{HTML}{F3C29F}
\definecolor{bg118}{HTML}{F6CCAA}
\definecolor{bg119}{HTML}{F8D5B3}
\definecolor{bg120}{HTML}{FADCBD}
\definecolor{bg121}{HTML}{FCE6C7}

\definecolor{bg39}{HTML}{FFE0B2} 

\definecolor{bg40}{HTML}{3CB371} 

\definecolor{bg43}{HTML}{ffe5d9}

\definecolor{bg15}{HTML}{7FFFD4}

\definecolor{bg17}{HTML}{F0FFFF}

\definecolor{bg18}{HTML}{F5FFFA}

\definecolor{bg19}{HTML}{F8F8FF}

\definecolor{bg20}{HTML}{FFFFFF}

\definecolor{bg21}{HTML}{E1F5FE}

\definecolor{bg22}{HTML}{B3E5FC}

\definecolor{bg23}{HTML}{81D4FA}

\definecolor{bg24}{HTML}{4FC3F7}

\definecolor{bg25}{HTML}{29B6F6}

\definecolor{bg26}{HTML}{03A9F4}

\definecolor{bg27}{HTML}{039BE5}

\definecolor{bg28}{HTML}{0288D1}

\definecolor{bg29}{HTML}{0277BD}

\definecolor{bg30}{HTML}{01579B}

\definecolor{bg16}{HTML}{FFCC99}

\definecolor{pg51}{HTML}{E8F5E9} 
\definecolor{pg52}{HTML}{C8E6C9} 
\definecolor{pg53}{HTML}{B9F6CA} 
\definecolor{pg54}{HTML}{A9DFBF} 
\definecolor{pg55}{HTML}{BCF5A6} 

\definecolor{pg56}{HTML}{BEF1CE} 
\definecolor{pg57}{HTML}{CEF6EC} 
\definecolor{pg58}{HTML}{B7F0B1} 
\definecolor{pg59}{HTML}{B1F2B5} 
\definecolor{pg60}{HTML}{9DF3C4} 

\definecolor{pg61}{HTML}{DEF7E0} 
\definecolor{pg62}{HTML}{E8F8DC} 

\definecolor{pg63}{HTML}{EBF7E7} 
\definecolor{pg64}{HTML}{F0FDF4} 

\definecolor{pg65}{HTML}{F1FEE7} 
\definecolor{pg66}{HTML}{F7FFF6} 
\definecolor{pg67}{HTML}{FCFFE7} 
\definecolor{pg68}{HTML}{F4FFD2} 
\definecolor{pg69}{HTML}{EEFFE2} 
\definecolor{pg70}{HTML}{E3FDF5} 

\definecolor{connect-color}{RGB}{0,0,0}
\definecolor{middle-color}{RGB}{255,255,255}
\definecolor{leaf-color}{RGB}{173,216,230}
\definecolor{line-color}{RGB}{25,25,112}

\usepackage{environ}

\newlength{\myl}
\expandafter\let\expandafter\origequation\csname equation*\endcsname
\expandafter\let\expandafter\endorigequation\csname endequation*\endcsname
\long\def\[#1\]{\begin{equation*}#1\end{equation*}}
\RenewEnviron{equation*}{
  \settowidth{\myl}{$\displaystyle\BODY$} 
  \origequation
    \ifdim\myl>\linewidth
      \resizebox{\linewidth}{!}{$\displaystyle\BODY$}
    \else
      \BODY 
    \fi
  \endorigequation
}

\title{Counter Turing Test ($CT^2$): Investigating AI-Generated Text Detection for Hindi - Ranking LLMs based on Hindi AI Detectability Index ($ADI_{hi}$)}

\author{\textbf{Ishan Kavathekar}$^{1}$ \quad \textbf{Anku Rani} \dag $^{2}$
\quad \textbf{Ashmit Chamoli}$^{1}$ \\ \textbf{Ponnurangam Kumaraguru}$^{1}$ \quad \textbf{Amit Sheth}$^{3}$ \quad \textbf{Amitava Das}$^{3}$
\\
$^{1}$ International Institute of Information Technology, Hyderabad \quad  \\
$^{2}$ Massachusetts Institute of Technology \quad   \\
$^{3}$ AI Institute, University of South Carolina \quad \\
\tt ishan.kavathekar@research.iiit.ac.in
}

\begin{document}
\maketitle

\renewcommand{\thefootnote}{\fnsymbol{footnote}}
\footnotetext[2]{Work was done when the author was at the AI Institute, University of South Carolina.}
\renewcommand*{\thefootnote}{\arabic{footnote}}
\setcounter{footnote}{0}

\begin{abstract}
The widespread adoption of Large Language Models (LLMs) and awareness around multilingual LLMs have raised concerns regarding the potential risks and repercussions linked to the misapplication of AI-generated text, necessitating increased vigilance. While these models are primarily trained for English, their extensive training on vast datasets covering almost the entire web, equips them with capabilities to perform well in numerous other languages. AI-Generated Text Detection (AGTD) has emerged as a topic that has already received immediate attention in research, with some initial methods having been proposed, soon followed by the emergence of techniques to bypass detection. In this paper, we report our investigation on AGTD for an indic language Hindi. Our major contributions are in four folds: i) examined 26 LLMs to evaluate their proficiency in generating Hindi text, ii) introducing the AI-generated news article in Hindi (AG\textsubscript{hi}) dataset, iii) evaluated the effectiveness of five recently proposed AGTD techniques: ConDA, J-Guard, RADAR, RAIDAR and Intrinsic Dimension Estimation for detecting AI-generated Hindi text, iv) proposed Hindi AI Detectability Index ($ADI_{hi}$) which shows a spectrum to understand the evolving landscape of eloquence of AI-generated text in Hindi. The code and dataset is available at \url{https://github.com/ishank31/Counter_Turing_Test}

\end{abstract}

\section{AGTD - The Necessity}

With the rise in the number of LLMs capable of generating text across languages, the risks associated with it have increased. AI-generated text could result in misinformation and fake news \cite{kreps2022all}, online manipulation can be used to create fake reviews and social media posts \cite{chernyaeva2022ai}, Phishing and scams where malicious actors may use AI-generated text to generate phishing emails \cite{basit2021comprehensive}, etc. 

In summary, as generative models grow, we need comparable detection techniques. AI text detection is necessary to safeguard individuals, organizations, and society from the potential negative consequences of malicious or misleading content generated by AI systems. It plays a crucial role in maintaining the integrity of online communication and upholding ethical standards in the use of AI technologies. We are the first to conduct experiments for AI-generated news article generation and detection techniques for the Hindi language. Hindi is the fourth most spoken first language in the world after Mandarin, Spanish, and English \cite{wiki:List_of_languages_by_number_of_native_speakers}. Taking inspiration from recent works of AI-generated text detection for English \cite{chakraborty2023counter} where they discussed $6$ detection techniques namely watermarking, perplexity estimation, burstiness estimation, negative log curvature, stylometric variation and classification-based approach. We extend it to regional languages like Hindi and cover five new detection techniques to assess AI-generated text detection for Hindi.

\begin{tcolorbox}
[colback=blue!5!white,colframe=black!65!black,title=\textbf{\footnotesize \textsc{Our Contributions}:} {\footnotesize A Counter Turing Test (\textbf{CT}$^2$) and AI Detectability Index for Hindi ($ADI_{hi}$)}]

\begin{itemize}[leftmargin=1mm]
\setlength\itemsep{0em}
\begin{spacing}{0.85}
    \item[\ding{224}]{\footnotesize 
    {\fontfamily{phv}\fontsize{8}{8}\selectfont
    Introducing the \emph{\ul{Counter Turing Test (CT$^2$)}} for Hindi, a benchmark that incorporates methods designed to provide a thorough assessment of the resilience of existing AGTD techniques in Hindi.
    }
    }

    \item[\ding{224}] {\footnotesize 
    {\fontfamily{phv}\fontsize{8}{8}\selectfont
    Conducting a thorough examination of \textbf{26} LLMs to generate an AI-generated news article in Hindi. (AG\textsubscript{hi}) dataset}
    }
    
    \item[\ding{224}] {\fontfamily{phv}\fontsize{8}{8}\selectfont
    Presenting the \textit{\ul{AI Detectability Index for Hindi ($ADI_{hi}$)}} as a metric for Language Models to assess whether their outputs can be identified as generated by artificial intelligence or not.} 
    
    \item[\ding{224}] {\footnotesize 
    {\fontfamily{phv}\fontsize{8}{8}\selectfont 
    Curated datasets and models is made available for open-source research and commercial use.
    
    }
    }
\vspace{-5mm}
    
\end{spacing}
\end{itemize}
\end{tcolorbox}

\begin{figure*}[!htb]
  \centering
  \resizebox{\textwidth}{!}{%
    \begin{forest}
      forked edges,
      for tree={
        grow=east,
        reversed=true,
        anchor=base west,
        parent anchor=east,
        child anchor=west,
        base=center,
        font=\large,
        rectangle,
        draw=hidden-draw,
        rounded corners,
        align=center,
        text centered,
        minimum width=5em,
        edge+={darkgray, line width=1pt},
        s sep=3pt,
        inner xsep=2pt,
        inner ysep=3pt,
        line width=0.8pt,
        ver/.style={rotate=90, child anchor=north, parent anchor=south, anchor=center},
      },
      where level=1{text width=15em,font=\normalsize,}{},
      where level=2{text width=14em,font=\normalsize,}{},
      where level=3{minimum width=10em,font=\normalsize,}{},
      where level=4{text width=26em,font=\normalsize,}{},
      where level=5{text width=20em,font=\normalsize,}{},
      [
        \textbf{AI-Generated Text}\\ \textbf{Detection Techniques}, for tree={fill=paired-light-red!70}
        [
          \textbf{Watermakring} , for tree={fill=paired-light-yellow!45}
          [
           \cite{pmlr-v202-kirchenbauer23a}; \cite{kuditipudi2023robust}, text width=54.3em, for tree={fill=bg1}
          ]
        ]
        [
          \textbf{Salient features of} \\ \textbf{AI-Generated Text}, for tree={fill=bg35}
            [
            \textbf{RAIDAR} \cite{mao2024raidar};
            \textbf{Intrinsic Dimension Estimation} \cite{tulchinskii2023intrinsic}, text width=54.3em, for tree={fill=bg2}
            ]
        ]
        [
          \textbf{Classification Methods}, for tree={fill=bg22}
          [
              \textbf{J-Guard} \cite{kumarage2023jguard};
              \textbf{ConDA} \cite{bhattacharjee2023conda}; 
              \textbf{RADAR} \cite{hu2023radar}, text width=54.3em, for tree={fill=pg67}
          ]
        ]
        [
            \textbf{Statistical Methods}, for tree={fill=bg39}
            [
            \textbf{Perplexity} \cite{tian2023gptzero};
            \textbf{Burstiness} \cite{tian2023gptzero};
            \textbf{Negative Log Curvature} \cite{10.5555/3618408.3619446}, text width=54.3em, for tree={fill=bg19}
            ]
        ]
      ]
    \end{forest}
  }
  \vspace{-5mm}
  \caption{Taxonomy of AI-Generated Text Detection techniques, showcasing various watermarking, feature-related, statistical, and classification-based techniques for detecting AI-generated text.} 
  \label{fig:types_of_techniques}
\end{figure*}
\section{Multilingual LLMs}
In this paper, we investigate the effectiveness of AGTD techniques on the  Hindi language.
This section discusses our selected LLMs and elaborates on our data generation methods.

\subsection{LLMs: Rationale and Coverage}
We chose a wide gamut of 26 LLMs that have exhibited exceptional results on a wide range of NLP tasks.
They are: (i) GPT-4 \cite{openai2024gpt4technicalreport}; (ii) GPT-3.5 \cite{chen2023robust}; (iii) GPT-2 (Base, Medium, Large, XL) \cite{radford2019language}; (iv) BARD (now Gemini) \cite{Bard2023}; (v) Bloom (560M, 3B, 7B) \cite{BigScience_BLOOM_2022} (vi) Bloomz (560M, 1B, 3B, 7B) \cite{muennighoff2022crosslingual}; (vii) mGPT (1.3B) \cite{shliazhko2023mgpt}; (viii) Mistral Instruct 7B \cite{jiang2023mistral}; (ix) Gemma-1.1 (2B, 7B) \cite{gemmateam2024gemmaopenmodelsbased}; (x) mT0 (Small, Base, Large, XL) \cite{muennighoff2022crosslingual}; (xi) mT5 (Small, Base, Large, XL) \cite{xue2021mt5} .

As the field is in a constant state of evolution, we acknowledge that this process will never reach its finality but instead will persist in its expansion. Therefore, we intend to maintain the Hindi leaderboard benchmark as an open platform for researchers, facilitating ongoing updates and contributions.

\subsection{Criteria of selection for Hindi LLM}
\label{sec:criteria}

We experimented with a total of 26 LLMs including variation in their parameter size. We reject a model if it generates no output, produces gibberish, engages in code-switching or generates output solely in English. Table \ref{tab:rejected_models_criteria} summarizes the rejection criteria for these dismissed models. Additional details about the selection criteria are provided in Appendix \ref{sec:acceptance_criteria}.

\begin{table}[!h]

\centering
\resizebox{1\columnwidth}{!}{
\begin{tabular}{ccccc}
\hline
\textbf{Model} & \textbf{No output} & \textbf{Gibberish output} & \textbf{English output} & \textbf{Code-switching} \\ \hline
Bloom-560M & \checkmark & -- & \checkmark & \checkmark \\
Bloom-3B & \checkmark & -- & -- & \checkmark \\
Bloom-7B & \checkmark & -- & \checkmark & \checkmark \\
Bloomz-560M & \checkmark & -- & -- & -- \\
Bloomz-1B & \checkmark & \checkmark & -- & -- \\
Bloomz-3B & \checkmark & -- & -- & \checkmark \\
Bloomz-7B & \checkmark & -- & -- & -- \\
GPT-2 Base & -- & \checkmark & \checkmark & \checkmark \\
GPT-2 Medium & -- & \checkmark & \checkmark & \checkmark \\
GPT-2 Large & -- & \checkmark & \checkmark & \checkmark \\
GPT-2 XL & -- & \checkmark & \checkmark & \checkmark \\
mGPT-1.3B & -- & \checkmark & -- & -- \\
Mistral-7B & -- & \checkmark & -- & -- \\
mT0 models & \checkmark & \checkmark & -- & -- \\
mT5 models & \checkmark & \checkmark & -- & -- \\ \hline
\end{tabular}
}
\caption{Criteria used for rejecting the dismissed models. Bloom and Bloomz models fail to generate outputs for most Hindi prompts. GPT-2 models produce gibberish or English outputs, with occasional instances of code-switching. Encoder-decoder models, mT5 and mT0 either produce no output or generate gibberish.}
\label{tab:rejected_models_criteria}
\end{table}

Through our experimentation and observation of the generated outputs, we rejected 21 models. Some of the outputs from these models are present in Appendix \ref{sec: Hindi_examples}. We have retained the responses for 100 data points from BBC Hindi for the rejected models, thereby providing a valuable resource for future research endeavors. This dataset exemplifies why certain models were deemed unfit for inclusion due to their inability to generate coherent and meaningful text. In summary, out of all the 26 LLMs tested for AI-generated news articles in Hindi, we have considered 5 models (BARD, GPT-3.5 Turbo, GPT-4, Gemma-1.1-2B-it, Gemma-1.1-7B-it).

\subsection{Hindi AGTD ($AG_{hi}$) dataset}
In this section, we detail the methodology employed for generating our $AG_{hi}$ dataset.

    \noindent \textbf{Human Written Articles}: The human-written articles dataset is derived from \href{https://www.bbc.com/hindi}{BBC} and \href{https://ndtv.in/}{NDTV} news platforms, encompassing various categories, including India, international affairs, sports, Bollywood, lifestyle, health, and more.

\begin{figure*}[ht!]
    \centering
    \begin{subfigure}[b]{0.32\textwidth}
        \centering
        \includegraphics[width=\textwidth]{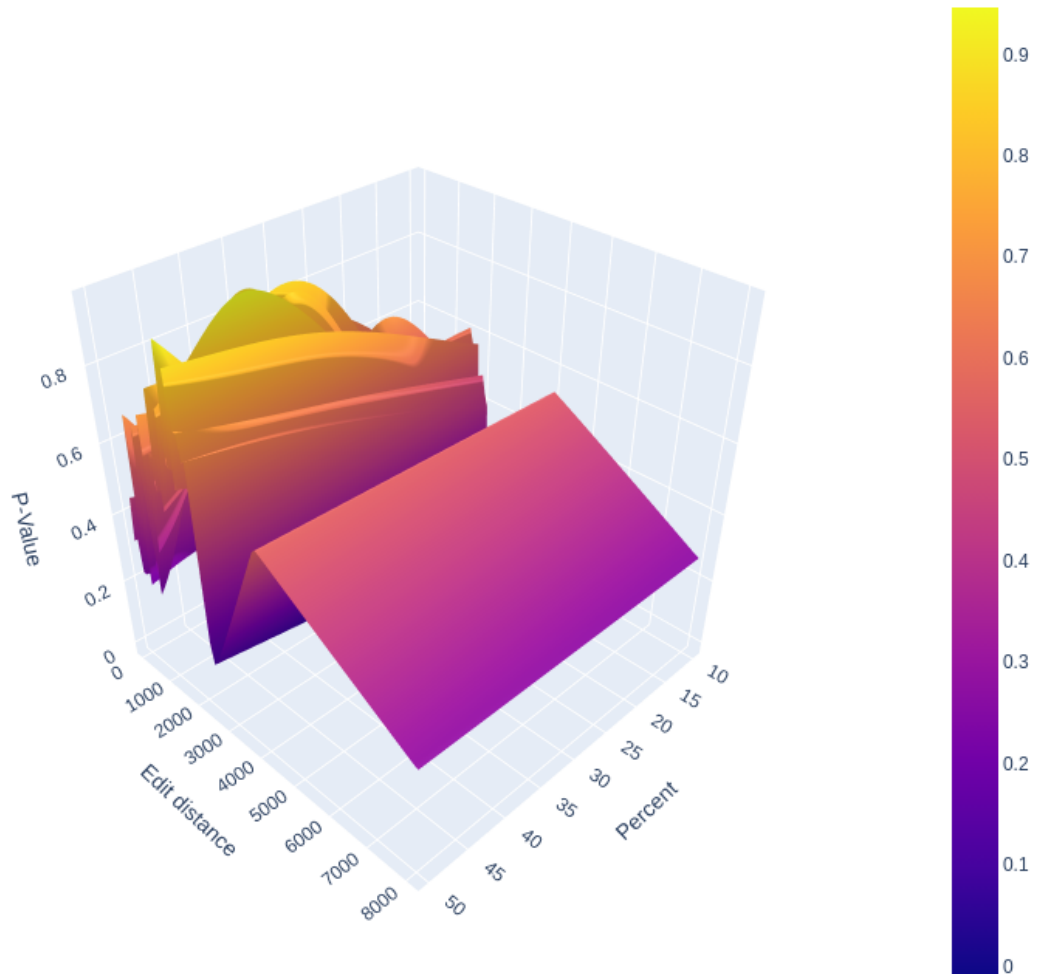}
        \caption{Edit Distance}
        \label{fig:watermark_edit}
    \end{subfigure}
    \hfill
    \begin{subfigure}[b]{0.32\textwidth}
        \centering
        \includegraphics[width=\textwidth]{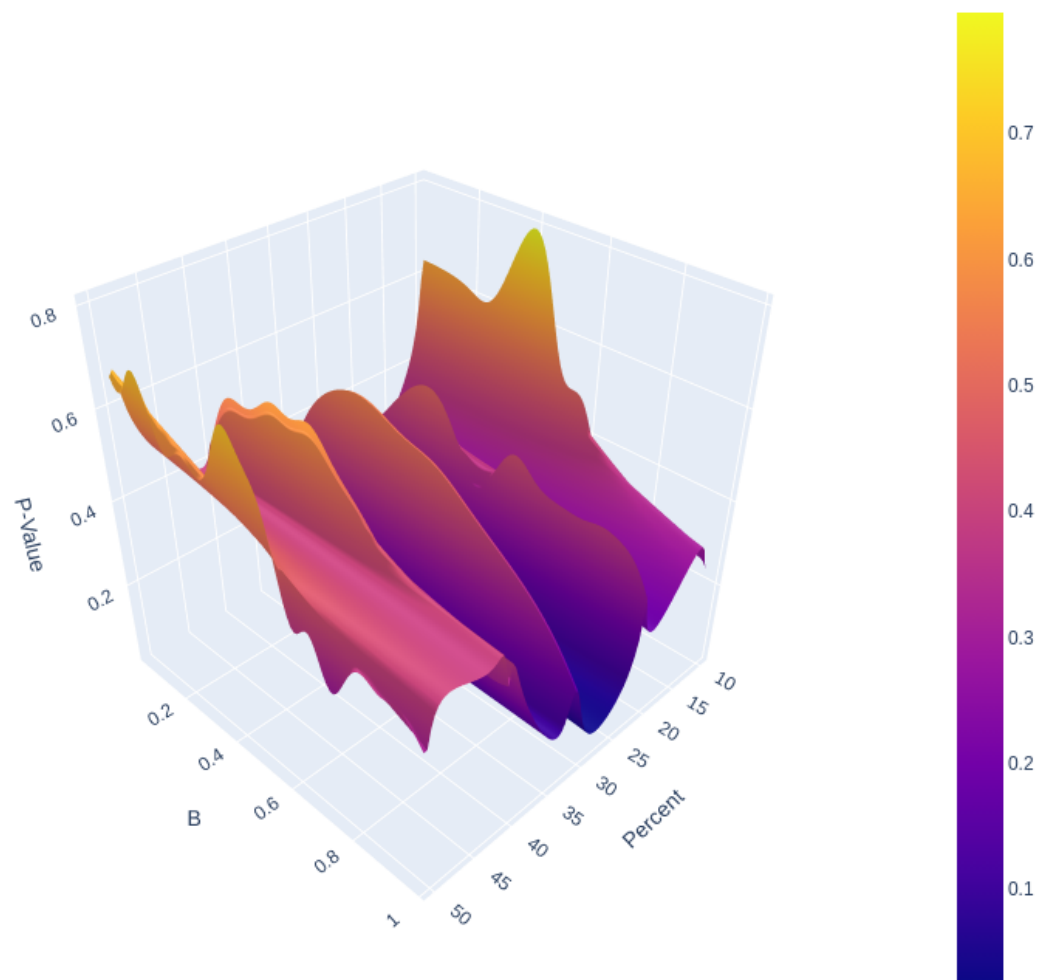}
        \caption{BLEU Score}
        \label{fig:watermark_bleu}
    \end{subfigure}
    \hfill
    \begin{subfigure}[b]{0.32\textwidth}
        \centering
        \includegraphics[width=\textwidth]{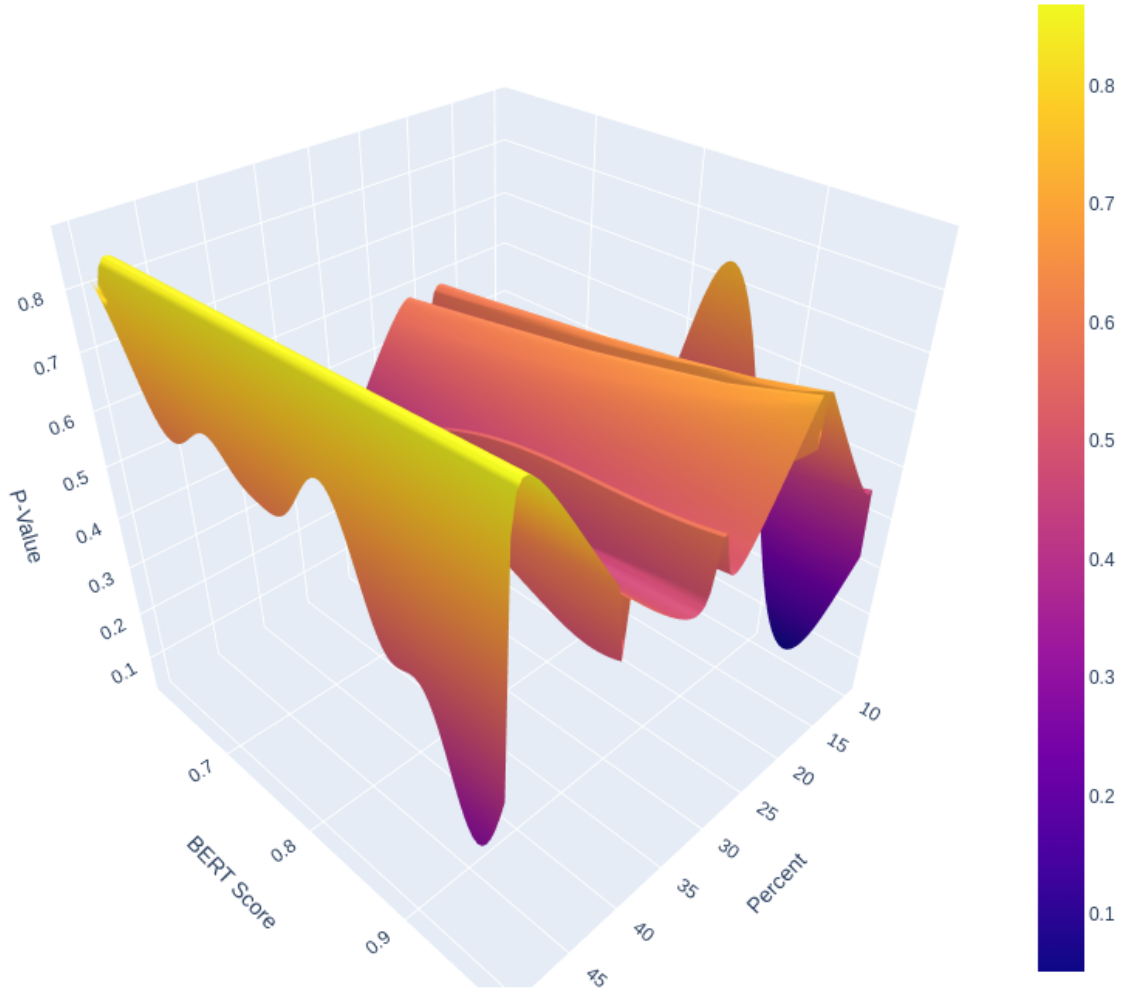}
        \caption{BERTScore}
        \label{fig:watermark_bert}
    \end{subfigure}
    \caption{Distortion of text vs detectability of watermark. We observe that the p-value increases even when the percentage of watermarked tokens increases. (a) We see variation in p-values for edit distances below 2000. However, for distances above 2000, p-values become constant and are not influenced by the percentage of tokens watermarked (b) Higher similarity to the original text, indicated by a high BLEU score, correlates with lower p-values (c) Semantic similarity does not influence p-values. However, p-values increase after 30\% watermarked tokens, reducing watermark detection reliability. }
    \label{fig:Watermarking}
\end{figure*}

    \noindent \textbf{AI-Generated Articles}: To obtain AI-generated responses, we employed state-of-the-art 5 LLMs. The headlines collected from the human-written articles were presented as prompts to these LLMs, which generated text responses. The five selected models for the curation of AI-generated articles, resulted in a total of 29,627 AI-generated news articles in Hindi from two Hindi news sources BBC and NDTV as shown in Table \ref{tab:dataset}. The details of the prompts and the hyperparameters used while producing the dataset are detailed in Appendix \ref{sec:hyperparameters}.

\begin{table}[!ht]
\centering
\resizebox{0.45\textwidth}{!}{
\begin{tabular}{lcc}
\toprule
\hline
\textbf{Data Sources} & \textbf{\begin{tabular}[c]{@{}c@{}} Human Written\\ News Articles\end{tabular}} & \textbf{\begin{tabular}[c]{@{}c@{}}AI Generated \\ News Articles\end{tabular}} \\ \hline \hline
\textbf{BBC}         & 1762                                                                  & 7390                                                                  \\ 
\textbf{NDTV}        & 5281                                                                  & 22237                                                                 \\ \hline \midrule
\textbf{Total}       & 7043                                                                  & 29627                                                                 \\ \hline \bottomrule
\end{tabular}
}
\caption {Statistics of human-written and AI-generated news articles in Hindi. The dataset comprises a total of 36,670 news articles.}
\label{tab:dataset}
\end{table}

\section{Related Works - SoTA methods}

The current AGTD methods can be broadly grouped into four categories: \textit{ (i) Watermarking, (ii) Methods based on features of AI-generated text  (iii) Classification based methods and (iv) Statistical methods}, as illustrated in Figure \ref{fig:types_of_techniques}.

Watermarking has long been an established method in computer vision to identify the source and ownership of content. \citet{pmlr-v202-kirchenbauer23a} were the first to present watermarking models for LLMs, though their initial proposal faced criticism. Studies by \citet{sadasivan2024aigenerated} and \citet{NEURIPS2023_575c4500} demonstrated that paraphrasing can effectively eliminate the watermark, rendering this method ineffective. In response, \citet{kirchenbauer2024reliability} introduced a more resilient method, which was more robust to paraphrasing. However, this method can be circumvented by a combination of replacing high entropy words and paraphrasing \cite{chakraborty2023counter}. In this paper, we are the first to discuss the balance between the distortion and detectability of the watermark in Section \ref{sec:watermarking}.   

Recent studies suggest that salient features of AI-generated text and operational characteristics of the LLMs can be utilized to effectively detect AI-generated content. In Section \ref{sec:salient_prop}, we explore two methods that leverage these features: (i) Intrinsic Dimension Estimation \cite{tulchinskii2023intrinsic} and (ii) RAIDAR \cite{mao2024raidar}

Classification methods address the problem of AI-generated text detection by framing it as a binary classification task. Present studies utilize the texts generated from LLMs to train the classifiers \cite{li2023origin, mao2024raidar}. We discuss three such methods: (i) RADAR \cite{hu2023radar}; (ii) J-Guard \cite{kumarage2023jguard} and (iii) ConDA \cite{bhattacharjee2023conda} in Section \ref{sec:Classification_methods}.

Statistical methods leverage the discrepancies in statistical characteristics of texts. They assess the deviations in measures such as perplexity, burstiness \cite{tian2023gptzero}, entropy and n-gram frequency to differentiate between human and AI-generated texts. DetectGPT \cite{10.5555/3618408.3619446} makes use of the observation that the AI-generated text lies in the negative curvature region of an LLM's log probability space to differentiate between human and AI-generated text. Recent studies, however, have criticized these methods for their unreliability \cite{chakraborty2023counter}. Therefore, we do not investigate them further.

\vspace{-0.3mm}
There has been significant exploration of ways to circumvent the detection techniques. \citet{NEURIPS2023_575c4500} and \citet{sadasivan2024aigenerated} have shown that the watermarking technique is vulnerable to paraphrasing attacks. \citet{lu2024large} proposed a Substitution-based In-Context example Optimization (SICO) method which can evade AI detectors without relying on an external paraphraser.
\section{Testing the tradeoffs of distortion vs. detectability in watermarking}
\label{sec:watermarking}

\begin{table*}[h]
\centering
\resizebox{1\textwidth}{!}{
\Huge
\begin{tabular}{clll}
\hline
\textbf{\begin{tabular}[c]{@{}c@{}}AGTD\\ Technique\end{tabular}} & \multicolumn{1}{c}{\textbf{Performance}} & \multicolumn{1}{c}{\textbf{Pros}} & \multicolumn{1}{c}{\textbf{Cons}} \\ \hline
\begin{tabular}[c]{@{}c@{}}\bf{Intrinsic Dimension}\\ \bf{Estimation}\end{tabular} & \begin{tabular}[c]{@{}l@{}} - Different LLMs exhibit distinct MLE and PHD values.\\ - GPT-4, GPT-3 and BARD showcase intrinsic dimension\\ 
\hspace{0.8em}similar to human text.\\ - Difference in intrinsic dimension of Gemma outputs and\\ \hspace{0.8em}human text make the responses more discernible.\end{tabular} & \begin{tabular}[c]{@{}l@{}} - Invariant property of text\\ - Language agnostic technique\\ - No training required\end{tabular} & \begin{tabular}[c]{@{}l@{}} - Models producing human-like responses share similar \\ \hspace{0.8em}intrinsic dimension as human text, making their outputs\\ \hspace{0.8em}harder to detect.\end{tabular} \\ \hline
\bf{RAIDAR} & \begin{tabular}[c]{@{}l@{}} - Responses of Gemma models are highly detectable\\ - The method fails to detect GPT-4, GPT-3, and BARD \\ \hspace{0.8em}responses, with a significant performance drop of 24-32\%.\end{tabular} & \begin{tabular}[c]{@{}l@{}} - Language agnostic technique\\ - No training required\end{tabular} & \begin{tabular}[c]{@{}l@{}} - Computational resources vary depending on the LLM\\  \hspace{0.8em}used for text rewriting.\\ - Performance is sensitive to both the model chosen for\\ \hspace{0.8em}rewriting and the prompt provided.\end{tabular} \\ \hline
\textbf{RADAR} & \begin{tabular}[c]{@{}l@{}} - Detection accuracy of GPT responses drops below 50\%,\\  \hspace{0.8em}performing worse than a random classifier.\\ - Higher precision than recall suggests the model classifies\\ \hspace{0.8em}human-written text well but struggles to detect \\ \hspace{0.8em}AI-generated text.\\ - Consistently low accuracy and F1-scores indicate RADAR's \\ \hspace{0.8em}difficulty in accurately identifying AI-generated text.\end{tabular} & \begin{tabular}[c]{@{}l@{}}- Identifies human text with great\\ \hspace{0.8em}precision\\ - Trained on paraphrased data along\\ \hspace{0.8em}with training data\end{tabular} & - Not trainable \\ \hline
\textbf{J-Guard} & \begin{tabular}[c]{@{}l@{}}- Outperforms other methods, likely due to its focus\\ \hspace{0.8em}on journalistic features and news article data.\\ - Cross-model analysis shows a 10-27\% performance dip \\ \hspace{0.8em}when trained on the BBC dataset and tested on NDTV.\\ - The model is less efficient when trained on data from \\ \hspace{0.8em}Gemma models compared to GPT or BARD data.\end{tabular} & \begin{tabular}[c]{@{}l@{}} - Computationally easy to train\\ - Performs better than other models \\ \hspace{0.8em}considered in the study\end{tabular} & \begin{tabular}[c]{@{}l@{}}- Specifically designed to detect AI-generated news articles.\\ - Performance is sensitive to the training data.\end{tabular} \\ \hline
\textbf{ConDA} & \begin{tabular}[c]{@{}l@{}}- ConDA's performance metrics, all below 50\%, highlight\\ \hspace{0.8em}its difficulty in handling the task effectively.\\ - Low precision and recall indicate frequent misclassification\\ \hspace{0.8em}of AI-generated and human-written text.\end{tabular} & \begin{tabular}[c]{@{}l@{}}- Utilizes unsupervised domain \\ \hspace{0.8em}adaptation and self-supervised \\ \hspace{0.8em}contrastive learning to leverage labeled \\ \hspace{0.8em}data from the source domain and unlabeled\\ \hspace{0.8em}data from the target domain effectively.\end{tabular} & - Significantly low performance on Hindi text \\ \hline
\end{tabular}
}
\caption{A brief description of performance, pros and cons of each AGTD technique.}
\end{table*}

To embed a watermark in text, targeted alterations of specific text units are required. While it is intuitive that increasing the number of alterations enhances the strength of the watermark, excessive changes can significantly distort the original text. Therefore, an effective watermarking method requires a delicate balance between distortion and detectability. To our knowledge, no prior work has addressed this issue comprehensively. Although \citet{kuditipudi2023robust} discussed distortion in the watermarked text at a high level, they refrained from quantifying this phenomenon. In this paper, we empirically study the balance between distortion and detectability based on the watermarking methods proposed by \citet{pmlr-v202-kirchenbauer23a}. We propose using Minimum Edit Distance to calculate lexical distortion, BLEU score \cite{10.3115/1073083.1073135} for syntactic distortion, and BERTScore \cite{zhang2019bertscore} for semantic distortion. For detectability, we utilize z-score and p-value as proposed by \citet{pmlr-v202-kirchenbauer23a}.

In our evaluation, we employ the Gemma-2B model for paraphrasing responses by Gemma-7B. We observe that after paraphrasing, the watermark present in the text becomes undetectable, evidenced by p-values greater than 0.01 in Figure \ref{fig:Watermarking}. Intuitively, one would expect that a higher number of watermarked tokens would result in paraphrasing having a lower impact on the watermark. However, our observations indicate that samples with the highest percentage of watermarked tokens (50\%) still exhibit high p-values, indicating almost complete elimination of the watermark. The semantic distortion of the text, as quantified by BERTScore, does not significantly affect watermark detectability. Additionally, we noticed that as the BLEU score increases, indicating that the paraphrased text is syntactically similar to the original, the p-value decreases, suggesting more reliable watermark detection compared to samples with lower BLEU scores.

\section{Methods based on salient properties of AI-generated texts}\label{sec:salient_prop}
This section discusses the methods which leverage the distinct features of the AI-generated text for detection.

\subsection{Intrinsic Dimension Estimation}

Intrinsic Dimension estimation \cite{tulchinskii2023intrinsic} introduces an invariant property for human-written text—namely, the intrinsic dimension of the underlying embedding manifold. The authors focus on the \textit{Persistence Homology Dimension (PHD)} which belongs to the class of fractal dimension approaches. They chose PHD due to its ability to capture both local and global dataset properties efficiently and robustly against noise. The hypothesis is that the human-texts exhibit higher PHD than that of AI-generated texts enabling a clear differentiation between the two. \citet{tulchinskii2023intrinsic} show that the PHD of most European languages is approximated to be \textbf{9$\pm$1}. However, our experiments reveal that  Hindi texts have a lower PHD, ranging from \textbf{6 to 7}. Moreover, the maximum likelihood estimation (MLE) values lie in the range of \textbf{9 to 10}.

\subsection{RAIDAR}

Generative AI Detection via Rewriting (RAIDAR) \cite{mao2024raidar} method suggests that text generated by auto-regressive generative models typically maintains a consistent structure, often leading other such  models to perceive this AI-generated text as high quality. RAIDAR observes that generative models alter AI-generated text less frequently compared to human-written text during rewriting. RAIDAR focuses on the symbolic word outputs of large language models (LLMs) over other features, leveraging the minimal character edit distance between original and rewritten text. In our experiments, we utilized six prompts and applied Gemma-2B to rewrite samples from the dataset. Additional details on the prompts are available in Appendix \ref{sec:RAIDAR_appendix}.
\section{Classification Based Methods} \label{sec:Classification_methods}
This section discusses classification-based methods for detecting AI-generated text. These methods utilize a range of techniques such as adversarial learning, self-supervised learning, and stylometry while training on both human-written and AI-generated text.


\subsection{RADAR}
Robust AI-text detector via adversarial learning (RADAR) \cite{hu2023radar} is a novel framework that employs adversarial training to enhance AGTD. RADAR presents a paraphraser and a detector as two opposing agents inspired by adversarial machine learning techniques. The paraphraser aims to generate realistic content that can bypass AI-text detection, whereas the detector is trained to enhance the detectability of the AI-generated text. The paraphraser rewrites the text generated by the LLMs to evade detection as AI-generated. Conversely, the detector learns to distinguish between human and AI-generated text using both the training data and the paraphraser's output. 


\begin{figure}[htb]
    \centering
    
    \begin{subfigure}[b]{\columnwidth} 
        \centering
        \includegraphics[width=\columnwidth]{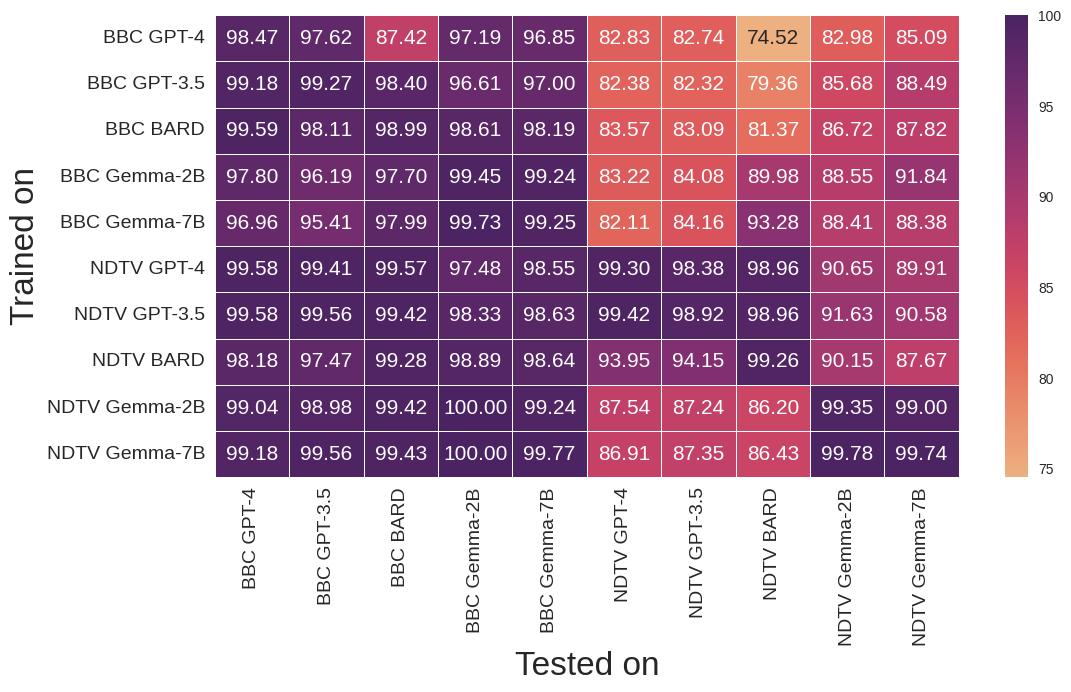}
        \caption{J-Guard cross-model F1 scores}
        \label{fig:jguard_cross_final}
    \end{subfigure}
    
    \vspace{0.5cm} 
    
    \begin{subfigure}[b]{0.3\columnwidth} 
        \centering
        \includegraphics[width=\columnwidth]{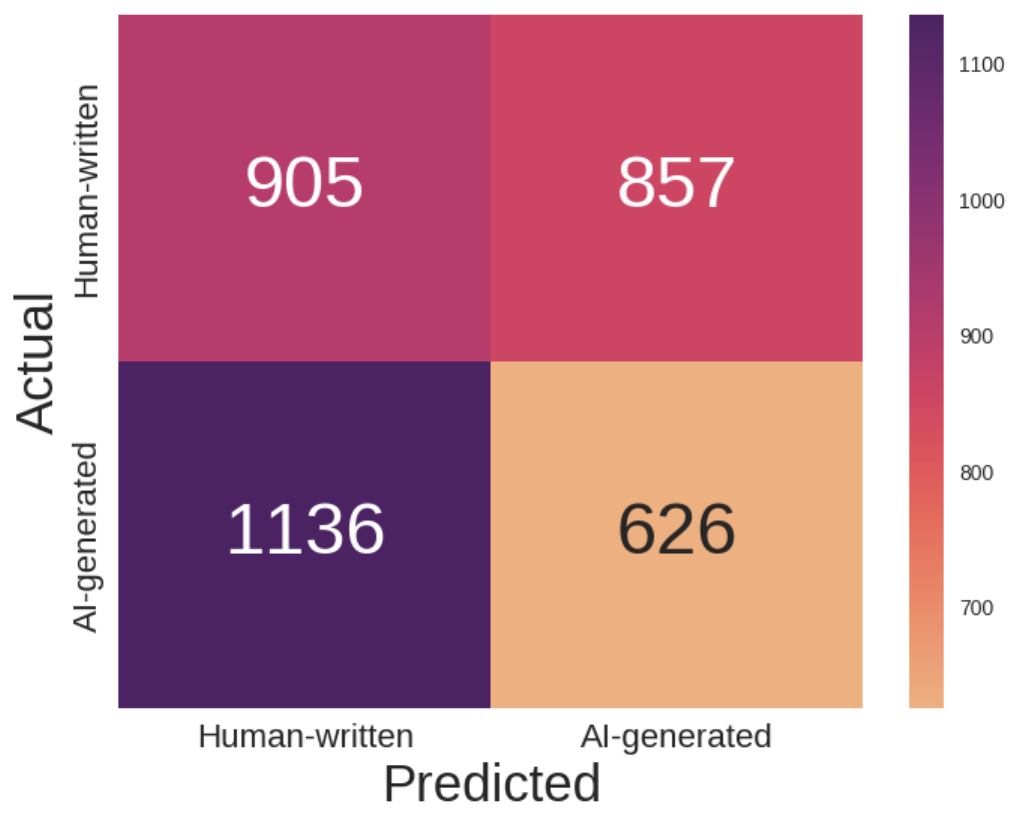}
        \caption{ConDA}
        \label{fig:conda_cf_final}
    \end{subfigure}%
    \hfill
    \begin{subfigure}[b]{0.3\columnwidth} 
        \centering
        \includegraphics[width=\columnwidth]{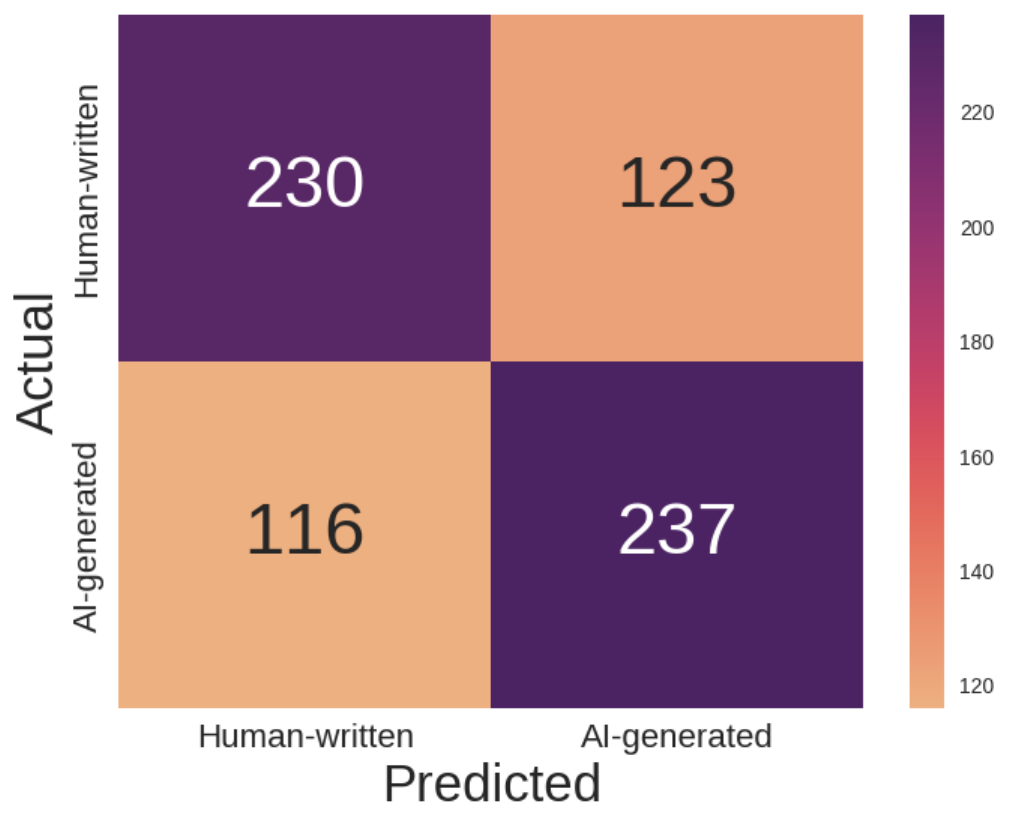}
        \caption{RAIDAR}
        \label{fig:raidar_df_final}
    \end{subfigure}%
    \hfill
    \begin{subfigure}[b]{0.3\columnwidth} 
        \centering
        \includegraphics[width=\columnwidth]{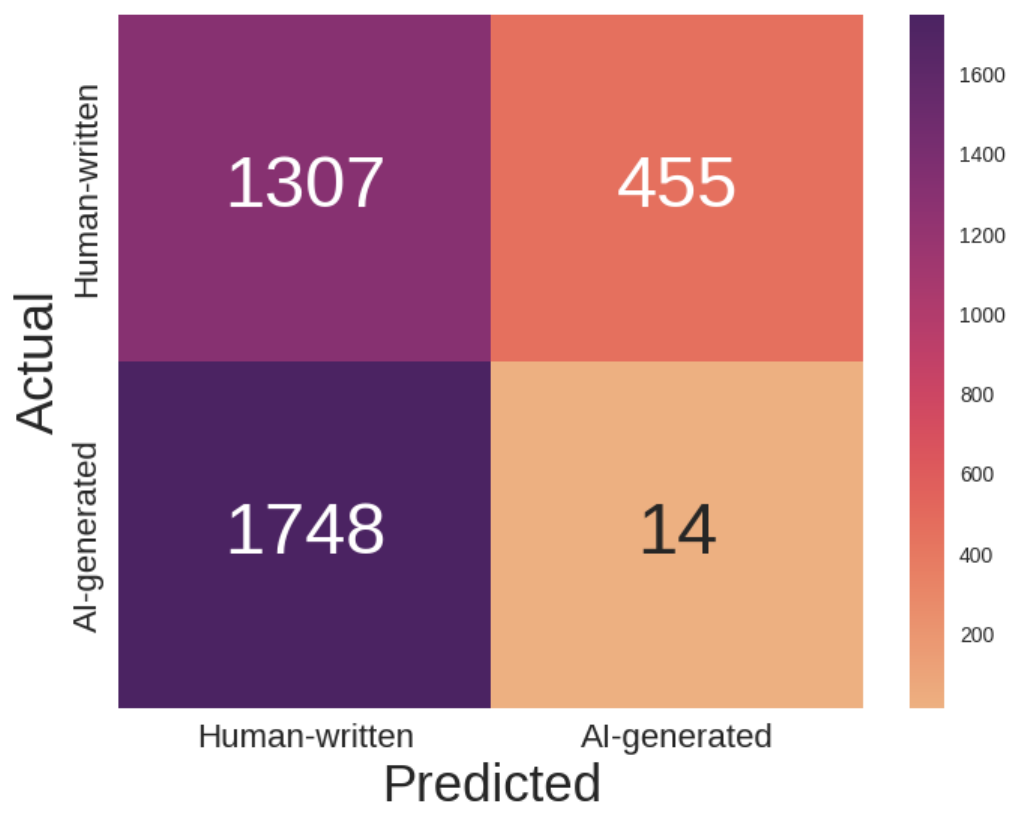}
        \caption{RADAR}
        \label{fig:radar_cf_final}
    \end{subfigure}
    
    \caption{(a) Models trained on BBC dataset and tested on NDTV dataset show a significant drop of 10-15\% in F1 score. ConDA and RAIDAR (b, c) misclassifies AI-generated text as human-written and vice versa, leading to high number of false positives and false negatives. RADAR (d) frequently classifies a given text as human-written, leading to misclassification of majority AI-generated text.}
    \label{fig:jguard_heatmap}
\end{figure}


\begin{figure*}[h]
    \centering
    \begin{subfigure}[b]{0.19\textwidth}
        \centering
        \includegraphics[width=\textwidth]{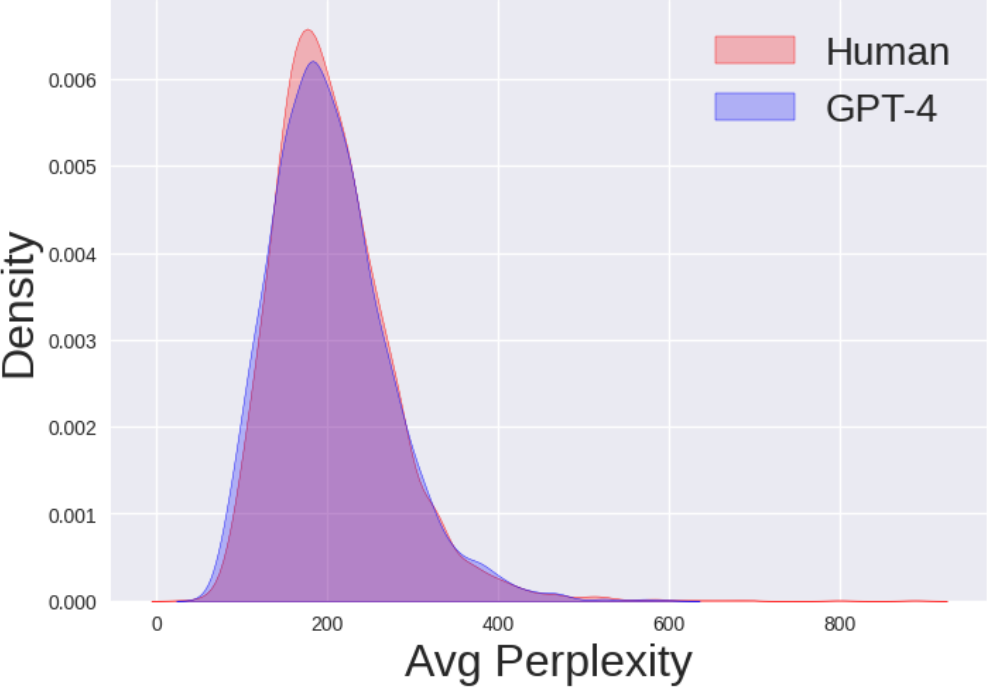}
        \caption{GPT-4}
        \label{fig:perp_gpt4}
    \end{subfigure}
    \hfill
    \begin{subfigure}[b]{0.19\textwidth}
        \centering
        \includegraphics[width=\textwidth]{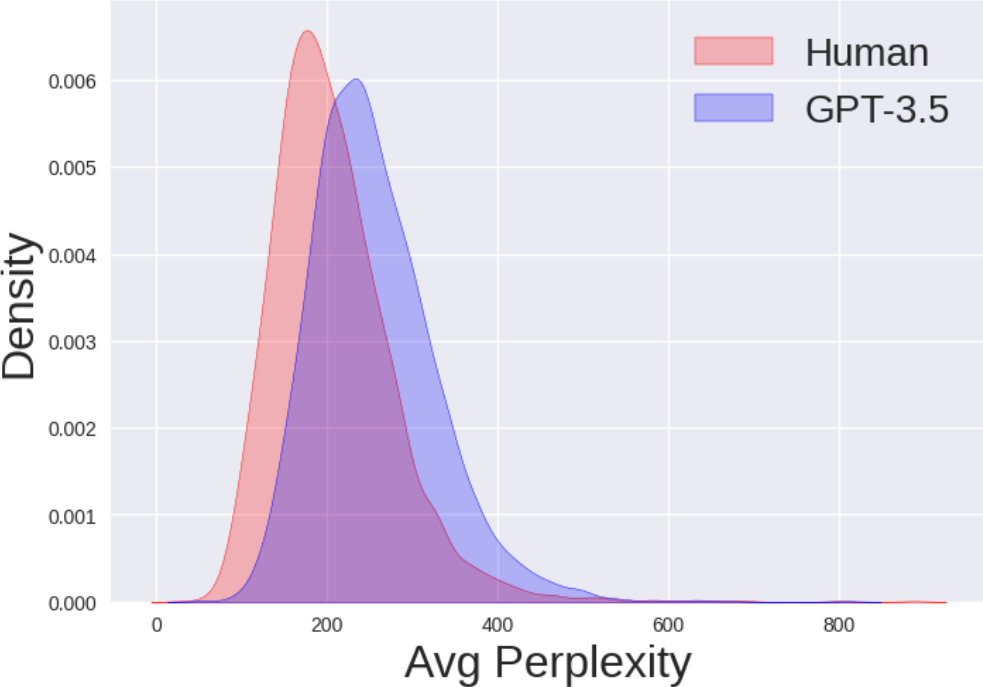}
        \caption{GPT-3.5}
        \label{fig:perp_gpt35}
    \end{subfigure}
    \hfill
    \begin{subfigure}[b]{0.19\textwidth}
        \centering
        \includegraphics[width=\textwidth]{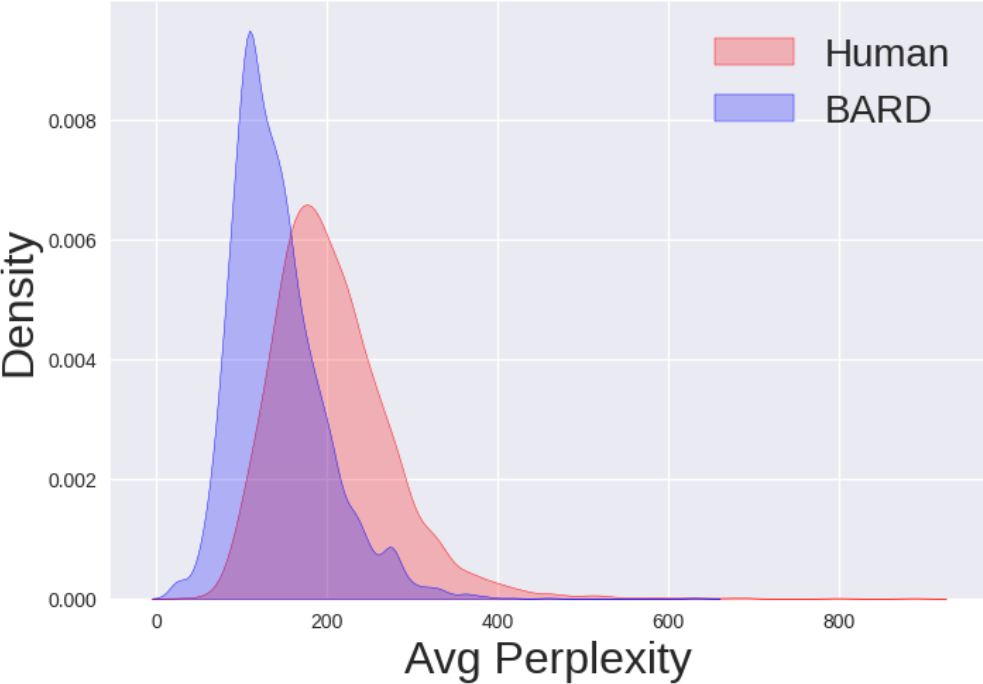}
        \caption{BARD}
        \label{fig:perp_bard}
    \end{subfigure}
    \hfill
    \begin{subfigure}[b]{0.19\textwidth}
        \centering
        \includegraphics[width=\textwidth]{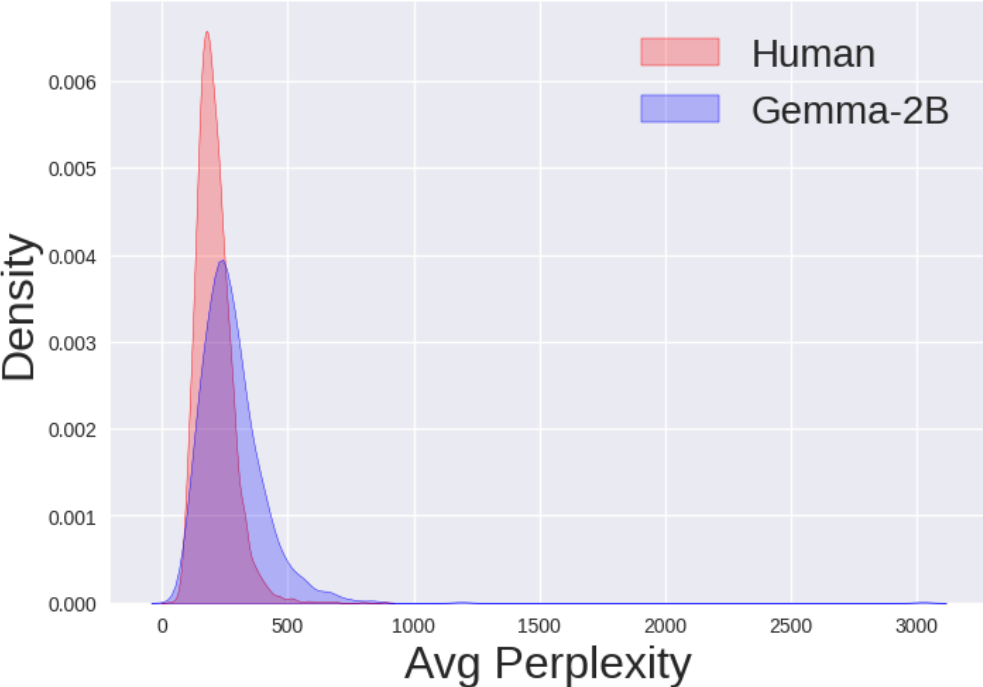}
        \caption{Gemma-2B}
        \label{fig:perp_gemma2b}
    \end{subfigure}
    \hfill
    \begin{subfigure}[b]{0.19\textwidth}
        \centering
        \includegraphics[width=\textwidth]{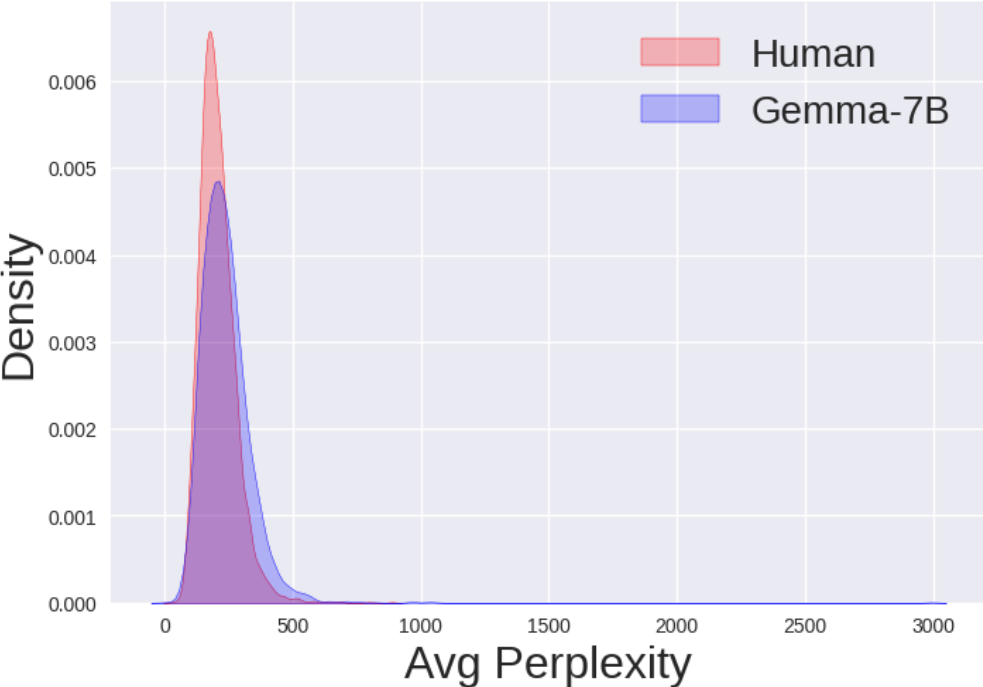}
        \caption{Gemma-7B}
        \label{fig:perp_gemma7b}
    \end{subfigure}
    \caption{Perplexity estimation for models. The perplexities of the responses generated by these LLMs are nearly identical to those of human texts. This similarity in perplexity makes it an unreliable factor for distinguishing between  human and AI-generated text.}
    \label{fig:perplexity_estimation}
\end{figure*}

\subsection{J-Guard}
Journalism Guided Adversarially Robust Detection of AI-generated News (J-Guard) \cite{kumarage2023jguard} is a framework designed to tackle the growing issue of AI-generated news. J-Guard leverages stylistic cues derived from journalistic features to distinguish human-written articles from AI-generated news articles. The premise is that deviation from the Associated Press (AP) Stylebook standards can indicate that an article is AI-generated. The framework extracts various journalistic features to quantify these deviations including, organization and grammar standards such as mean word count, word count of leading paragraph, punctuation use and standard formatting violations such as date, time and number formats. 

\subsection{ConDA}
The Contrastive domain adaptation framework (ConDA) \cite{bhattacharjee2023conda} addresses the problem of AI-generated text detection by framing it as an unsupervised domain adaptation task where the domains are different LLMs. The framework assumes access to labeled source data and unlabeled target data. This framework blends standard domain adaptation techniques with the representation power of contrastive learning to learn domain invariant representations that are effective for the final unsupervised detection task. ConDA leverages the power of both, unsupervised domain adaptation and self-supervised representation learning.

\section{Overall Analysis}
\vspace{-2mm}
Table \ref{tab:Main_results} provides a summary of the results for the AGTD techniques. Our experiments highlight the fragility of existing AGTD methods. RADAR and ConDA exhibit poor performance in effectively detecting AI-generated text. 
While intrinsic dimension estimation can effectively differentiate the responses of certain models as AI-generated, its performance is not consistent across models. This is evident from the variation in intrinsic dimensions across models. J-Guard outperforms the other techniques, but struggles in cross-model scenarios, revealing its limitations, as seen in Figure \ref{fig:jguard_heatmap}. Additional results are provided in Appendix \ref{sec:app_results}. \\
\indent We found that responses generated by black-box LLMs such as GPT-4, GPT-3.5 and BARD are particularly challenging to detect, possibly due to their large parameter sizes. In contrast, outputs from open-source models  like Gemma are easier to identify. Interestingly, we did not observe a significant difference in detectability across open-source models with varying parameter sizes. As the LLMs increasingly generate human-like text, detection becomes more difficult, since most techniques rely on comparing AI-generated text to human-written content.

\begin{table*}[!h]
\centering
\resizebox{1\textwidth}{!}{
\begin{tabular}{cccccccccc}
\hline
\multirow{2}{*}{\textbf{\begin{tabular}[c]{@{}c@{}}Detection \\ Techniques\end{tabular}}} & \multirow{2}{*}{\textbf{Models}} & \multicolumn{4}{c|}{\textbf{\begin{tabular}[c]{@{}c@{}}News Source 1\\ {[}BBC Data{]}\end{tabular}}} & \multicolumn{4}{c}{\textbf{\begin{tabular}[c]{@{}c@{}}News Source 2\\ {[}NDTV Data{]}\end{tabular}}} \\ \cline{3-10} 
 &  & \textbf{Accuracy} & \textbf{Precision} & \textbf{Recall} & \textbf{F1-score} & \textbf{Accuracy} & \textbf{Precision} & \textbf{Recall} & \textbf{F1-score} \\ \hline
\multirow{5}{*}{\textbf{RADAR}} & GPT-4 & \cellcolor[HTML]{FF7251}37.486 & \cellcolor[HTML]{FF5353}2.985 & \cellcolor[HTML]{FF5353}0.795 & \cellcolor[HTML]{FF5353}1.255 & \cellcolor[HTML]{FF8751}49.205 & \cellcolor[HTML]{FF7251}25.862 & \cellcolor[HTML]{FF5353}0.852 & \cellcolor[HTML]{FF5353}1.650 \\
 & GPT-3.5 & \cellcolor[HTML]{FF7251}37.089 & \cellcolor[HTML]{FF5353}0.000 & \cellcolor[HTML]{FF5353}0.000 & \cellcolor[HTML]{FF5353}0.000 & \cellcolor[HTML]{FF8751}48.959 & \cellcolor[HTML]{FF5353}12.838 & \cellcolor[HTML]{FF5353}0.360 & \cellcolor[HTML]{FF5353}0.699 \\
 & BARD & \cellcolor[HTML]{FFC950}72.211 & \cellcolor[HTML]{FF5353}14.634 &  \cellcolor[HTML]{FFAB51}50.000 & \cellcolor[HTML]{FF7251}22.642 &  \cellcolor[HTML]{FFAB51}53.380 & \cellcolor[HTML]{FFC950}79.024 & \cellcolor[HTML]{FF5353}9.203 & \cellcolor[HTML]{FF5353}16.486 \\
 & Gemma-2B &  \cellcolor[HTML]{FFAB51}50.748 &  \cellcolor[HTML]{FFAB51}52.448 & \cellcolor[HTML]{FF5353}16.026 & \cellcolor[HTML]{FF7251}24.55 &  \cellcolor[HTML]{FFAB51}53.557 & \cellcolor[HTML]{FFC450}61.423 & \cellcolor[HTML]{FF5353}19.125 & \cellcolor[HTML]{FF7251}29.169 \\
 & Gemma-7B & \cellcolor[HTML]{FFC450}62.103 & \cellcolor[HTML]{FFC950}71.616 & \cellcolor[HTML]{FF8751}40.098 &  \cellcolor[HTML]{FFAB51}51.411 & \cellcolor[HTML]{FFC450}64.415 & \cellcolor[HTML]{FFE24F}80.287 & \cellcolor[HTML]{FF7251}38.213 &  \cellcolor[HTML]{FFAB51}51.781 \\ \hline
\multirow{5}{*}{\textbf{J-Guard}} & GPT-4 & \cellcolor[HTML]{7CC84F}98.440 & \cellcolor[HTML]{58C050}99.718 & \cellcolor[HTML]{7CC84F}97.245 & \cellcolor[HTML]{7CC84F}98.466 & \cellcolor[HTML]{58C050}99.242 & \cellcolor[HTML]{58C050}99.229 & \cellcolor[HTML]{58C050}99.229 & \cellcolor[HTML]{58C050}99.229 \\
 & GPT-3.5 & \cellcolor[HTML]{58C050}99.291 & \cellcolor[HTML]{58C050}99.128 & \cellcolor[HTML]{58C050}99.417 & \cellcolor[HTML]{58C050}99.272 & \cellcolor[HTML]{7CC84F}98.958 & \cellcolor[HTML]{58C050}99.606 & \cellcolor[HTML]{7CC84F}98.249 & \cellcolor[HTML]{7CC84F}98.923 \\
 & BARD & \cellcolor[HTML]{58C050}99.007 & \cellcolor[HTML]{58C050}99.709 & \cellcolor[HTML]{7CC84F}98.281 & \cellcolor[HTML]{7CC84F}98.990 & \cellcolor[HTML]{58C050}99.290 & \cellcolor[HTML]{58C050}99.505 & \cellcolor[HTML]{58C050}99.016 & \cellcolor[HTML]{58C050}99.260 \\
 & Gemma-2B & \cellcolor[HTML]{58C050}99.467 & \cellcolor[HTML]{58C050}99.454 & \cellcolor[HTML]{58C050}99.454 & \cellcolor[HTML]{58C050}99.454 & \cellcolor[HTML]{58C050}99.344 & \cellcolor[HTML]{7CC84F}98.996 & \cellcolor[HTML]{58C050}99.711 & \cellcolor[HTML]{58C050}99.352 \\
 & Gemma-7B & \cellcolor[HTML]{58C050}99.237 & \cellcolor[HTML]{58C050}99.246 & \cellcolor[HTML]{58C050}99.246 & \cellcolor[HTML]{58C050}99.246 & \cellcolor[HTML]{58C050}99.733 & \cellcolor[HTML]{58C050}99.529 & \cellcolor[HTML]{58C050}99.947 & \cellcolor[HTML]{58C050}99.738 \\ \hline
\multirow{5}{*}{\textbf{ConDA}} & GPT-4 & \cellcolor[HTML]{FF8751}43.445 & \cellcolor[HTML]{FF8751}42.212 & \cellcolor[HTML]{FF7251}35.528 & \cellcolor[HTML]{FF7251}38.582 &  \cellcolor[HTML]{FFAB51}51.856 &  \cellcolor[HTML]{FFAB51}52.736 & \cellcolor[HTML]{FF7251}35.770 & \cellcolor[HTML]{FF8751}42.627 \\
 & GPT-3.5 & \cellcolor[HTML]{FF8751}45.658 & \cellcolor[HTML]{FF8751}45.099 & \cellcolor[HTML]{FF7251}39.955 & \cellcolor[HTML]{FF8751}42.371 &  \cellcolor[HTML]{FFAB51}50.587 &  \cellcolor[HTML]{FFAB51}50.899 & \cellcolor[HTML]{FF7251}33.232 & \cellcolor[HTML]{FF8751}40.211 \\
 & BARD & \cellcolor[HTML]{FF8751}47.645 & \cellcolor[HTML]{FF8751}47.456 & \cellcolor[HTML]{FF8751}43.927 & \cellcolor[HTML]{FF8751}45.623 &  \cellcolor[HTML]{FFAB51}55.245 &  \cellcolor[HTML]{FFAB51}57.030 & \cellcolor[HTML]{FF8751}42.548 & \cellcolor[HTML]{FF8751}48.736 \\
 & Gemma-2B &  \cellcolor[HTML]{FFAB51}53.739 &  \cellcolor[HTML]{FFAB51}53.700 &  \cellcolor[HTML]{FFAB51}54.274 &  \cellcolor[HTML]{FFAB51}53.985 &  \cellcolor[HTML]{FFAB51}59.679 & \cellcolor[HTML]{FFC450}63.607 & \cellcolor[HTML]{FF8751}45.248 &  \cellcolor[HTML]{FFAB51}52.879 \\
 & Gemma-7B &  \cellcolor[HTML]{FFAB51}52.353 &  \cellcolor[HTML]{FFAB51}52.323 &  \cellcolor[HTML]{FFAB51}52.995 &  \cellcolor[HTML]{FFAB51}52.657 &  \cellcolor[HTML]{FFAB51}57.491 &  \cellcolor[HTML]{FFAB51}59.537 & \cellcolor[HTML]{FF8751}46.762 & \cellcolor[HTML]{FFAB51}52.382 \\ \hline
\multirow{5}{*}{\textbf{RAIDAR}} & GPT-4 & \cellcolor[HTML]{FFC450}66.147 & \cellcolor[HTML]{FFC450}65.833 & \cellcolor[HTML]{FFC450}67.134 & \cellcolor[HTML]{FFC450}66.48 & \cellcolor[HTML]{FFC450}69.584 & \cellcolor[HTML]{FFC450}67.814 & \cellcolor[HTML]{FFC950}74.551 & \cellcolor[HTML]{FFC950}71.023 \\
 & GPT-3.5 & \cellcolor[HTML]{FFC450}64.589 & \cellcolor[HTML]{FFC450}64.345 & \cellcolor[HTML]{FFC450}65.439 & \cellcolor[HTML]{FFC450}64.888 & \cellcolor[HTML]{FFC450}60.549 & \cellcolor[HTML]{FFC450}60.43 & \cellcolor[HTML]{FFC450}61.116 & \cellcolor[HTML]{FFC450}60.771 \\
 & BARD & \cellcolor[HTML]{FFC950}74.22 & \cellcolor[HTML]{FFC950}74.085 & \cellcolor[HTML]{FFC950}74.504 & \cellcolor[HTML]{FFC950}74.294 & \cellcolor[HTML]{FFE24F}89.64 & \cellcolor[HTML]{FFE24F}88.582 & \cellcolor[HTML]{B8D44F}91.012 & \cellcolor[HTML]{FFE24F}89.781 \\
 & Gemma-2B & \cellcolor[HTML]{7CC84F}98.404 & \cellcolor[HTML]{7CC84F}98.925 & \cellcolor[HTML]{7CC84F}97.872 & \cellcolor[HTML]{7CC84F}98.396 & \cellcolor[HTML]{7CC84F}96.939 & \cellcolor[HTML]{7CC84F}96.532 & \cellcolor[HTML]{7CC84F}97.376 & \cellcolor[HTML]{7CC84F}96.952 \\
 & Gemma-7B & \cellcolor[HTML]{7CC84F}98.476 & \cellcolor[HTML]{58C050}99.688 & \cellcolor[HTML]{7CC84F}97.256 & \cellcolor[HTML]{7CC84F}98.457 & \cellcolor[HTML]{B8D44F}94.712 & \cellcolor[HTML]{7CC84F}95.146 & \cellcolor[HTML]{B8D44F}94.231 & \cellcolor[HTML]{B8D44F}94.686 \\ \hline
\textbf{} &  & \multicolumn{2}{c}{\textbf{MLE}} & \multicolumn{2}{c}{\textbf{PHD}} & \multicolumn{2}{c}{\textbf{MLE}} & \multicolumn{2}{c}{\textbf{PHD}} \\ \hline
\multirow{6}{*}{\textbf{\begin{tabular}[c]{@{}c@{}}Intrinsic \\ Dimension\end{tabular}}} & \begin{tabular}[c]{@{}c@{}}Human\\ written\end{tabular} & \multicolumn{2}{c}{10.016} & \multicolumn{2}{c}{6.967} & \multicolumn{2}{c}{9.592} & \multicolumn{2}{c}{6.781} \\
 & GPT-4 & \multicolumn{2}{c}{9.541} & \multicolumn{2}{c}{7.002} & \multicolumn{2}{c}{9.416} & \multicolumn{2}{c}{6.900} \\
 & GPT-3.5 & \multicolumn{2}{c}{9.796} & \multicolumn{2}{c}{6.882} & \multicolumn{2}{c}{9.549} & \multicolumn{2}{c}{6.720} \\
 & BARD & \multicolumn{2}{c}{7.272} & \multicolumn{2}{c}{3.120} & \multicolumn{2}{c}{7.061} & \multicolumn{2}{c}{3.105} \\
 & Gemma-2B & \multicolumn{2}{c}{4.368} & \multicolumn{2}{c}{3.004} & \multicolumn{2}{c}{4.537} & \multicolumn{2}{c}{3.118} \\
 & Gemma-7B & \multicolumn{2}{c}{5.354} & \multicolumn{2}{c}{3.597} & \multicolumn{2}{c}{5.577} & \multicolumn{2}{c}{3.744} \\ \hline
\end{tabular}
}
\caption{Results showcasing the efficacy of various AI-Generated Text Detection (AGTD) methods. Results compare performance metrics across different techniques, highlighting their effectiveness in accurately identifying AI-generated text versus human-written text}
\label{tab:Main_results}
\end{table*}

\section{AI Detectability Index for Hindi ($ADI_{hi}$)}
Given the rapid advancements in LLMs, the existing AGTD techniques may prove to be ineffective for the newer models. We propose AI Detectability Index for Hindi $(ADI_{hi})$ as a benchmark to assess and rank LLMs according to the detectability of the model's responses.
\begin{figure}[H] 
    \centering 
    \includegraphics[width= \linewidth]{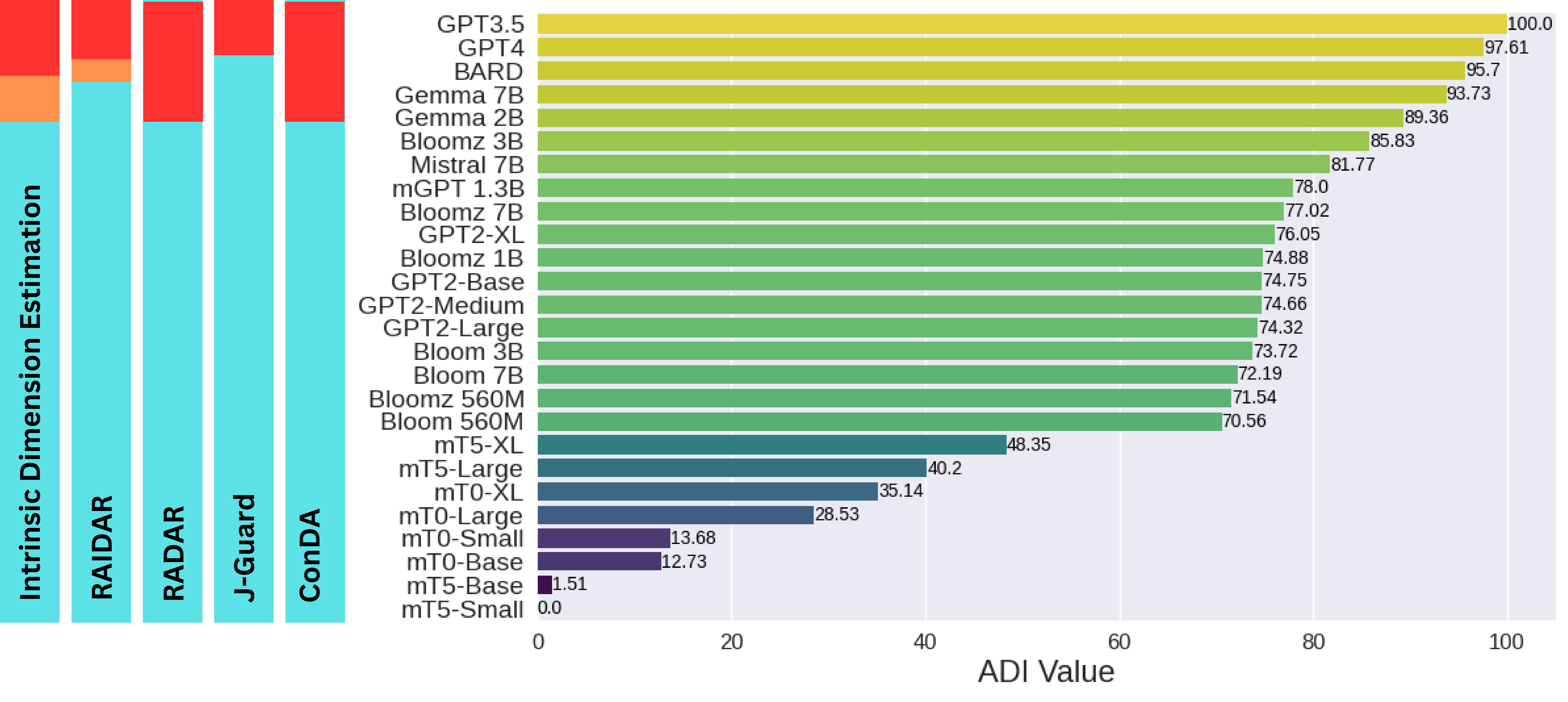} 
    \caption{Right: ADI Spectrum of a diverse set of LLMs based on their detectability.
    Left: Bars represent AI-generated text detectability by a specific method. Blue indicates fully detectable, orange indicates unreliable detection, and red indicates not detectable.}
    \label{fig:ADI}
\end{figure}

\subsection{Limitations of ADI proposed by \citet{chakraborty2023counter}}
Previous work by \citet{chakraborty2023counter} focuses on perplexity and burstiness as the factors to quantify the detectability of the model responses. However, the text generated by the newer LLMs is indistinguishable from human-written text. Furthermore, the perplexity and burstiness of larger models 
like GPT-4, GPT-3.5 and BARD
resemble human-written text with a small variance, as illustrated in Figure \ref{fig:perplexity_estimation}. \citet{liang2023gpt} and \citet{chakraborty2023possibilities} have shown that perplexity and burstiness are not reliable detectors of human writing. To overcome this, we assess the divergence between the AI-generated text and human-written text to quantify the detectability of the model. 
\subsection{ADI - Proposed by us}
For every pair of human-written article and AI-generated article in our $AG_{hi}$ dataset, we identify the common set of words between them, denoted as $V$. For each word $w_i$ in $V$, we extract the sentences $S_h$ and $S_{ai}$ from the human-written and AI-generated text respectively, containing $w_i$. Using these sentences, we generate the co-occurrence vectors $C_h$ and $C_{ai}$.

\begin{figure*}[htb]
    \centering
    \small
    \includegraphics[width=\linewidth]{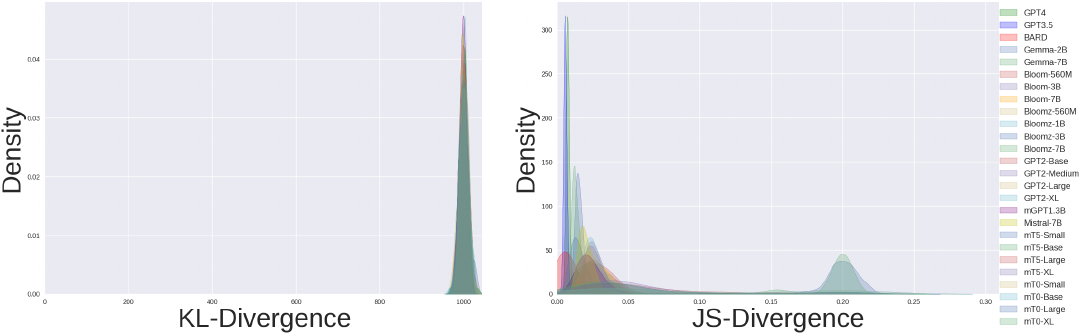}
    \caption{Comparison of Divergence values using KL-Divergence and Jenson-Shannon Divergence. When employing KL-divergence, all the divergence values cluster around a point. In contrast, using JS-Divergence reveals a more distributed spectrum of values. \textit{Note: The divergence values of 1000 in the KL-divergence case are merely indicative; the actual value approaches infinity.} }
    \label{fig:divergence_comparison}
\end{figure*}


This helps us capture the semantic, syntactic, and lexical features of the text. These co-occurrence vectors are then transformed into probability distributions by normalizing the frequency counts. A detailed explanation of this can be found in Appendix \ref{sec:distribution} To quantify the divergence between these distributions we initially employed KL-divergence (KLD) \cite{csiszar1975divergence}. However, KLD lacks the capability to handle zero values in the probability distribution. This results in the divergence values escalating to infinity, causing the models to cluster closely together and overlap, rendering the spectrum discernible.
To address this we adopt Jenson-Shannon divergence (JSD) \cite{JSD}, a more robust and symmetric version of KL-divergence. Figure \ref{fig:divergence_comparison} illustrates the comparison between the performance of KLD and JSD. To assess overall divergence, summation has been taken over all the data points $U$ as depicted in Equation \ref{eq:ADI}. After calculating the mean divergence of all 26 models initially, we adopt Yeo-Johnson power transformation \cite{10.1093/biomet/87.4.954} to make the data more normally distributed. This is crucial for balanced and unbiased scaling. Finally, the values are scaled between 0-100 using the min-max normalization \cite{Min-Max} for better readability and interpretability. The resulting ADIs are then ranked and scaled providing a comparative spectrum as presented in Figure \ref{fig:ADI}. A  higher value of $ADI_{hi}$ corresponds to greater difficulty in detecting the model's responses as AI-generated.

\vspace{-6mm}
\begin{equation}
  ADI_{x} = \frac{1}{U} * \sum_{j=1}^{U} \Big[\{\sum_{i=1}^{|V|} JSD_{j}(P_{h}^{i}\ ||\ P_{ai}^{i})\} * \frac{1}{|V|} \Big]
  \label{eq:ADI}
\end{equation}

Three groups of models can be observed from the ADI spectrum, namely: easy-to-detect, detectable and difficult-to-detect. The mT0 and mT5 models are situated in the realm of easy-to-detect range while models like Bloom, Bloomz and GPT-2 are detectable. The remaining models are regarded as nearly undetectable using the existing SoTA AGTD techniques.

From the methods we considered, it is unlikely that any of them would be effective for models with high ADI, as shown by our experiments and results. With advancements in LLM technology, the current AGTD methods would become more ineffective. Recognizing this, the ADI spectrum will serve as a crucial tool for assessing the detectability of LLMs.

\vspace{-2mm}
\section{Conclusion}
Our research contends that SoTA AGTD techniques are susceptible to fragility. These techniques, while effective in English often struggle or perform poorly when applied to Hindi, highlighting the need for language-specific considerations in AI-generated text detection. We experimented with 26 distinct LLMs to create the $AG_{hi}$ dataset and support the assertion. We introduce the AI Detectability Index for Hindi ($ADI_{hi}$), and present a means to assess and rank LLMs based on their detectability levels. The excitement and success of LLMs have resulted in their extensive proliferation, and this trend is anticipated to persist regardless of the future course it takes. In light of this, the \ul{{CT}$^2$} benchmark and the \emph{$ADI_{hi}$} will continue to play a vital role in catering to the scientific community.

\newpage
\section{Discussion And Limitations}

In this paper, we address the critical issue of AI-generated text detection in the context of the Hindi language. Despite the valuable contributions, there are certain limitations inherent in this work as discussed in the following points.

\begin{itemize}
    \item{Exploring the temperature hyperparameter: 
    We experiment with temperature hyperparameters while selecting the LLMs. However, we generate $AG_{hi}$ considering a constant temperature. Investigating the influence of temperature on the detectability of the generated text would provide valuable insights.}

    \item{Text consistency in experiments:
    We generate only a single response per headline while forming the dataset. However, future work can involve generating multiple responses for each headline and evaluating the detectability of these responses.}

    \item{Temporal Limitations:
    Due to the absence of an archive feature on the BBC and NDTV websites, we opted to compile a varied set of headlines without being bound by temporal limitations. However, our selection criteria for LLMs focuses on the quality of the text generated. Furthermore, none of the AGTD techniques evaluated in the study assess the text based on its factuality. Therefore, this decision does not affect the validity of our results.
    }

    \item{Generalization to other languages:
    The study primarily focuses on the Hindi language, and the findings may not be directly applicable to other languages with distinct linguistic characteristics. Future research could explore the extension of these insights to a broader range of languages.}
    
    \item{Evolution of LLMs:
    The rapidly evolving nature of LLMs raises the possibility that newer models, not included in the study, may exhibit different behaviors. As such, the generalizability of the findings to future LLMs may be limited.}

    \item{Dynamic AI-generated text detection landscape:
    The research evaluates AGTD techniques based on the current state of detection methods. However, the dynamic nature of the AI-generated text detection methods suggests that new strategies may emerge, potentially impacting the long-term efficacy of the proposed techniques.}
    

\end{itemize}

\section{Ethical Considerations}
Our experiments reveal the constraints of AGTD methods in Hindi. It is crucial to note that while we envision $ADI_{hi}$ as a tool for constructive purposes, there exists the potential for misuse by malicious entities, especially in generating AI-generated text like fake news that is indistinguishable from human-written content. We strongly caution against any such misuse of our findings.

\section*{Acknowledgments}
We would like to thank Harshavardhan Nemani, Mann Khatri, Prashant Kodali, Ritwik Mishra and other members of Precog group, IIIT-Hyderabad for their valuable feedback.

\bibliography{anthology,custom}

\begin{thebibliography}{37}
\expandafter\ifx\csname natexlab\endcsname\relax\def\natexlab#1{#1}\fi

\bibitem[{Bard(2023)}]{Bard2023}
Google~AI Bard. 2023.
\newblock \href {https://blog.google/technology/ai/bard-google-ai-search-updates/} {An important next step on our ai journey}.
\newblock [Online; accessed 06-December-2023].

\bibitem[{Basit et~al.(2021)Basit, Zafar, Liu, Javed, Jalil, and Kifayat}]{basit2021comprehensive}
Abdul Basit, Maham Zafar, Xuan Liu, Abdul~Rehman Javed, Zunera Jalil, and Kashif Kifayat. 2021.
\newblock A comprehensive survey of ai-enabled phishing attacks detection techniques.
\newblock \emph{Telecommunication Systems}, 76:139--154.

\bibitem[{Bhattacharjee et~al.(2023)Bhattacharjee, Kumarage, Moraffah, and Liu}]{bhattacharjee2023conda}
Amrita Bhattacharjee, Tharindu Kumarage, Raha Moraffah, and Huan Liu. 2023.
\newblock \href {http://arxiv.org/abs/2309.03992} {Conda: Contrastive domain adaptation for ai-generated text detection}.

\bibitem[{BigScience(2022)}]{BigScience_BLOOM_2022}
BigScience. 2022.
\newblock Bigscience language open-science open-access multilingual (bloom) language model.
\newblock \url{https://huggingface.co/bigscience/bloom}.
\newblock International, May 2021-May 2022.

\bibitem[{Chakraborty et~al.(2023{\natexlab{a}})Chakraborty, Tonmoy, Zaman, Sharma, Barman, Gupta, Gautam, Kumar, Jain, Chadha et~al.}]{chakraborty2023counter}
Megha Chakraborty, SM~Tonmoy, SM~Zaman, Krish Sharma, Niyar~R Barman, Chandan Gupta, Shreya Gautam, Tanay Kumar, Vinija Jain, Aman Chadha, et~al. 2023{\natexlab{a}}.
\newblock Counter turing test ct\^{} 2: Ai-generated text detection is not as easy as you may think--introducing ai detectability index.
\newblock \emph{arXiv preprint arXiv:2310.05030}.

\bibitem[{Chakraborty et~al.(2023{\natexlab{b}})Chakraborty, Bedi, Zhu, An, Manocha, and Huang}]{chakraborty2023possibilities}
Souradip Chakraborty, Amrit~Singh Bedi, Sicheng Zhu, Bang An, Dinesh Manocha, and Furong Huang. 2023{\natexlab{b}}.
\newblock \href {http://arxiv.org/abs/2304.04736} {On the possibilities of ai-generated text detection}.

\bibitem[{Chen et~al.(2023)Chen, Ye, Zu, Xu, Zheng, Peng, Zhou, Gui, Zhang, and Huang}]{chen2023robust}
Xuanting Chen, Junjie Ye, Can Zu, Nuo Xu, Rui Zheng, Minlong Peng, Jie Zhou, Tao Gui, Qi~Zhang, and Xuanjing Huang. 2023.
\newblock How robust is gpt-3.5 to predecessors? a comprehensive study on language understanding tasks.
\newblock \emph{arXiv preprint arXiv:2303.00293}.

\bibitem[{Chernyaeva et~al.(2022)Chernyaeva, Hong, Park, Kim, and Ren}]{chernyaeva2022ai}
O~Chernyaeva, TH~Hong, YK~Park, YH~Kim, and G~Ren. 2022.
\newblock Ai generating and detecting manipulated online customers reviews.
\newblock pages 270--275.

\bibitem[{Csisz{\'a}r(1975)}]{csiszar1975divergence}
Imre Csisz{\'a}r. 1975.
\newblock I-divergence geometry of probability distributions and minimization problems.
\newblock \emph{The annals of probability}, pages 146--158.

\bibitem[{Hu et~al.(2023)Hu, Chen, and Ho}]{hu2023radar}
Xiaomeng Hu, Pin-Yu Chen, and Tsung-Yi Ho. 2023.
\newblock \href {http://arxiv.org/abs/2307.03838} {Radar: Robust ai-text detection via adversarial learning}.

\bibitem[{Jenson-Shannon-Divergence()}]{JSD}
Jenson-Shannon-Divergence.
\newblock \href {https://en.wikipedia.org/wiki/Jensen%E2%80%93Shannon_divergence} {[link]}.

\bibitem[{Jiang et~al.(2023)Jiang, Sablayrolles, Mensch, Bamford, Chaplot, de~las Casas, Bressand, Lengyel, Lample, Saulnier, Lavaud, Lachaux, Stock, Scao, Lavril, Wang, Lacroix, and Sayed}]{jiang2023mistral}
Albert~Q. Jiang, Alexandre Sablayrolles, Arthur Mensch, Chris Bamford, Devendra~Singh Chaplot, Diego de~las Casas, Florian Bressand, Gianna Lengyel, Guillaume Lample, Lucile Saulnier, Lélio~Renard Lavaud, Marie-Anne Lachaux, Pierre Stock, Teven~Le Scao, Thibaut Lavril, Thomas Wang, Timothée Lacroix, and William~El Sayed. 2023.
\newblock \href {http://arxiv.org/abs/2310.06825} {Mistral 7b}.

\bibitem[{Kirchenbauer et~al.(2023)Kirchenbauer, Geiping, Wen, Katz, Miers, and Goldstein}]{pmlr-v202-kirchenbauer23a}
John Kirchenbauer, Jonas Geiping, Yuxin Wen, Jonathan Katz, Ian Miers, and Tom Goldstein. 2023.
\newblock \href {https://proceedings.mlr.press/v202/kirchenbauer23a.html} {A watermark for large language models}.
\newblock In \emph{Proceedings of the 40th International Conference on Machine Learning}, volume 202 of \emph{Proceedings of Machine Learning Research}, pages 17061--17084. PMLR.

\bibitem[{Kirchenbauer et~al.(2024)Kirchenbauer, Geiping, Wen, Shu, Saifullah, Kong, Fernando, Saha, Goldblum, and Goldstein}]{kirchenbauer2024reliability}
John Kirchenbauer, Jonas Geiping, Yuxin Wen, Manli Shu, Khalid Saifullah, Kezhi Kong, Kasun Fernando, Aniruddha Saha, Micah Goldblum, and Tom Goldstein. 2024.
\newblock \href {http://arxiv.org/abs/2306.04634} {On the reliability of watermarks for large language models}.

\bibitem[{Kreps et~al.(2022)Kreps, McCain, and Brundage}]{kreps2022all}
Sarah Kreps, R~Miles McCain, and Miles Brundage. 2022.
\newblock All the news that’s fit to fabricate: Ai-generated text as a tool of media misinformation.
\newblock \emph{Journal of experimental political science}, 9(1):104--117.

\bibitem[{Krishna et~al.(2023)Krishna, Song, Karpinska, Wieting, and Iyyer}]{NEURIPS2023_575c4500}
Kalpesh Krishna, Yixiao Song, Marzena Karpinska, John Wieting, and Mohit Iyyer. 2023.
\newblock \href {https://proceedings.neurips.cc/paper_files/paper/2023/file/575c450013d0e99e4b0ecf82bd1afaa4-Paper-Conference.pdf} {Paraphrasing evades detectors of ai-generated text, but retrieval is an effective defense}.
\newblock In \emph{Advances in Neural Information Processing Systems}, volume~36, pages 27469--27500. Curran Associates, Inc.

\bibitem[{Kuditipudi et~al.(2023)Kuditipudi, Thickstun, Hashimoto, and Liang}]{kuditipudi2023robust}
Rohith Kuditipudi, John Thickstun, Tatsunori Hashimoto, and Percy Liang. 2023.
\newblock Robust distortion-free watermarks for language models.
\newblock \emph{arXiv preprint arXiv:2307.15593}.

\bibitem[{Kumarage et~al.(2023)Kumarage, Bhattacharjee, Padejski, Roschke, Gillmor, Ruston, Liu, and Garland}]{kumarage2023jguard}
Tharindu Kumarage, Amrita Bhattacharjee, Djordje Padejski, Kristy Roschke, Dan Gillmor, Scott Ruston, Huan Liu, and Joshua Garland. 2023.
\newblock \href {http://arxiv.org/abs/2309.03164} {J-guard: Journalism guided adversarially robust detection of ai-generated news}.

\bibitem[{Li et~al.(2023)Li, Wang, Ren, Sun, and Qiu}]{li2023origin}
Linyang Li, Pengyu Wang, Ke~Ren, Tianxiang Sun, and Xipeng Qiu. 2023.
\newblock \href {http://arxiv.org/abs/2304.14072} {Origin tracing and detecting of llms}.

\bibitem[{Liang et~al.(2023)Liang, Yuksekgonul, Mao, Wu, and Zou}]{liang2023gpt}
Weixin Liang, Mert Yuksekgonul, Yining Mao, Eric Wu, and James Zou. 2023.
\newblock \href {http://arxiv.org/abs/2304.02819} {Gpt detectors are biased against non-native english writers}.

\bibitem[{Lu et~al.(2024)Lu, Liu, He, Wang, Ong, and Tang}]{lu2024large}
Ning Lu, Shengcai Liu, Rui He, Qi~Wang, Yew-Soon Ong, and Ke~Tang. 2024.
\newblock \href {http://arxiv.org/abs/2305.10847} {Large language models can be guided to evade ai-generated text detection}.

\bibitem[{Mao et~al.(2024)Mao, Vondrick, Wang, and Yang}]{mao2024raidar}
Chengzhi Mao, Carl Vondrick, Hao Wang, and Junfeng Yang. 2024.
\newblock \href {http://arxiv.org/abs/2401.12970} {Raidar: generative ai detection via rewriting}.

\bibitem[{Mitchell et~al.(2023)Mitchell, Lee, Khazatsky, Manning, and Finn}]{10.5555/3618408.3619446}
Eric Mitchell, Yoonho Lee, Alexander Khazatsky, Christopher~D. Manning, and Chelsea Finn. 2023.
\newblock Detectgpt: zero-shot machine-generated text detection using probability curvature.
\newblock In \emph{Proceedings of the 40th International Conference on Machine Learning}, ICML'23. JMLR.org.

\bibitem[{Muennighoff et~al.(2022)Muennighoff, Wang, Sutawika, Roberts, Biderman, Scao, Bari, Shen, Yong, Schoelkopf et~al.}]{muennighoff2022crosslingual}
Niklas Muennighoff, Thomas Wang, Lintang Sutawika, Adam Roberts, Stella Biderman, Teven~Le Scao, M~Saiful Bari, Sheng Shen, Zheng-Xin Yong, Hailey Schoelkopf, et~al. 2022.
\newblock Crosslingual generalization through multitask finetuning.
\newblock \emph{arXiv preprint arXiv:2211.01786}.

\bibitem[{Normalization()}]{Min-Max}
Normalization.
\newblock \href {https://en.wikipedia.org/wiki/Feature_scaling#Rescaling_(min-max_normalization)} {[link]}.

\bibitem[{OpenAI et~al.(2024)OpenAI, Achiam, Adler, Agarwal, Ahmad, Akkaya, Aleman, and et~al.}]{openai2024gpt4technicalreport}
OpenAI, Josh Achiam, Steven Adler, Sandhini Agarwal, Lama Ahmad, Ilge Akkaya, Florencia~Leoni Aleman, and Diogo~Almeida et~al. 2024.
\newblock \href {http://arxiv.org/abs/2303.08774} {Gpt-4 technical report}.

\bibitem[{Papineni et~al.(2002)Papineni, Roukos, Ward, and Zhu}]{10.3115/1073083.1073135}
Kishore Papineni, Salim Roukos, Todd Ward, and Wei-Jing Zhu. 2002.
\newblock \href {https://doi.org/10.3115/1073083.1073135} {Bleu: a method for automatic evaluation of machine translation}.
\newblock In \emph{Proceedings of the 40th Annual Meeting on Association for Computational Linguistics}, ACL '02, page 311–318, USA. Association for Computational Linguistics.

\bibitem[{Radford et~al.(2019)Radford, Wu, Child, Luan, Amodei, Sutskever et~al.}]{radford2019language}
Alec Radford, Jeffrey Wu, Rewon Child, David Luan, Dario Amodei, Ilya Sutskever, et~al. 2019.
\newblock Language models are unsupervised multitask learners.
\newblock \emph{OpenAI blog}, 1(8):9.

\bibitem[{Sadasivan et~al.(2024)Sadasivan, Kumar, Balasubramanian, Wang, and Feizi}]{sadasivan2024aigenerated}
Vinu~Sankar Sadasivan, Aounon Kumar, Sriram Balasubramanian, Wenxiao Wang, and Soheil Feizi. 2024.
\newblock \href {http://arxiv.org/abs/2303.11156} {Can ai-generated text be reliably detected?}

\bibitem[{Shliazhko et~al.(2023)Shliazhko, Fenogenova, Tikhonova, Mikhailov, Kozlova, and Shavrina}]{shliazhko2023mgpt}
Oleh Shliazhko, Alena Fenogenova, Maria Tikhonova, Vladislav Mikhailov, Anastasia Kozlova, and Tatiana Shavrina. 2023.
\newblock \href {http://arxiv.org/abs/2204.07580} {mgpt: Few-shot learners go multilingual}.

\bibitem[{Team et~al.(2024)Team, Mesnard, Hardin, Dadashi, Bhupatiraju, Pathak, and et~al.}]{gemmateam2024gemmaopenmodelsbased}
Gemma Team, Thomas Mesnard, Cassidy Hardin, Robert Dadashi, Surya Bhupatiraju, Shreya Pathak, and Laurent~Sifre et~al. 2024.
\newblock \href {http://arxiv.org/abs/2403.08295} {Gemma: Open models based on gemini research and technology}.

\bibitem[{Tian and Cui(2023)}]{tian2023gptzero}
Edward Tian and Alexander Cui. 2023.
\newblock \href {https://gptzero.me} {Gptzero: Towards detection of ai-generated text using zero-shot and supervised methods}.

\bibitem[{Tulchinskii et~al.(2023)Tulchinskii, Kuznetsov, Kushnareva, Cherniavskii, Barannikov, Piontkovskaya, Nikolenko, and Burnaev}]{tulchinskii2023intrinsic}
Eduard Tulchinskii, Kristian Kuznetsov, Laida Kushnareva, Daniil Cherniavskii, Serguei Barannikov, Irina Piontkovskaya, Sergey Nikolenko, and Evgeny Burnaev. 2023.
\newblock \href {http://arxiv.org/abs/2306.04723} {Intrinsic dimension estimation for robust detection of ai-generated texts}.

\bibitem[{Wikipedia(2023)}]{wiki:List_of_languages_by_number_of_native_speakers}
Wikipedia. 2023.
\newblock {List of languages by number of native speakers} --- {W}ikipedia{,} the free encyclopedia.
\newblock \href{http://en.wikipedia.org/w/index.php?title=List\%20of\%20languages\%20by\%20number\%20of\%20native\%20speakers&oldid=1187628042}{Article Link.}
\newblock [Online; accessed 06-December-2023].

\bibitem[{Xue et~al.(2021)Xue, Constant, Roberts, Kale, Al-Rfou, Siddhant, Barua, and Raffel}]{xue2021mt5}
Linting Xue, Noah Constant, Adam Roberts, Mihir Kale, Rami Al-Rfou, Aditya Siddhant, Aditya Barua, and Colin Raffel. 2021.
\newblock \href {http://arxiv.org/abs/2010.11934} {mt5: A massively multilingual pre-trained text-to-text transformer}.

\bibitem[{Yeo and Johnson(2000)}]{10.1093/biomet/87.4.954}
In‐Kwon Yeo and Richard~A. Johnson. 2000.
\newblock \href {https://doi.org/10.1093/biomet/87.4.954} {{A new family of power transformations to improve normality or symmetry}}.
\newblock \emph{Biometrika}, 87(4):954--959.

\bibitem[{Zhang et~al.(2019)Zhang, Kishore, Wu, Weinberger, and Artzi}]{zhang2019bertscore}
Tianyi Zhang, Varsha Kishore, Felix Wu, Kilian~Q Weinberger, and Yoav Artzi. 2019.
\newblock Bertscore: Evaluating text generation with bert.
\newblock \emph{arXiv preprint arXiv:1904.09675}.

\end{thebibliography}

\appendix

\newpage
\onecolumn
\section*{\textbf{Appendix}}
This section provides supplementary material in the form of additional examples, implementation details,
etc. to bolster the reader’s understanding of the concepts presented in this work.
\section{Model Selection and Data Generation}
In this section we provide additional information about  criteria for model selection, methods used for data generation and hyperparameters applied in generating the dataset.

\subsection{Acceptance and Rejection criteria}\label{sec:acceptance_criteria}
The criteria used to determine acceptance or rejection of a model are as follows:\\
\noindent
\textbf {Language Consistency}: If the response is only in English, the model is rejected.

\noindent
\textbf{Code-Switching}: If the response starts in Hindi but later switches to English, the model is rejected.

\noindent
\textbf{Gibberish Output}: Models that produce unintelligible or gibberish responses are rejected.

\noindent
\textbf{No Output}: Models producing no output are trivially rejected.
\\

\noindent The news headline along with an instruction is prompted to the LLM. We assess a hundred responses from the LLM manually and reject a sample if it produces only English responses, engages in code-switching, generates gibberish output, or fails to produce any output. Besides these criteria, the model must generate five unique Hindi sentences. Therefore, responses that repeat sentences are excluded. If over 70 out of 100 responses meet these rejection criteria, the model is rejected. 

\subsection{Examples from $AG_{hi}$ dataset} \label{sec: Hindi_examples}
\begin{figure*}[!h]
    \centering
    \small
    \includegraphics[width=1\textwidth]{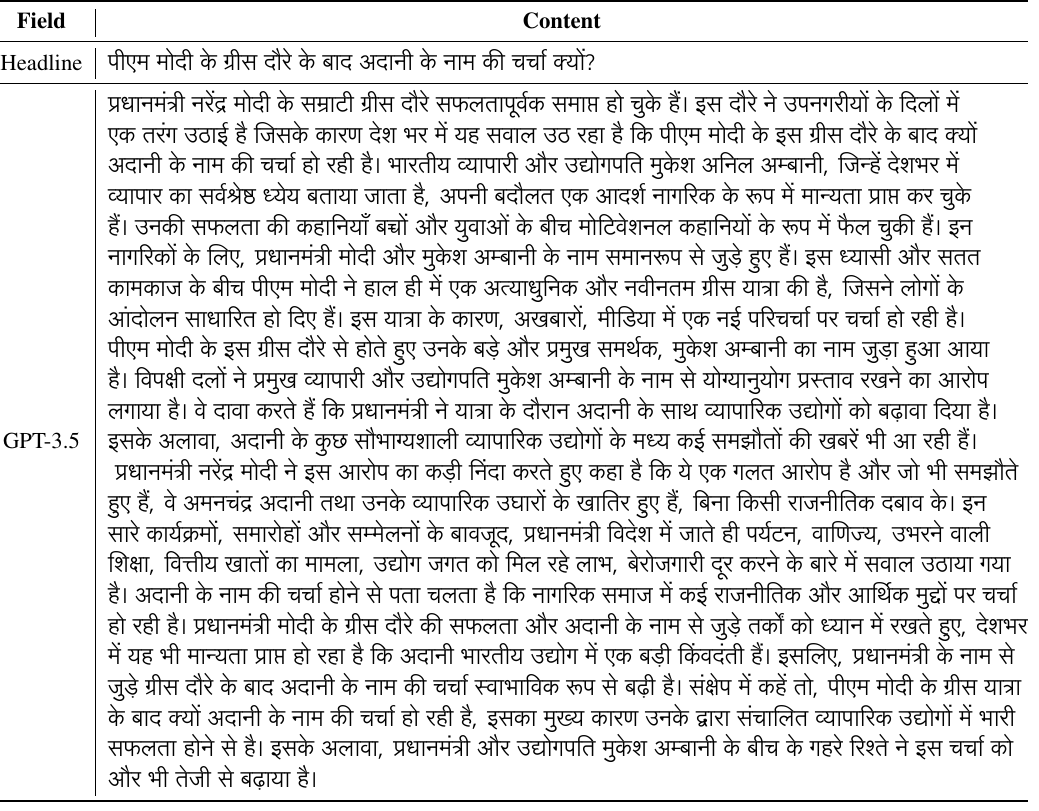}
    
    \caption{Samples of articles generated by accepted models (GPT-4, GPT-3.5, BARD, Gemma-2B and Gemma-7B).}
    \label{fig:Hindi_examples_2}
\end{figure*}

\begin{figure*}[!h]
    \centering
    \small
    \includegraphics[width=1\textwidth, height=20cm]{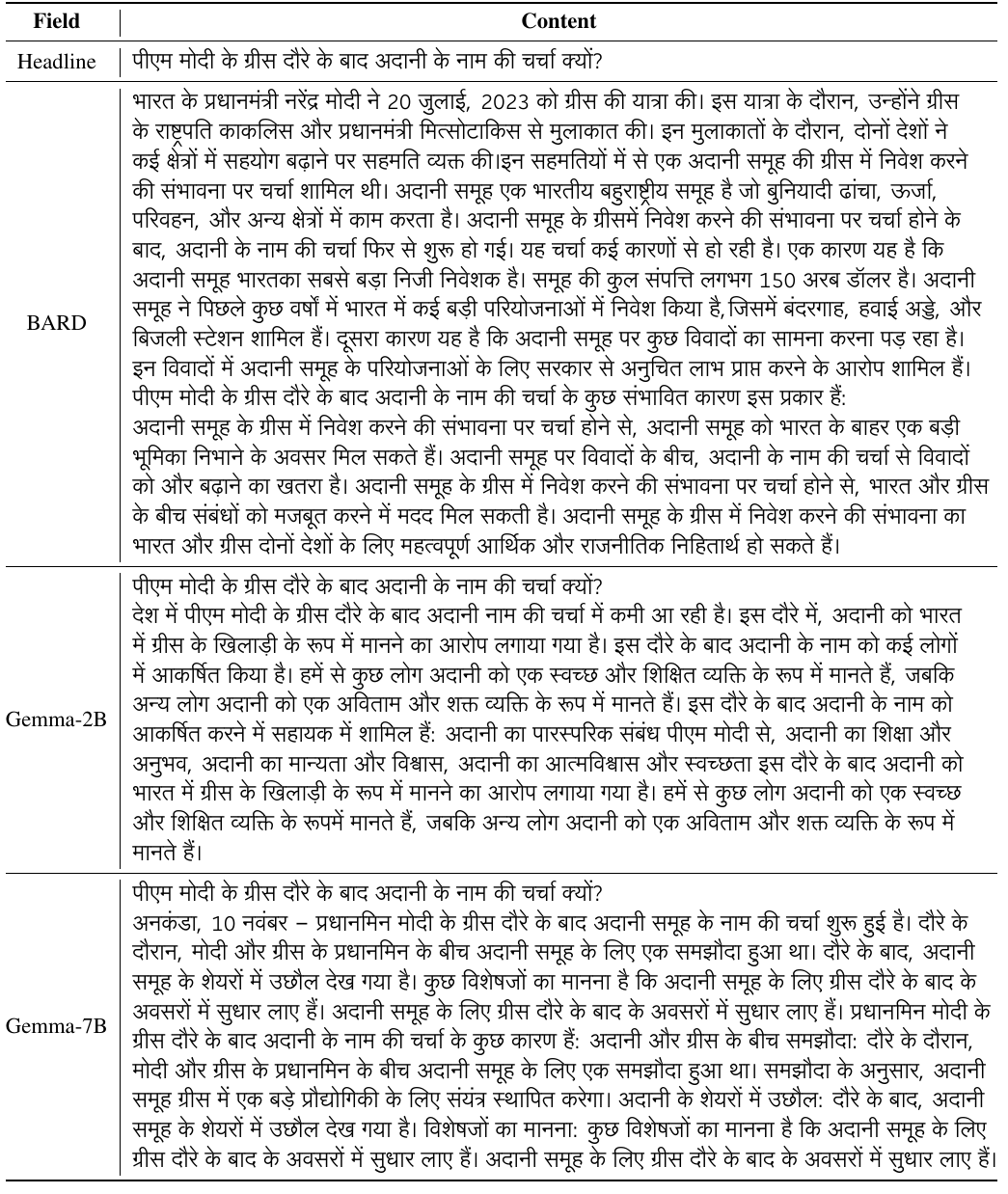}
    
    \caption{Samples of articles generated by accepted models (GPT-4, GPT-3.5, BARD, Gemma-2B and Gemma-7B).}
    \label{fig:Hindi_examples_3}
\end{figure*}

\begin{figure*}[!h]
    \centering
    \small
    \includegraphics[width=1\textwidth]{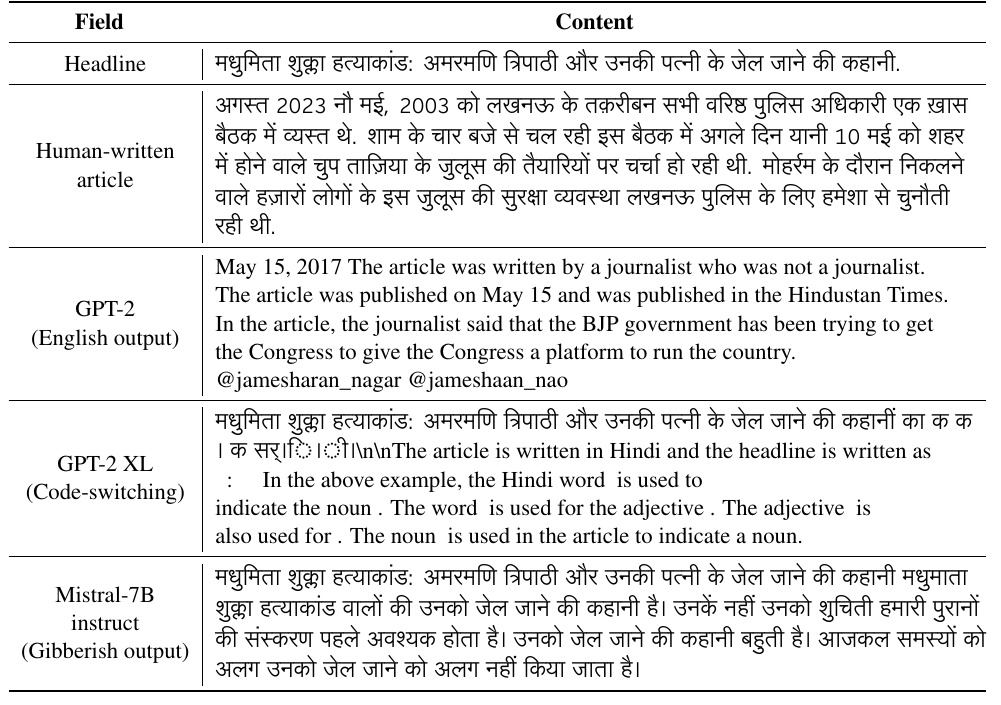}
    
    \caption{Examples illustrating outputs corresponding to different rejection criteria. We show outputs from GPT-2, GPT-2 XL and Mistral-7B instruct.}
    \label{fig:Hindi_examples_4}
\end{figure*}

We present articles generated by both accepted and dismissed models in Fig \ref{fig:Hindi_examples_2}, \ref{fig:Hindi_examples_2}, \ref{fig:Hindi_examples_3}  and Fig \ref{fig:Hindi_examples_4}, showcasing various types of rejection criteria along with specific examples.

\subsection{Prompts for Data Generation}\label{sec:prompts}
For better responses we add an instruction along with the headlines while prompting the LLMs. We experimented with various prompts for generating news articles in Hindi. Some examples include:
\begin{enumerate}[leftmargin=25pt]
    \item Expand this headline into a Hindi news article.
    \item Write a Hindi news article for the headline.
    \item Consider the given headline and write a news article for it in Hindi.
    \item Generate a Hindi news article from the given headline.
\end{enumerate}

\noindent We also experiment with Hindi instructions but observe that while larger models like GPT-4, GPT3.5 and BARD were able to generate the desired responses, other models either failed to generate any response or produced responses falling mainly in the rejection criteria mentioned in Section \ref{sec:criteria}. The use of Hindi instructions increased the number of responses falling under the aforementioned rejection criteria, therefore, we chose to use English instructions exclusively for our experiments. 


\noindent Although GPT-4, GPT-3.5 and BARD were able to generate all responses in Hindi, Gemma models could not. Hence, we only consider Hindi responses from these models in our experiments. Table \ref{tab:data_stat} summarizes the statistics for each model's responses included in $AG_{hi}$ dataset. 

\begin{table*}[h]
\centering
\small
\begin{tabular}{ccc}
\cline{2-3}
\textbf{} & \multicolumn{2}{c}{\textbf{Data source}} \\ \hline
\textbf{Model} & \textbf{BBC} & \textbf{NDTV} \\ \hline
GPT-4 & 1762 & 5280 \\
GPT-3.5 & 1762 & 5280 \\
BARD & 1762 & 5280 \\
Gemma-2B & 468 & 1715 \\
Gemma-7B & 1636 & 4679 \\ \hline
Total & 7390 & 22234 \\ \hline
\end{tabular}
\caption{Data samples statistics for individual models. The combined BBC and NDTV datasets contain a total of 29,624 AI-generated data points, providing a substantial basis for evaluating the performance and generalization capabilities of the models. }
\label{tab:data_stat}
\end{table*}

\subsection{Hyperparameters for models}\label{sec:hyperparameters}
We list the hyperparameters employed during text generation for the included models. Various hyperparameters were tested to evaluate the rejected models, but their outcomes did not meet our criteria, resulting in exclusion from further consideration. Table \ref{tab:Hyperparameters} provides a comprehensive overview of all the hyperparameters for the models.

\begin{table}[!h]
\small
\centering
\begin{tabular}{cc}
\hline
\textbf{Model} & \textbf{Hyperparameters} \\ \hline
GPT-4 & \begin{tabular}[c]{@{}c@{}}temperature: 1\\ max\_tokens: 500\\ frequency\_penalty: 0\end{tabular} \\
GPT-3.5 Turbo & \begin{tabular}[c]{@{}c@{}}temperature: 1\\ max\_tokens: 500\\ frequency\_penalty: 0\end{tabular} \\
BARD & - \\
Gemma-2B & \begin{tabular}[c]{@{}c@{}}temperature: 1\\ max\_tokens: 500\end{tabular} \\
Gemma-7B & \begin{tabular}[c]{@{}c@{}}temperature: 1\\ max\_tokens: 500\end{tabular} \\ \hline
\end{tabular}
\caption{Hyperparameters used to generate text from different models. No hyperparameters are available for BARD as the data was collected directly from the website.}
\label{tab:Hyperparameters}
\end{table}

\section{Results}\label{sec:app_results}
In this section, we discuss additional results from the aforementioned AI-Generated Text Detection techniques. 

\subsection{Main Results}
\begin{figure*}[h]
    \centering
    \begin{subfigure}[b]{0.19\textwidth}
        \centering
        \includegraphics[width=\textwidth]{img/RAIDAR_CF_Final/RAIDAR_BBC_GPT4.pdf}
        
        \caption{BBC GPT-4}
        \label{fig:perp_gpt4}
    \end{subfigure}
    \hfill
    \begin{subfigure}[b]{0.19\textwidth}
        \centering
        \includegraphics[width=\textwidth]{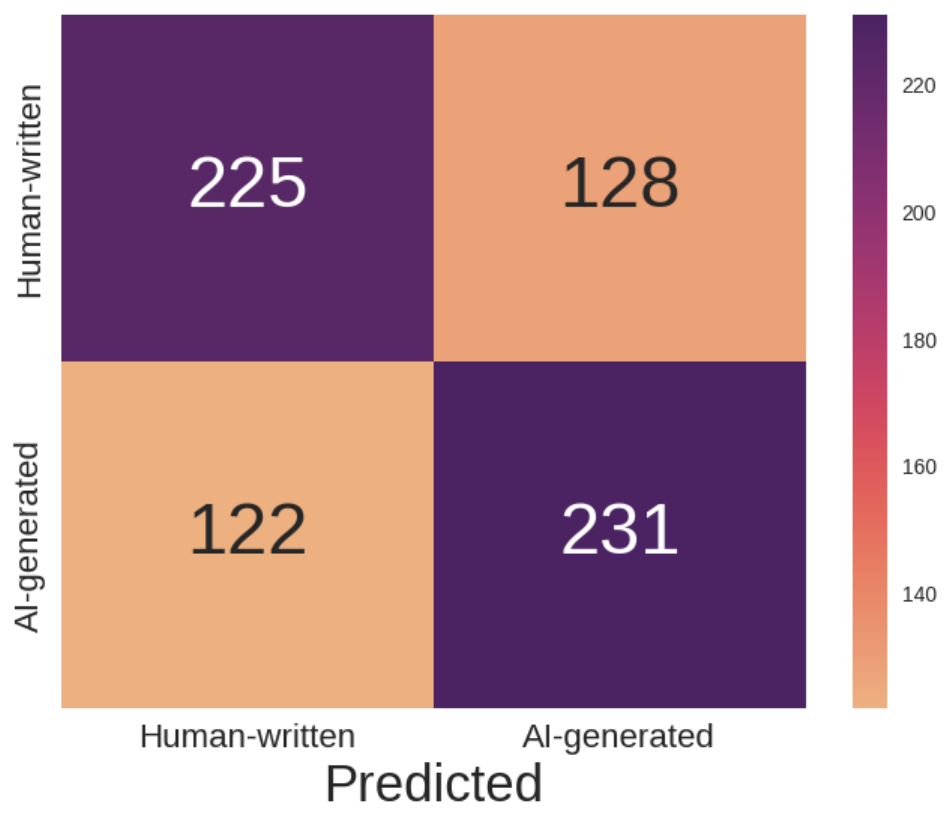}
        
        \caption{BBC GPT-3.5}
        \label{fig:perp_gpt35}
    \end{subfigure}
    \hfill
    \begin{subfigure}[b]{0.19\textwidth}
        \centering
        \includegraphics[width=\textwidth]{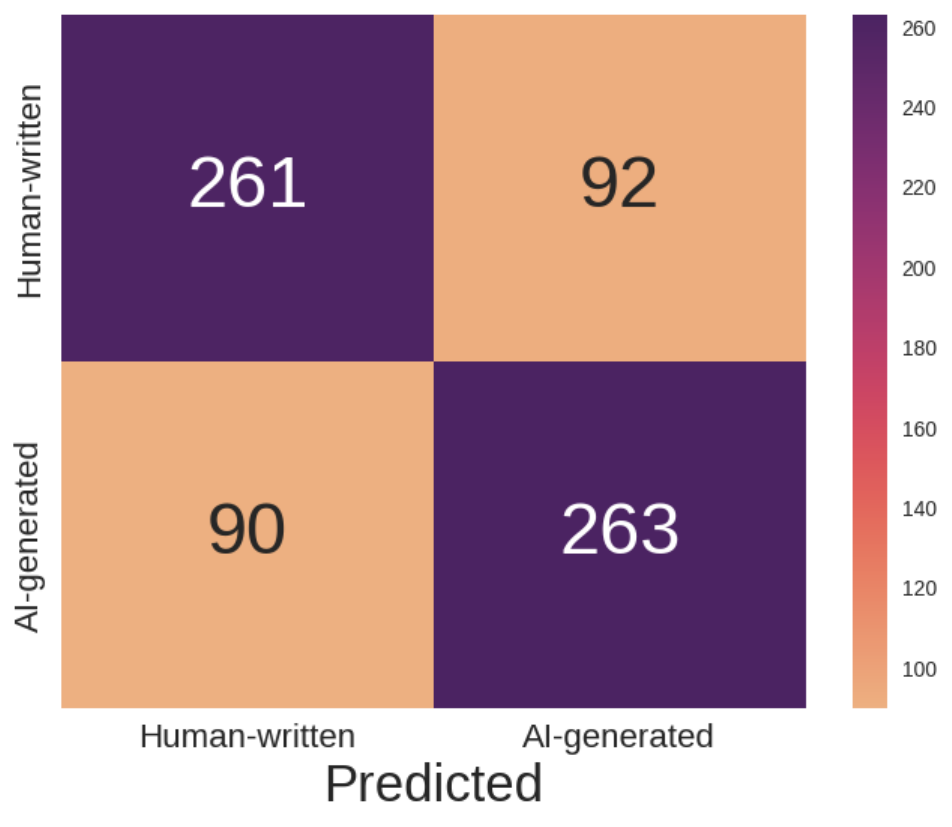}
        \caption{BBC BARD}
        \label{fig:perp_bard}
    \end{subfigure}
    \hfill
    \begin{subfigure}[b]{0.19\textwidth}
        \centering
        \includegraphics[width=\textwidth]{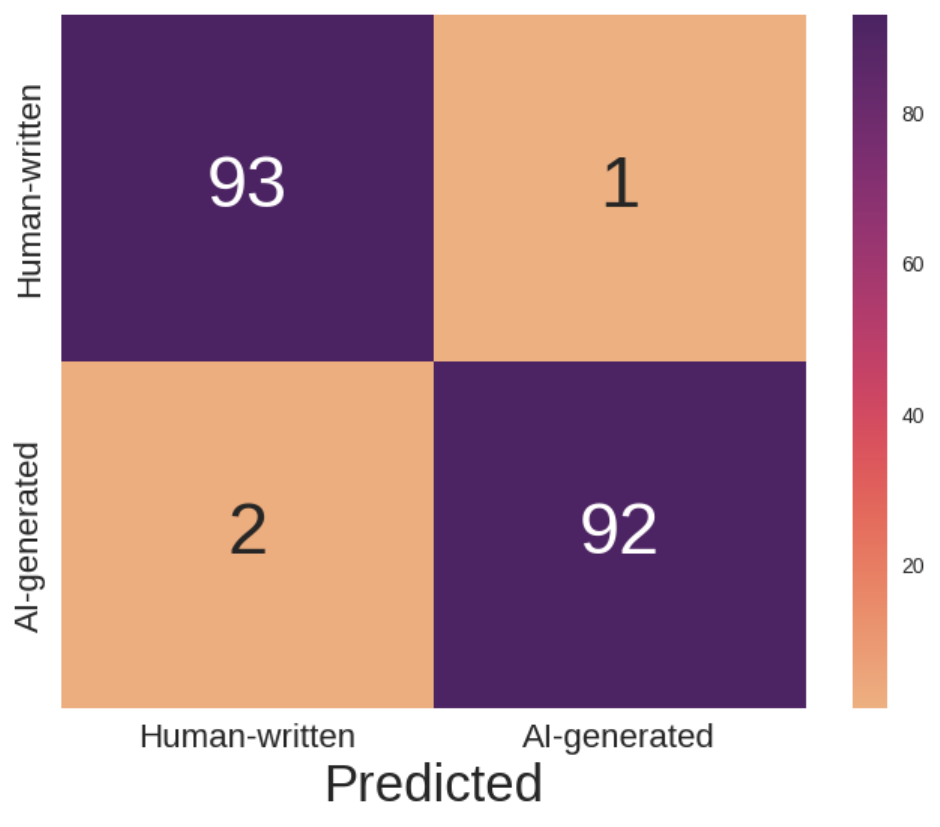}
        \caption{BBC Gemma-2B}
        \label{fig:perp_gemma2b}
    \end{subfigure}
    \hfill
    \begin{subfigure}[b]{0.19\textwidth}
        \centering
        \includegraphics[width=\textwidth]{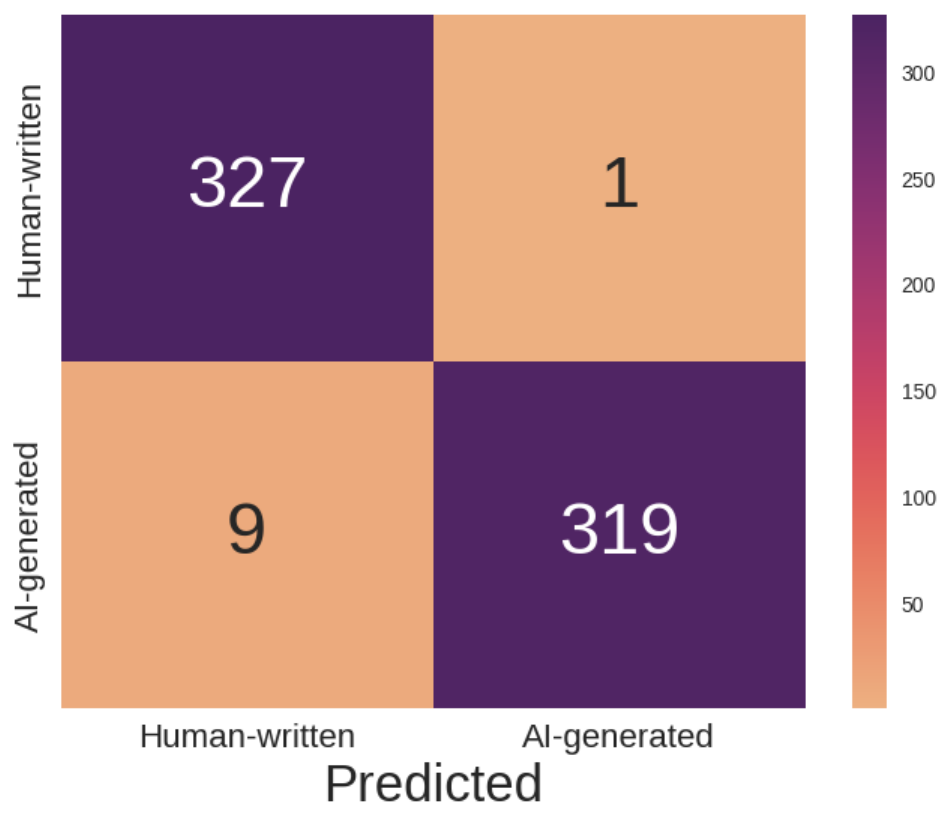}
        \caption{BBC Gemma-7B}
        \label{fig:perp_gemma7b}
    \end{subfigure}
    \\
    \begin{subfigure}[b]{0.19\textwidth}
        \centering
        \includegraphics[width=\textwidth]{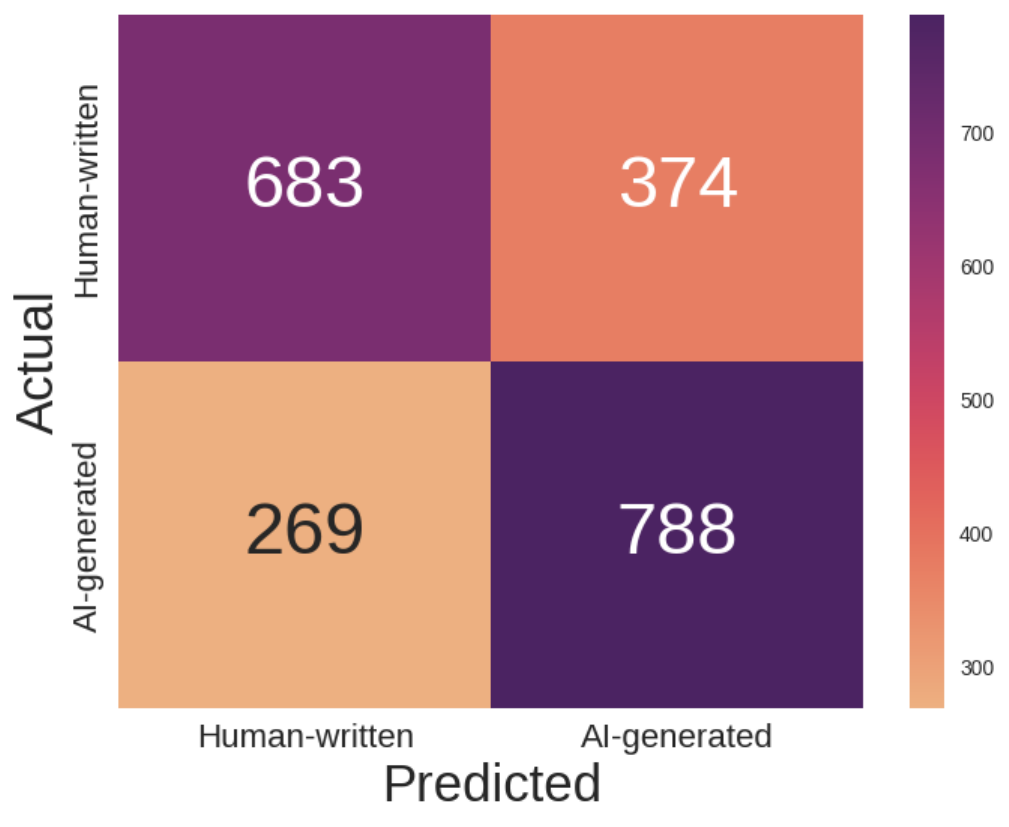}
        \caption{NDTV GPT-4}
        \label{fig:perp_model1}
    \end{subfigure}
    \hfill
    \begin{subfigure}[b]{0.19\textwidth}
        \centering
        \includegraphics[width=\textwidth]{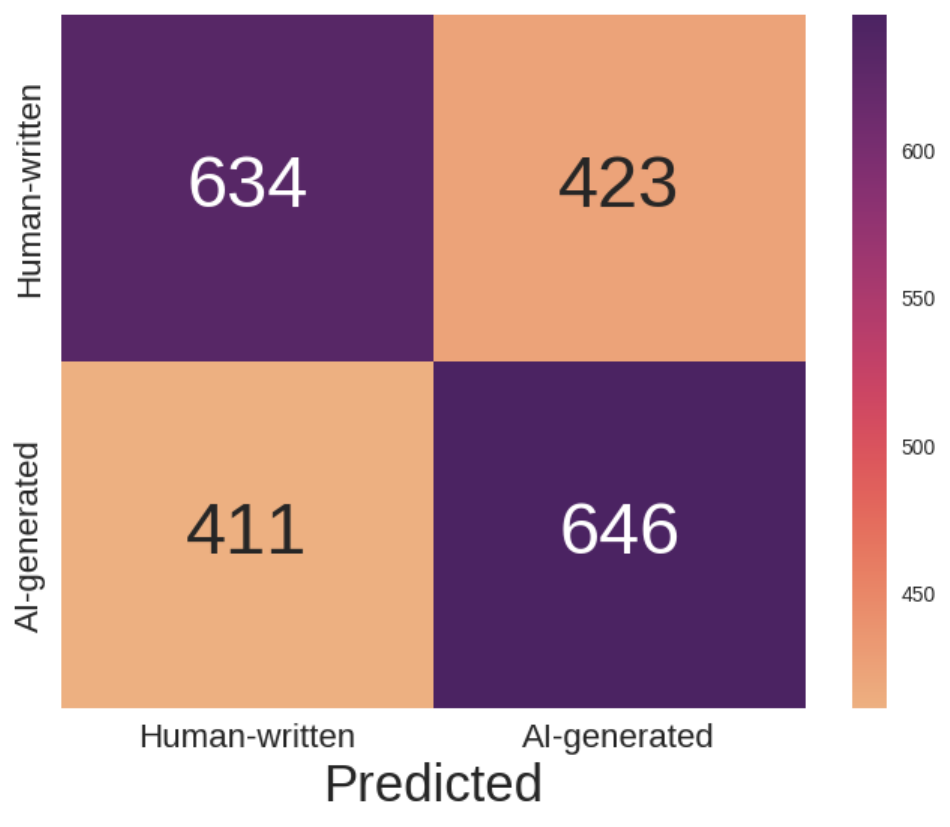}
        \caption{NDTV GPT-3.5}
        \label{fig:perp_model2}
    \end{subfigure}
    \hfill
    \begin{subfigure}[b]{0.19\textwidth}
        \centering
        \includegraphics[width=\textwidth]{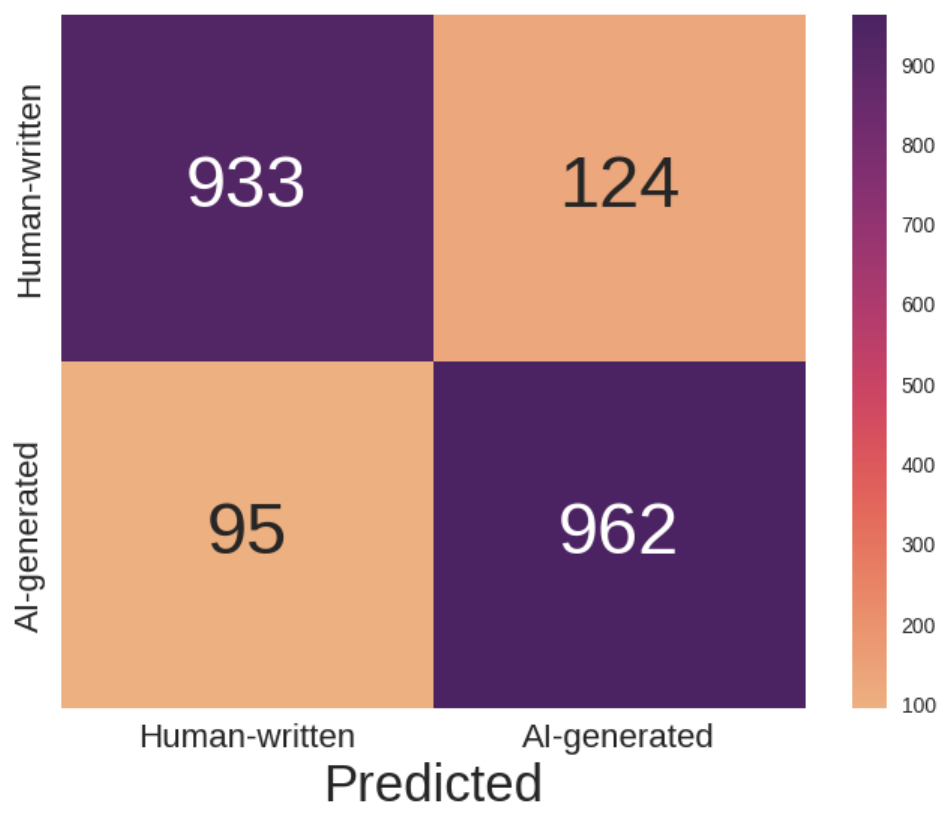}
        \caption{NDTV BARD}
        \label{fig:perp_model3}
    \end{subfigure}
    \hfill
    \begin{subfigure}[b]{0.20\textwidth}
        \centering
        \includegraphics[width=\textwidth]{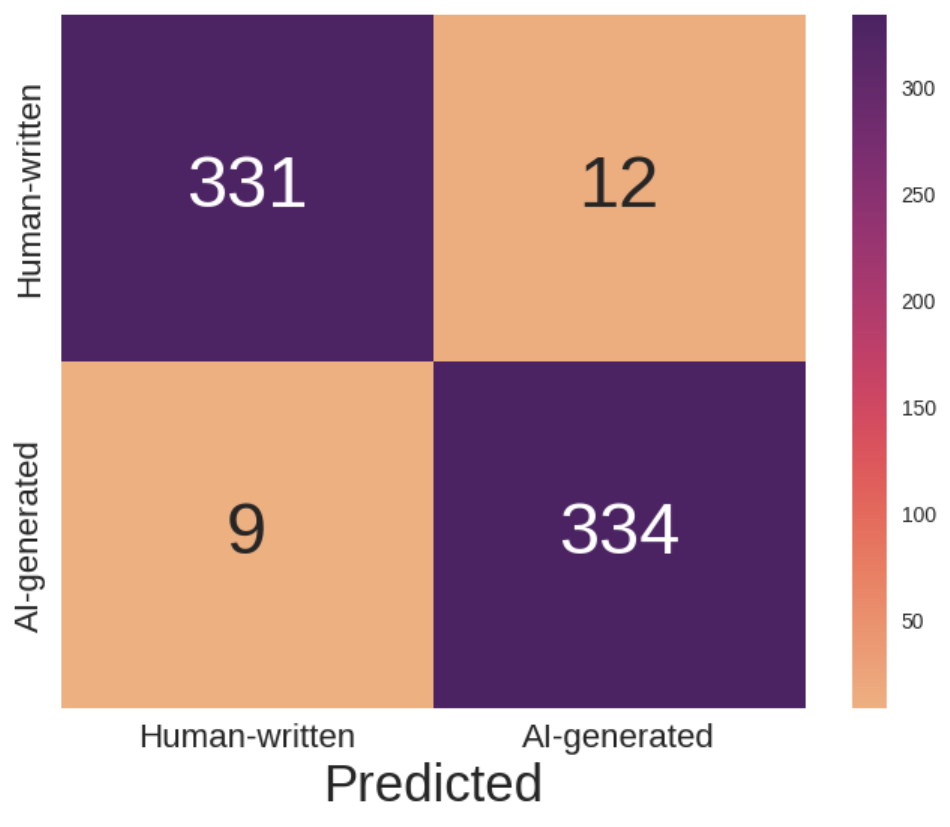}
        \caption{NDTV Gemma-2B}
        \label{fig:perp_model4}
    \end{subfigure}
    \hfill
    \begin{subfigure}[b]{0.20\textwidth}
        \centering
        \includegraphics[width=\textwidth]{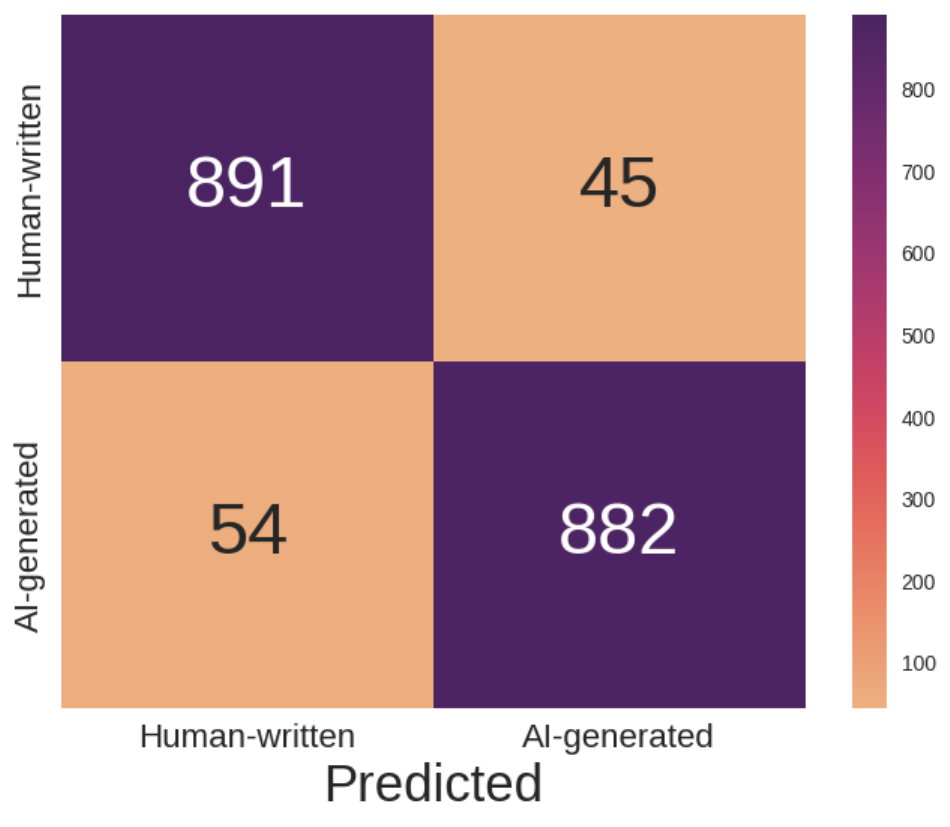}
        \caption{NDTV Gemma-7B}
        \label{fig:perp_model5}
    \end{subfigure}
    \caption{RAIDAR Confusion Matrices. GPT models exhibit higher misclassification rates compared to other models. Similar numbers of false positives (FP) and false negatives (FN) across models and datasets indicate a trade-off-aware approach of the model.}
    \label{fig:raidar_cf}
\end{figure*}

\begin{figure*}[h]
    \centering
    \begin{subfigure}[b]{0.19\textwidth}
        \centering
        \includegraphics[width=\textwidth]{img/RADAR_CF_Final/RADAR_BBC_GPT4.pdf}
        \caption{BBC GPT-4}
        \label{fig:perp_gpt4}
    \end{subfigure}
    \hfill
    \begin{subfigure}[b]{0.19\textwidth}
        \centering
        \includegraphics[width=\textwidth]{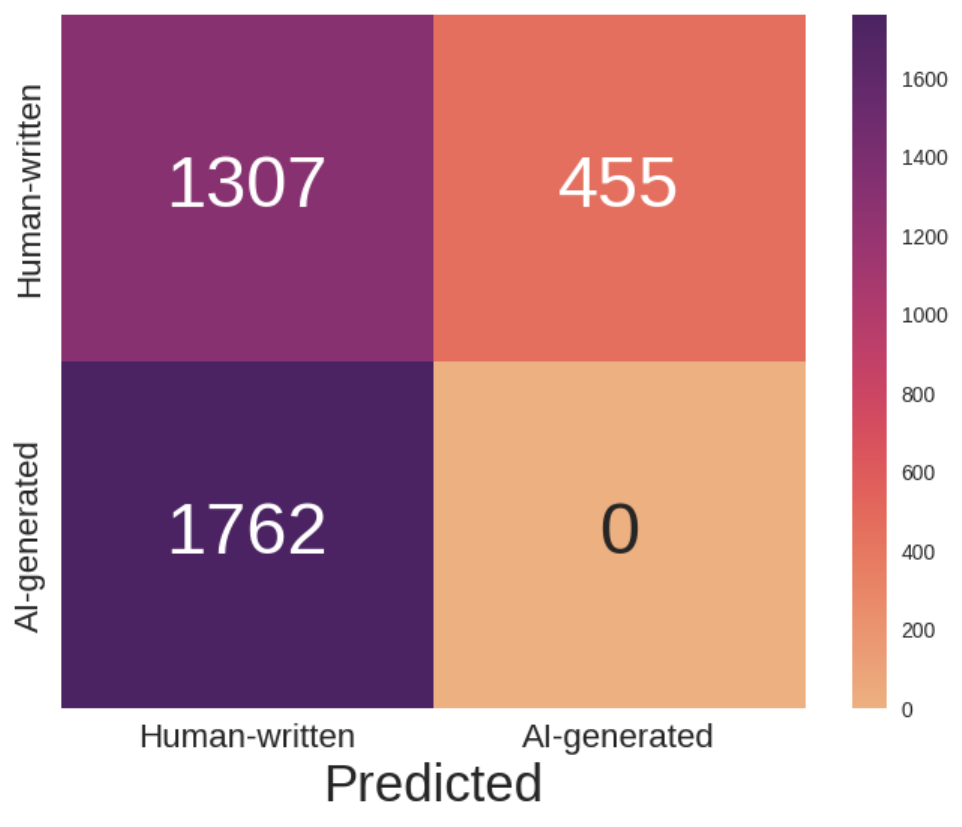}
        \caption{BBC GPT-3.5}
        \label{fig:perp_gpt35}
    \end{subfigure}
    \hfill
    \begin{subfigure}[b]{0.19\textwidth}
        \centering
        \includegraphics[width=\textwidth]{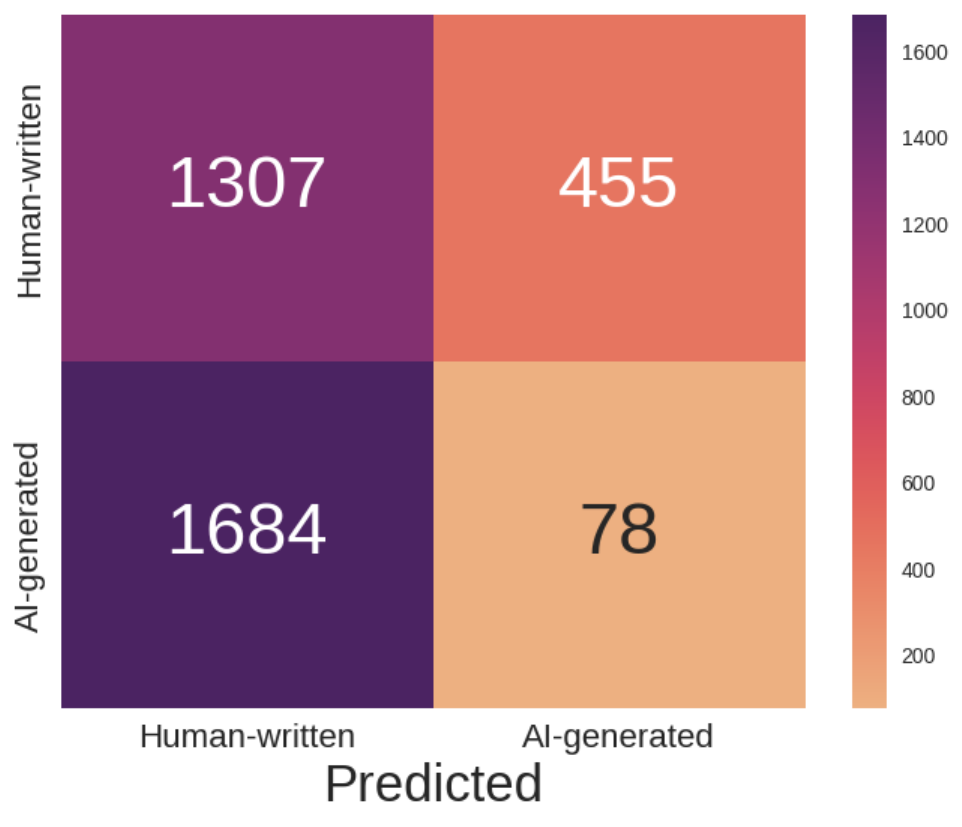}
        \caption{BBC BARD}
        \label{fig:perp_bard}
    \end{subfigure}
    \hfill
    \begin{subfigure}[b]{0.19\textwidth}
        \centering
        \includegraphics[width=\textwidth]{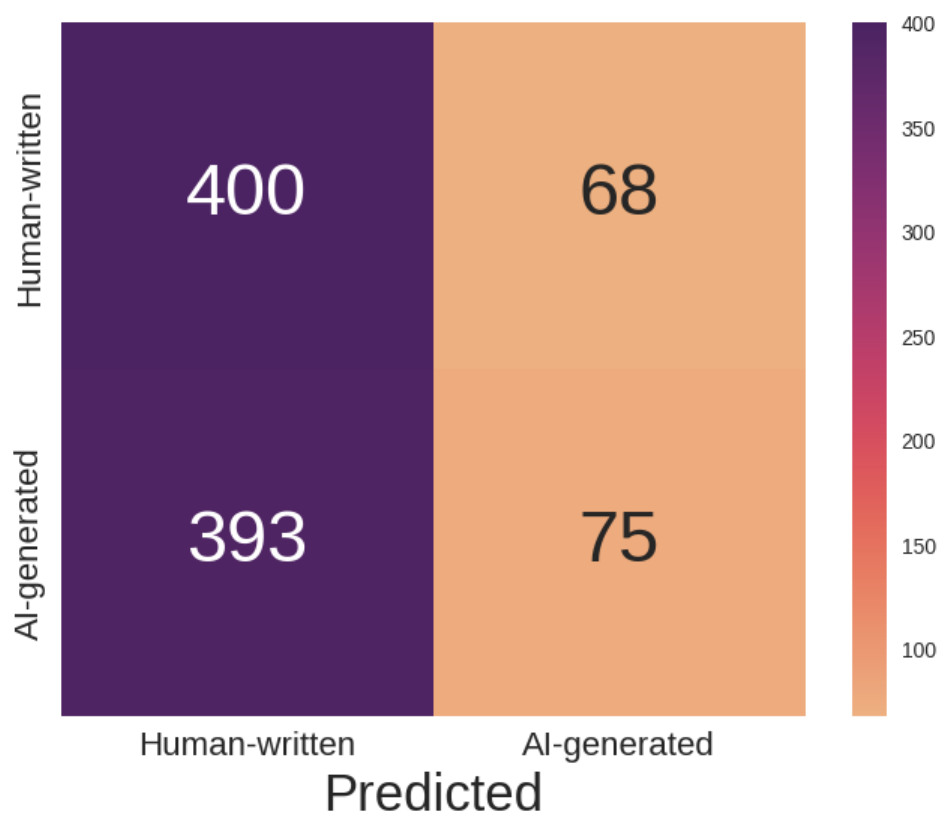}
        \caption{BBC Gemma-2B}
        \label{fig:perp_gemma2b}
    \end{subfigure}
    \hfill
    \begin{subfigure}[b]{0.19\textwidth}
        \centering
        \includegraphics[width=\textwidth]{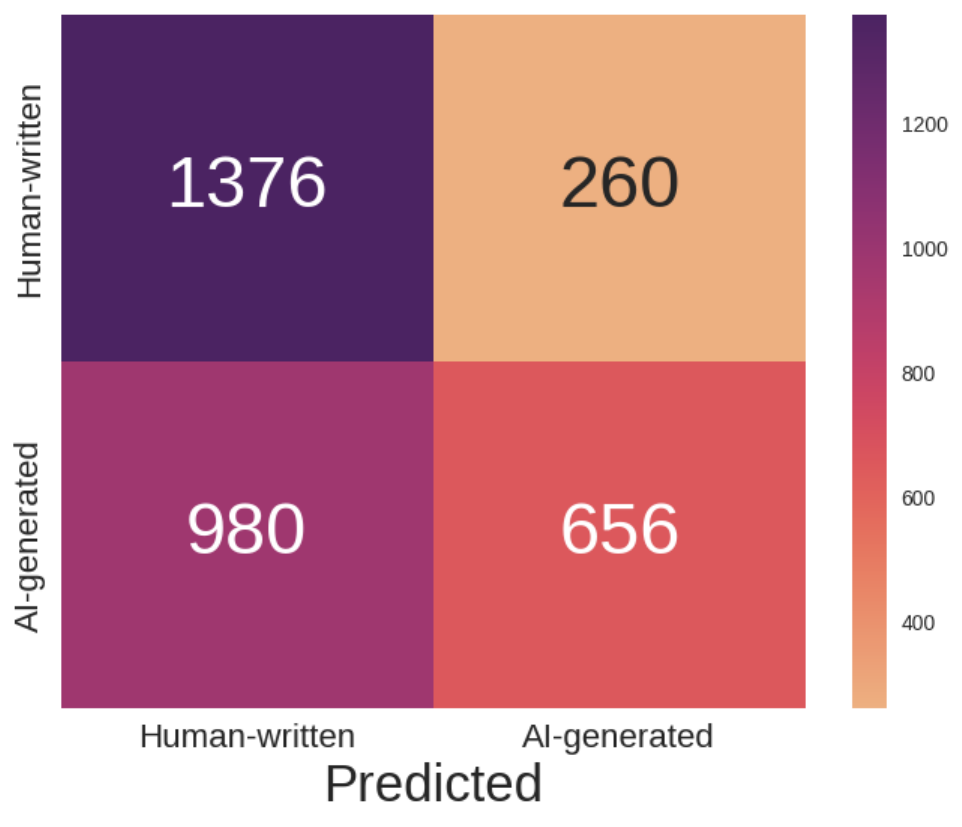}
        \caption{BBC Gemma-7B}
        \label{fig:perp_gemma7b}
    \end{subfigure}
    \\
    \begin{subfigure}[b]{0.19\textwidth}
        \centering
        \includegraphics[width=\textwidth]{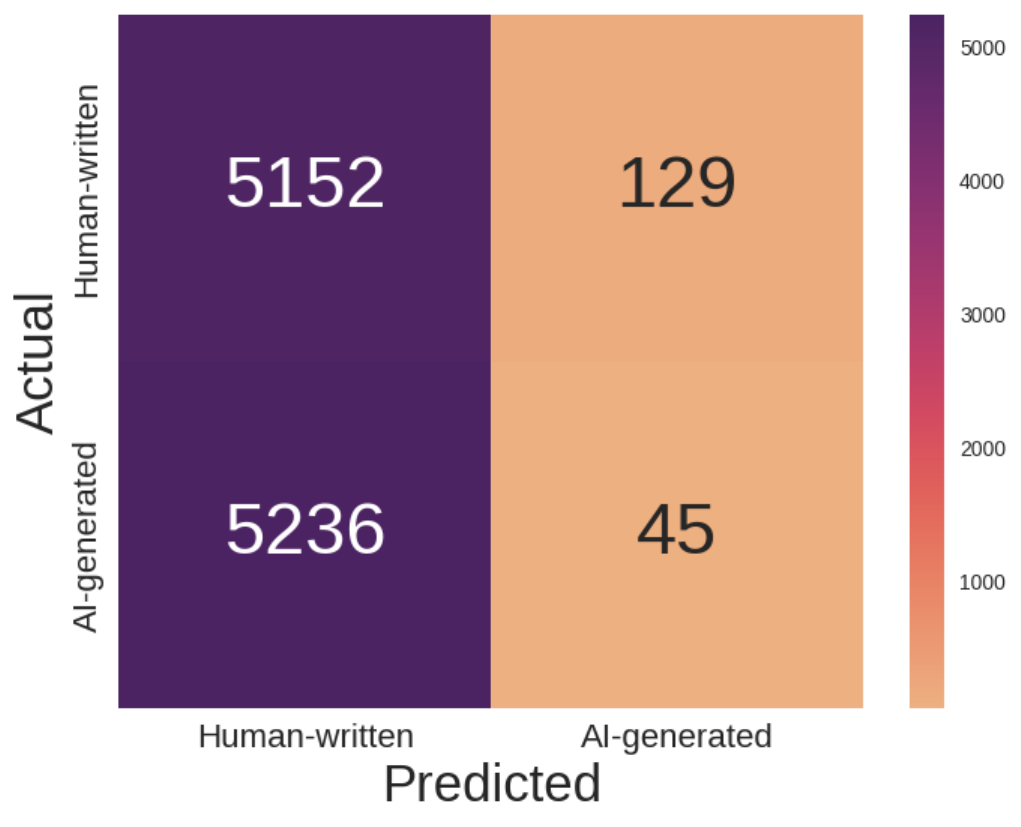}
        \caption{NDTV GPT-4}
        \label{fig:perp_model1}
    \end{subfigure}
    \hfill
    \begin{subfigure}[b]{0.19\textwidth}
        \centering
        \includegraphics[width=\textwidth]{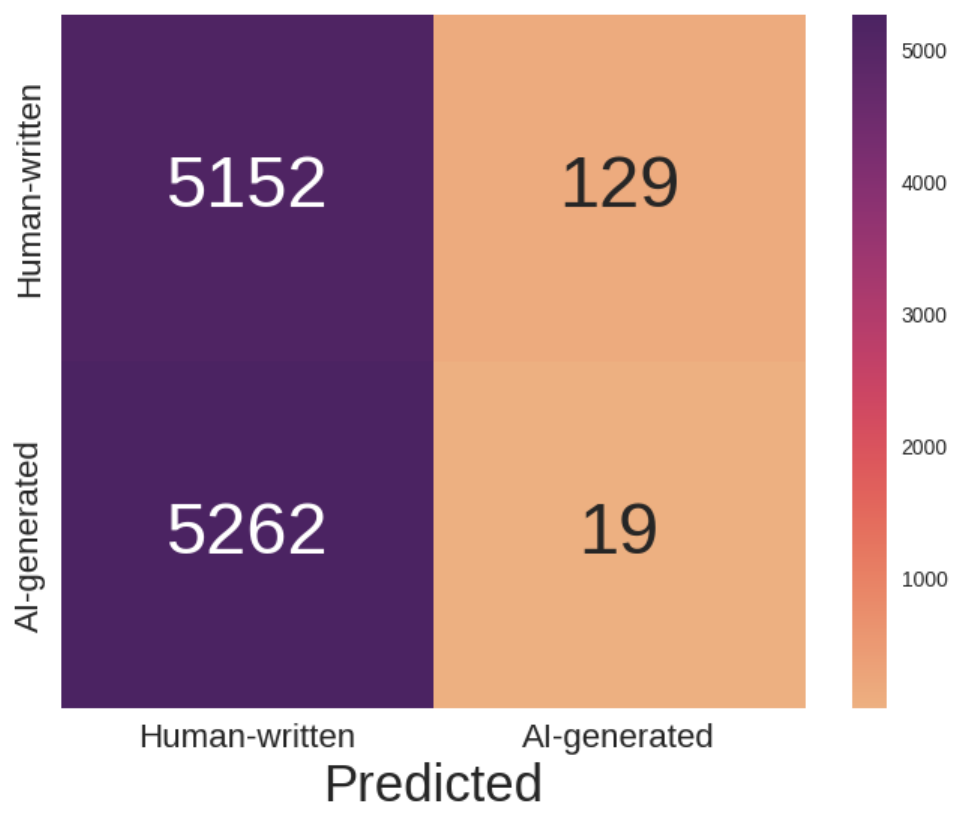}
        \caption{NDTV GPT-3.5}
        \label{fig:perp_model2}
    \end{subfigure}
    \hfill
    \begin{subfigure}[b]{0.19\textwidth}
        \centering
        \includegraphics[width=\textwidth]{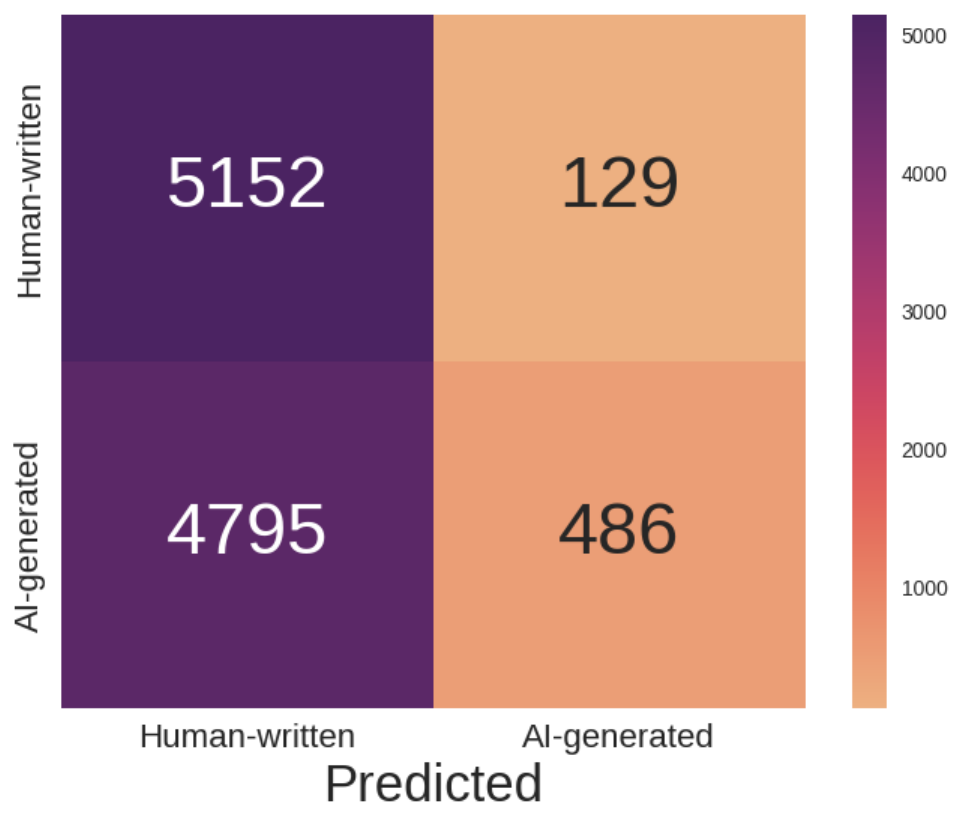}
        \caption{NDTV BARD}
        \label{fig:perp_model3}
    \end{subfigure}
    \hfill
    \begin{subfigure}[b]{0.20\textwidth}
        \centering
        \includegraphics[width=\textwidth]{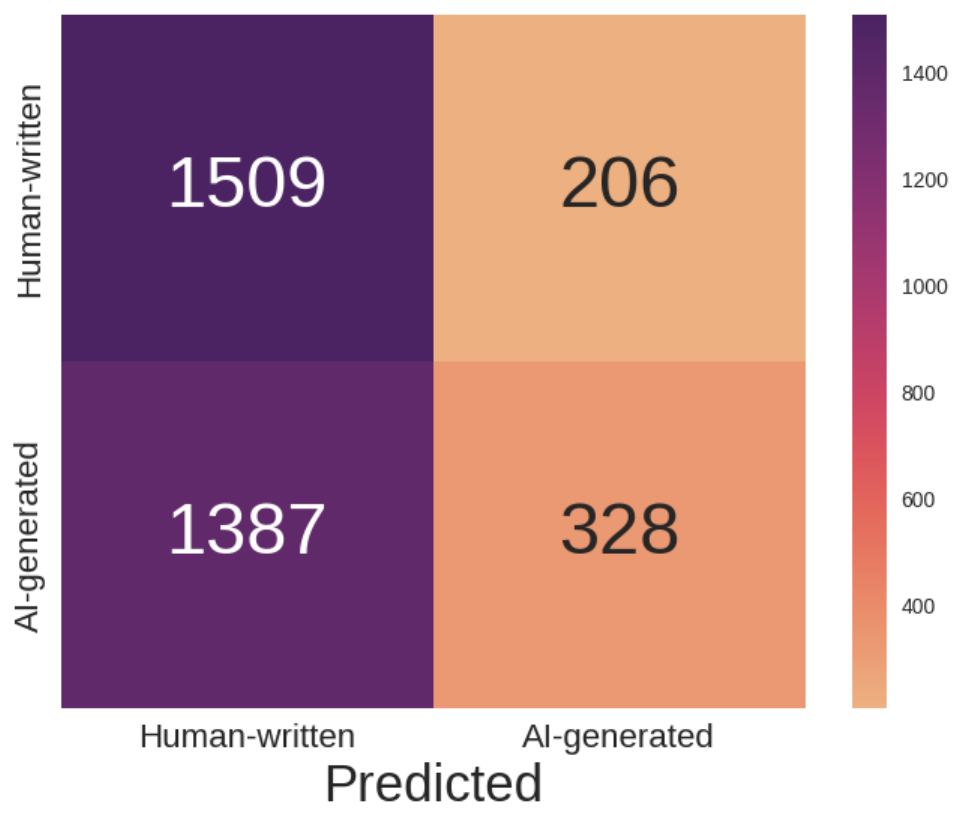}
        \caption{NDTV Gemma-2B}
        \label{fig:perp_model4}
    \end{subfigure}
    \hfill
    \begin{subfigure}[b]{0.20\textwidth}
        \centering
        \includegraphics[width=\textwidth]{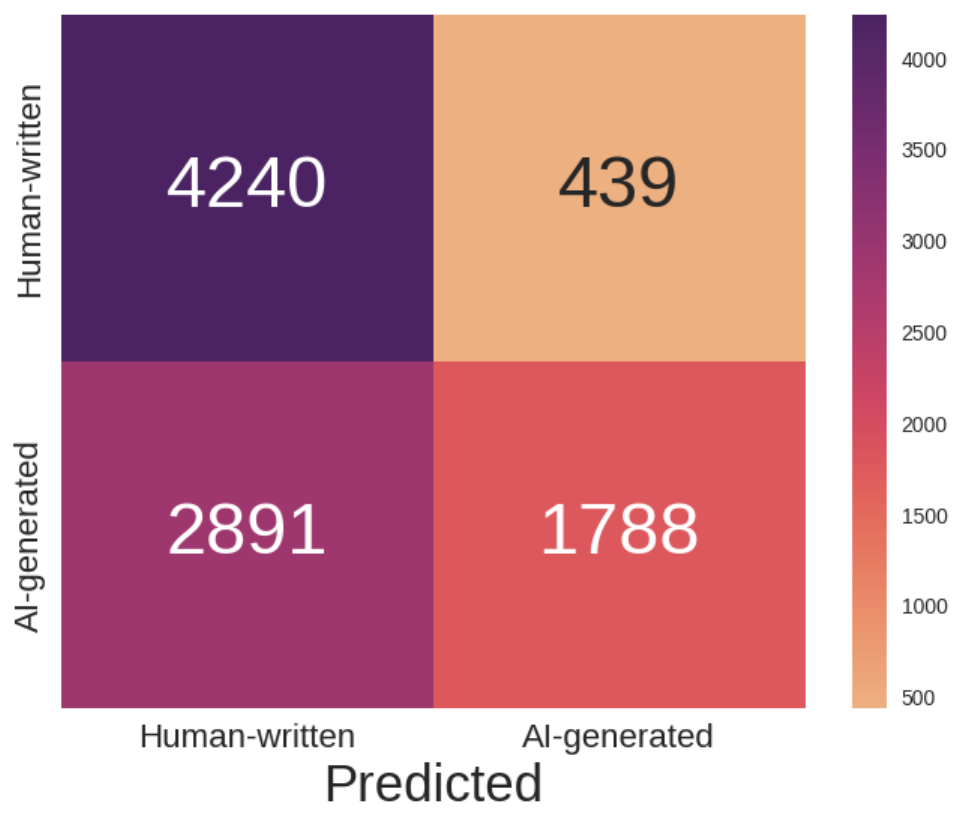}
        \caption{NDTV Gemma-7B}
        \label{fig:perp_model5}
    \end{subfigure}
    \caption{RADAR Confusion Matrices. A significant disparity can be observed in predictions between AI-generated and human-written classes, with AI-generated classes being predicted much less frequently. This suggests that the model exhibits a bias towards identifying text as human-written rather than AI-generated, reflecting a potential challenge in accurately distinguishing between the two classes. }
    \label{fig:radar_cf}
\end{figure*}

\begin{figure*}[h]
    \centering
    \begin{subfigure}[b]{0.19\textwidth}
        \centering
        \includegraphics[width=\textwidth]{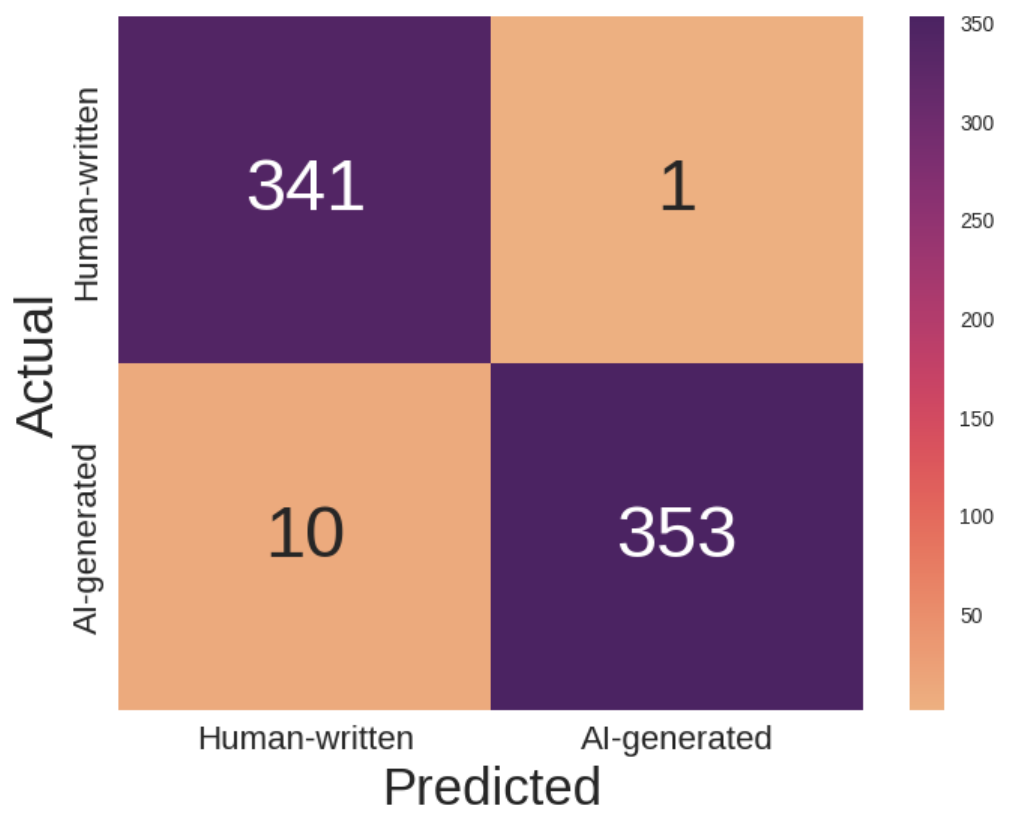}
        \caption{BBC GPT-4}
        \label{fig:perp_gpt4}
    \end{subfigure}
    \hfill
    \begin{subfigure}[b]{0.19\textwidth}
        \centering
        \includegraphics[width=\textwidth]{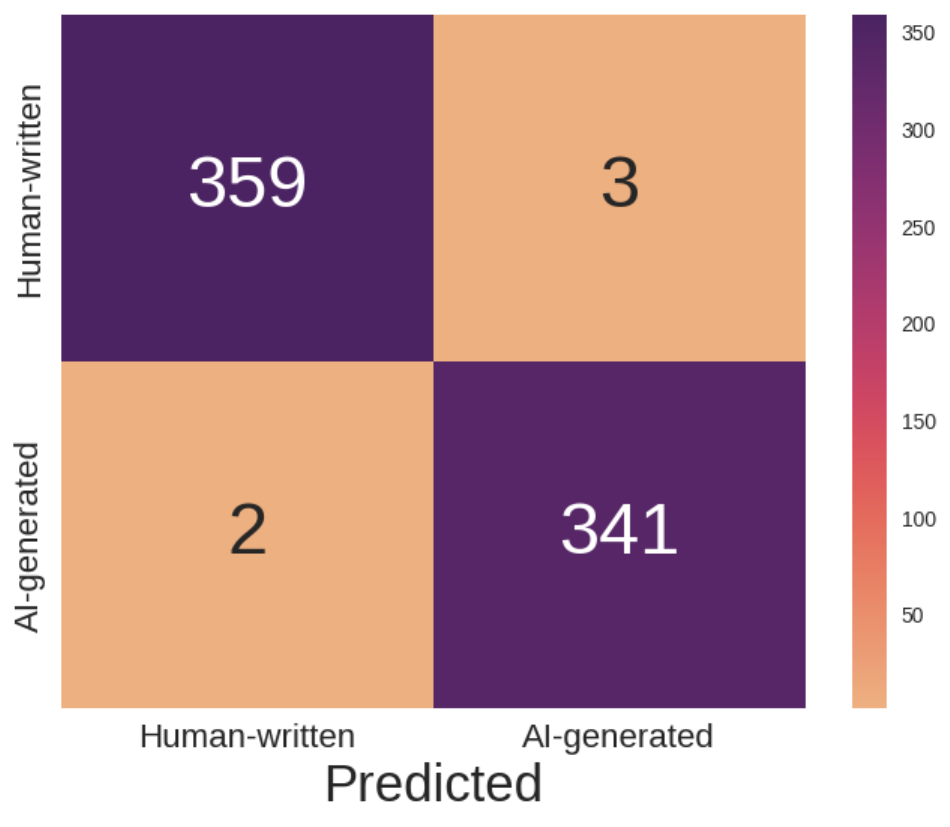}
        \caption{BBC GPT-3.5}
        \label{fig:perp_gpt35}
    \end{subfigure}
    \hfill
    \begin{subfigure}[b]{0.19\textwidth}
        \centering
        \includegraphics[width=\textwidth]{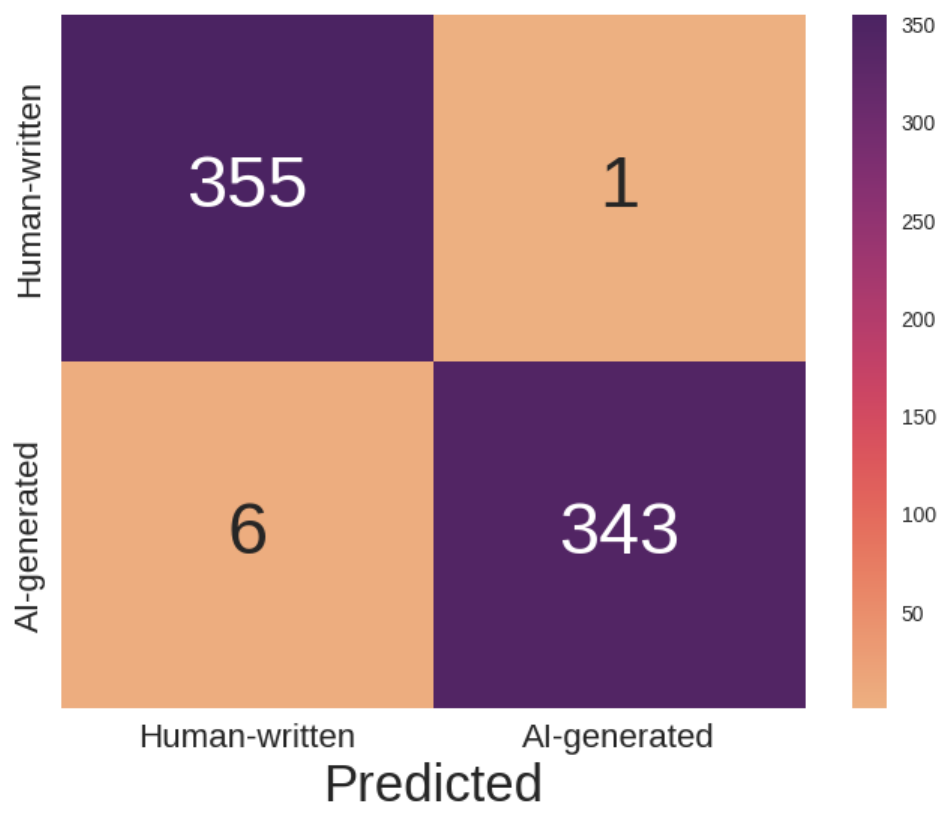}
        \caption{BBC BARD}
        \label{fig:perp_bard}
    \end{subfigure}
    \hfill
    \begin{subfigure}[b]{0.19\textwidth}
        \centering
        \includegraphics[width=\textwidth]{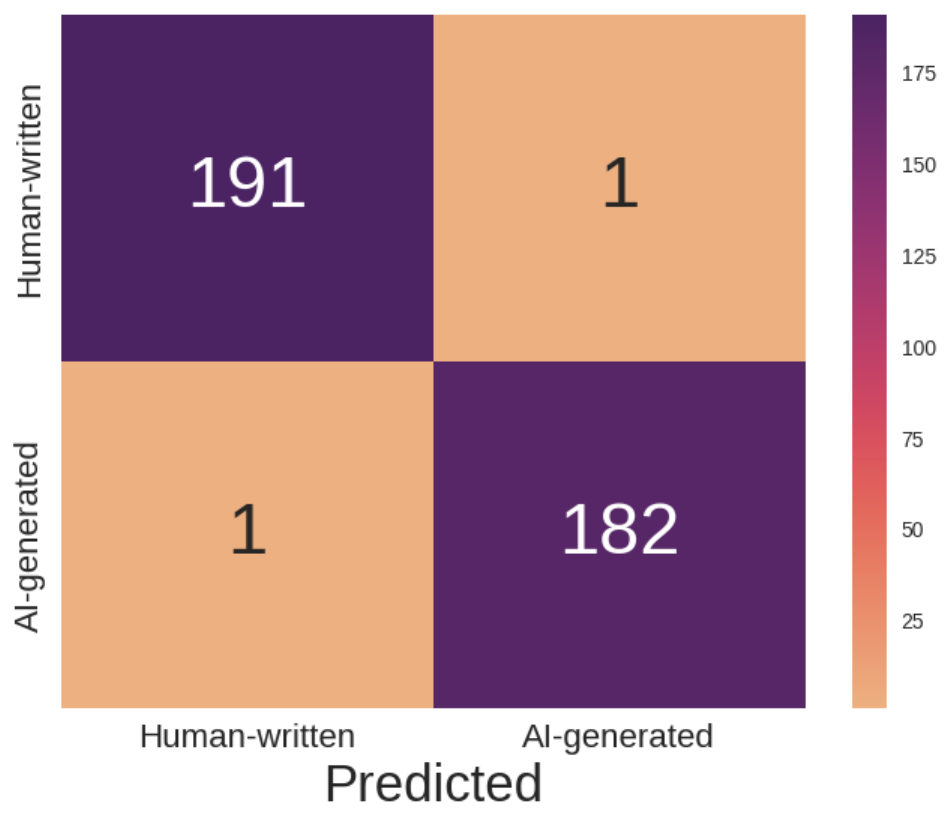}
        \caption{BBC Gemma-2B}
        \label{fig:perp_gemma2b}
    \end{subfigure}
    \hfill
    \begin{subfigure}[b]{0.19\textwidth}
        \centering
        \includegraphics[width=\textwidth]{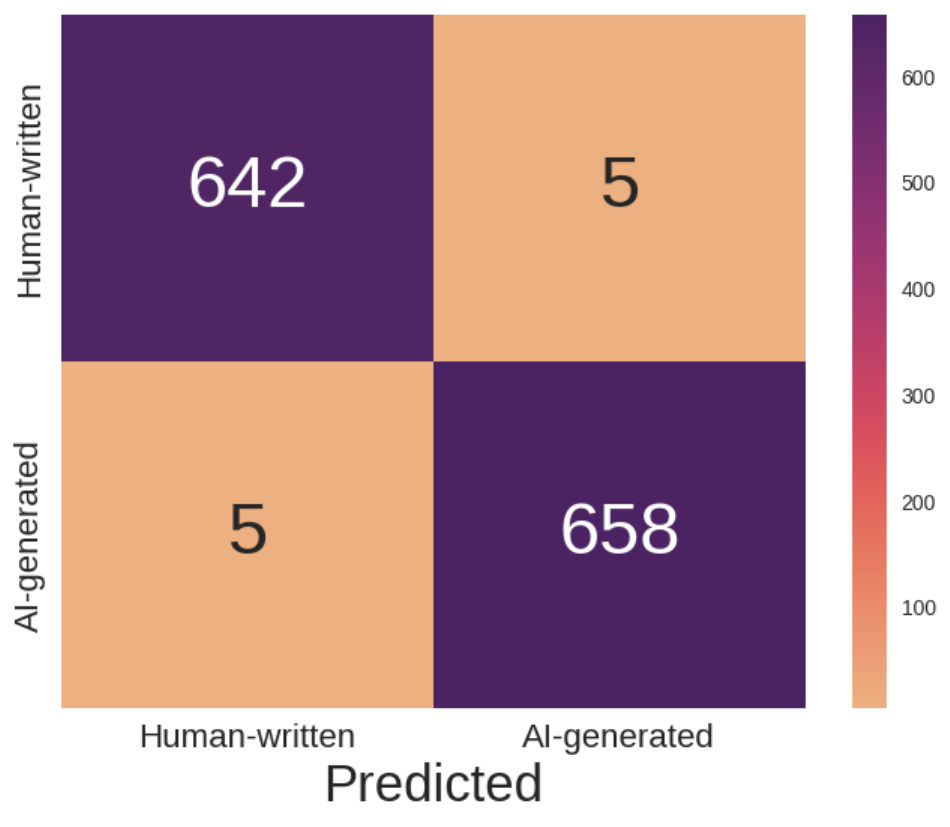}
        \caption{BBC Gemma-7B}
        \label{fig:perp_gemma7b}
    \end{subfigure}
    \\
    \begin{subfigure}[b]{0.19\textwidth}
        \centering
        \includegraphics[width=\textwidth]{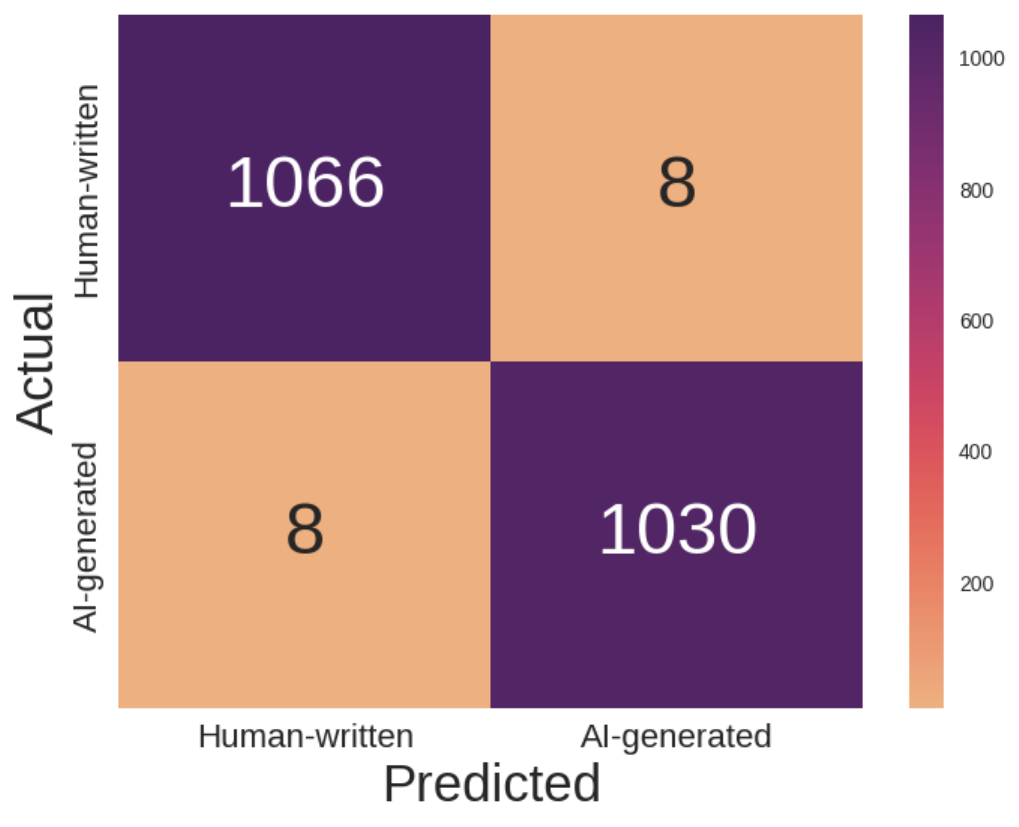}
        \caption{NDTV GPT-4}
        \label{fig:perp_model1}
    \end{subfigure}
    \hfill
    \begin{subfigure}[b]{0.19\textwidth}
        \centering
        \includegraphics[width=\textwidth]{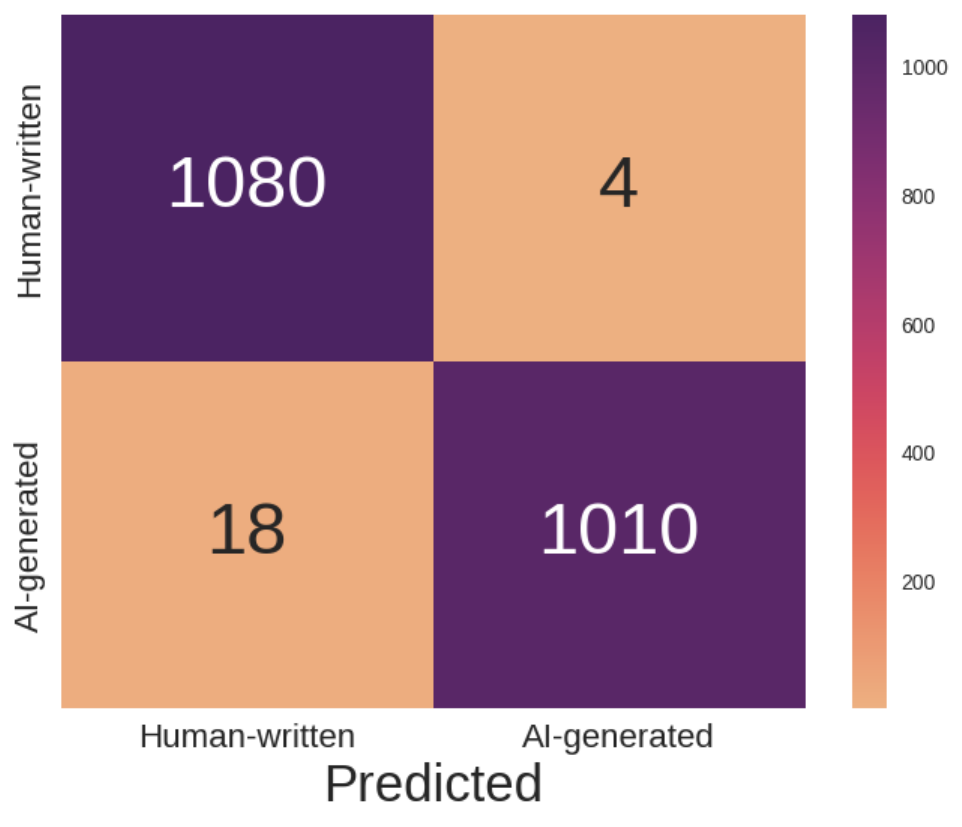}
        \caption{NDTV GPT-3.5}
        \label{fig:perp_model2}
    \end{subfigure}
    \hfill
    \begin{subfigure}[b]{0.19\textwidth}
        \centering
        \includegraphics[width=\textwidth]{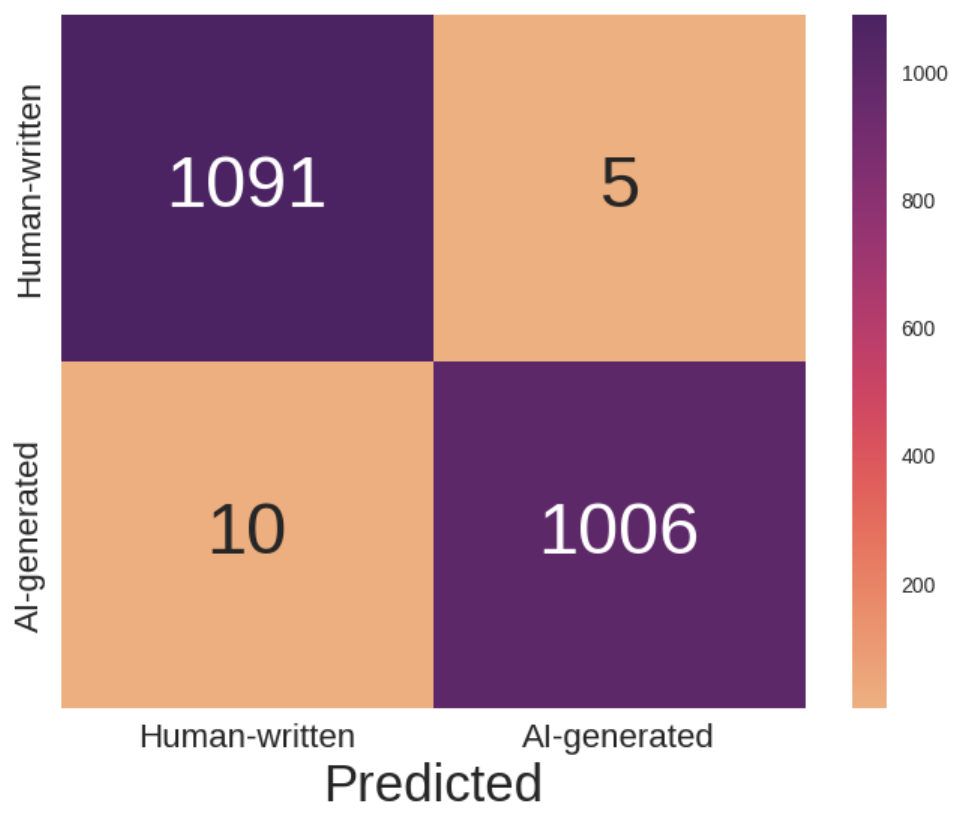}
        \caption{NDTV BARD}
        \label{fig:perp_model3}
    \end{subfigure}
    \hfill
    \begin{subfigure}[b]{0.20\textwidth}
        \centering
        \includegraphics[width=\textwidth]{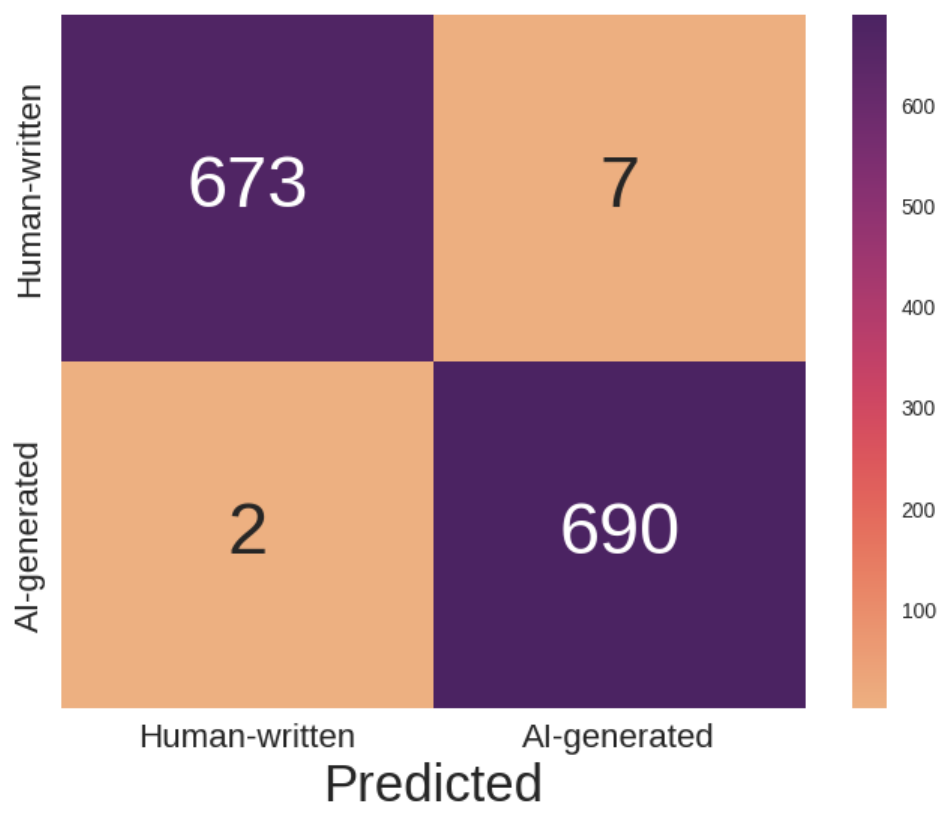}
        \caption{NDTV Gemma-2B}
        \label{fig:perp_model4}
    \end{subfigure}
    \hfill
    \begin{subfigure}[b]{0.20\textwidth}
        \centering
        \includegraphics[width=\textwidth]{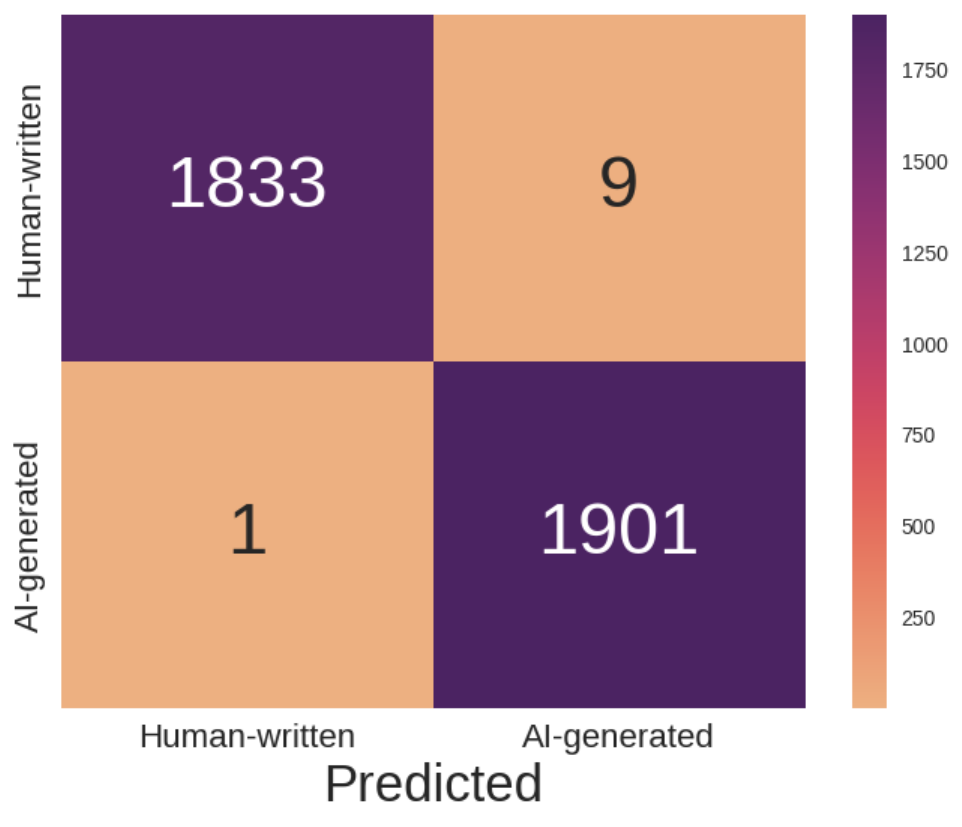}
        \caption{NDTV Gemma-7B}
        \label{fig:perp_model5}
    \end{subfigure}
    \caption{J-Guard Confusion Matrices. The models trained and tested on the same model responses are able to distinguish AI-generated text from human-written text with a few exceptions. Instances of false positives and false negatives are minimal.}
    \label{fig:j_guard_cf}
\end{figure*}

\begin{figure*}[h]
    \centering
    \begin{subfigure}[b]{0.19\textwidth}
        \centering
        \includegraphics[width=\textwidth]{img/ConDA_CF_Final/Conda_BBC_GPT4.pdf}
        \caption{BBC GPT-4}
        \label{fig:perp_gpt4}
    \end{subfigure}
    \hfill
    \begin{subfigure}[b]{0.19\textwidth}
        \centering
        \includegraphics[width=\textwidth]{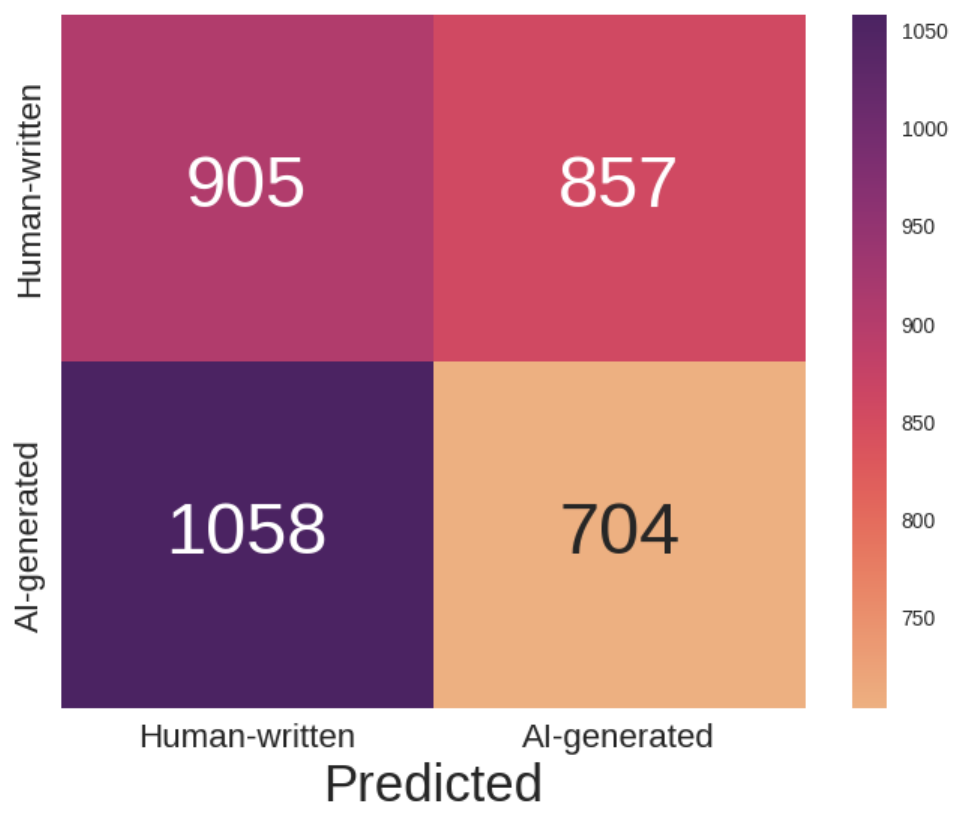}
        \caption{BBC GPT-3.5}
        \label{fig:perp_gpt35}
    \end{subfigure}
    \hfill
    \begin{subfigure}[b]{0.19\textwidth}
        \centering
        \includegraphics[width=\textwidth]{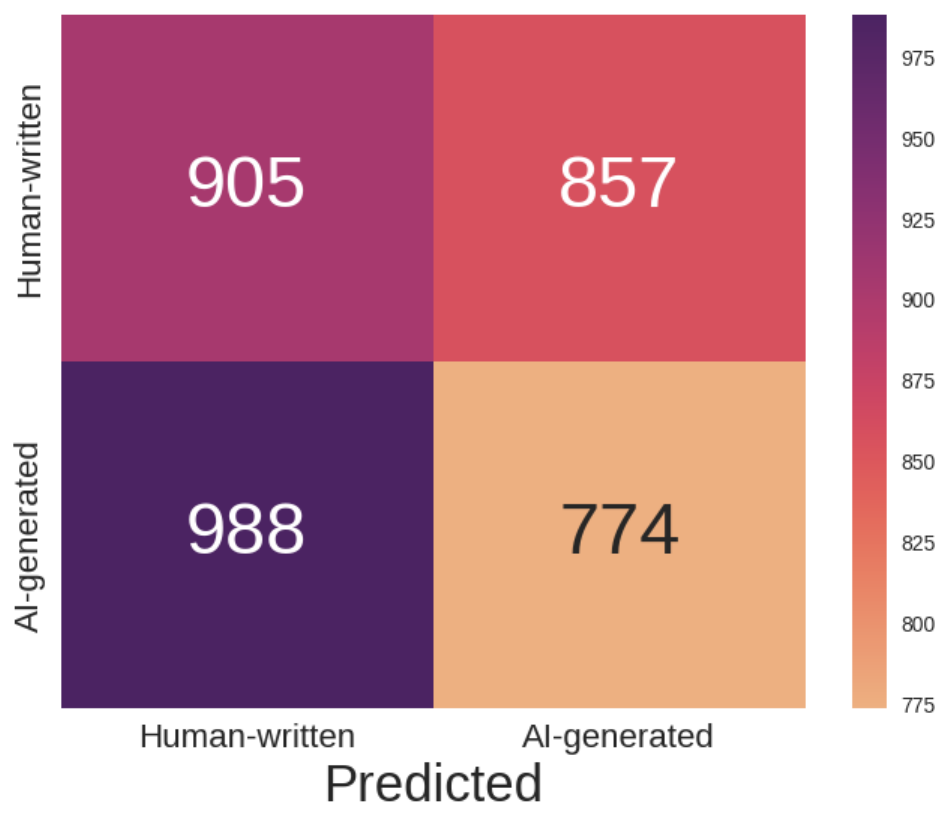}
        \caption{BBC BARD}
        \label{fig:perp_bard}
    \end{subfigure}
    \hfill
    \begin{subfigure}[b]{0.19\textwidth}
        \centering
        \includegraphics[width=\textwidth]{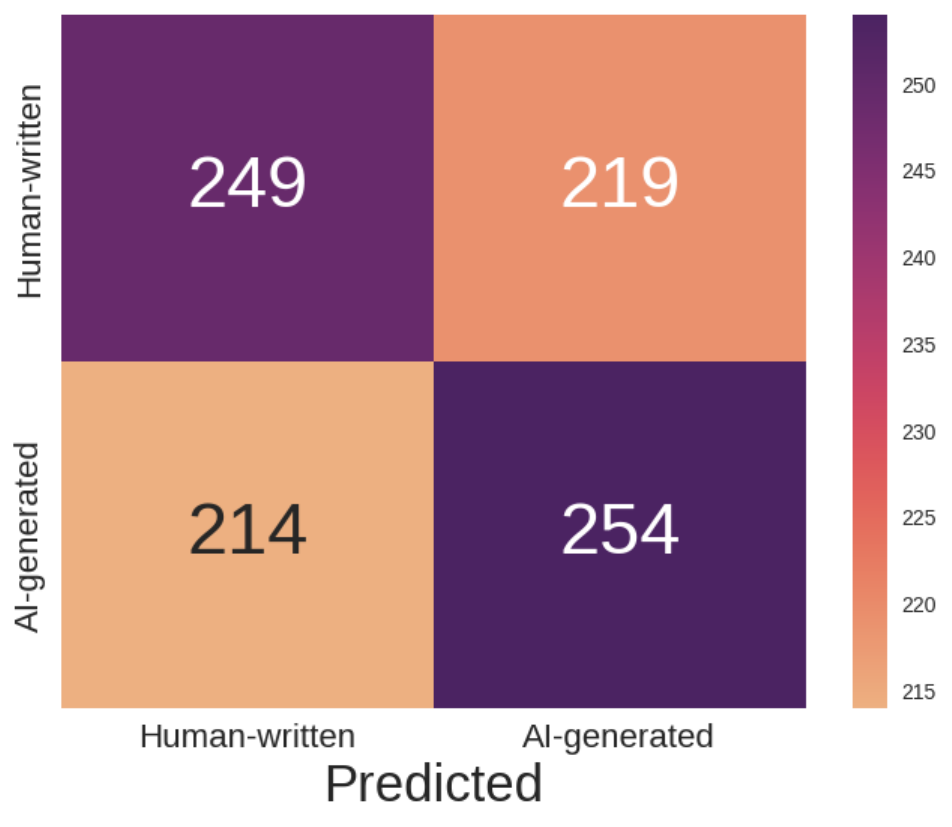}
        \caption{BBC Gemma-2B}
        \label{fig:perp_gemma2b}
    \end{subfigure}
    \hfill
    \begin{subfigure}[b]{0.19\textwidth}
        \centering
        \includegraphics[width=\textwidth]{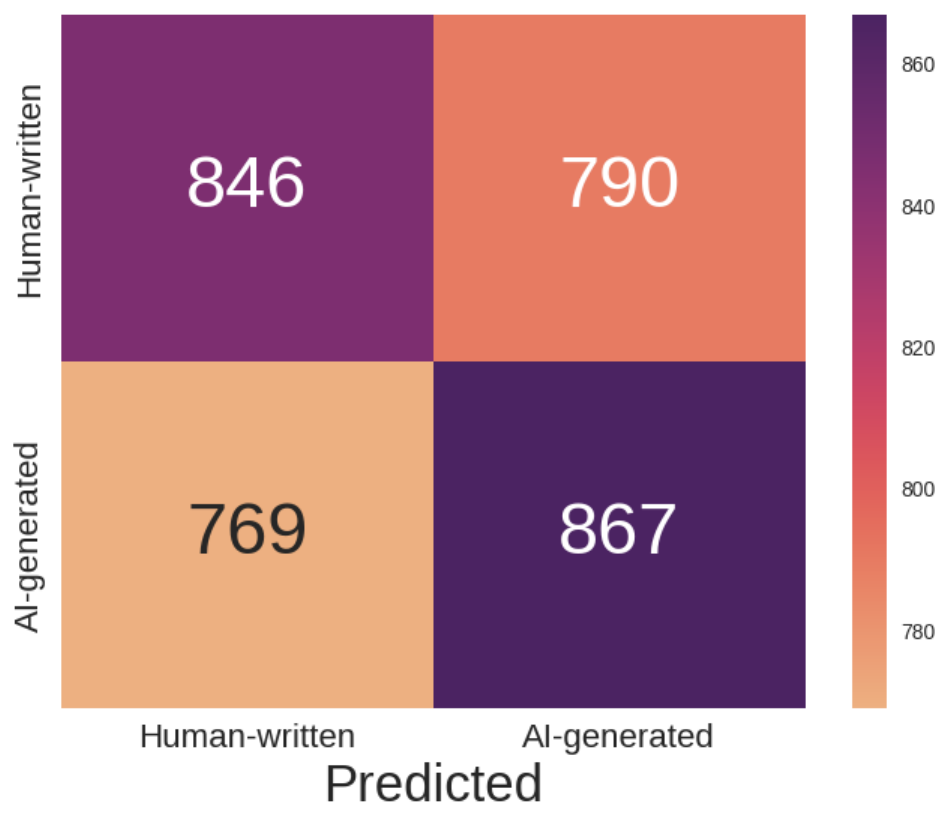}
        \caption{BBC Gemma-7B}
        \label{fig:perp_gemma7b}
    \end{subfigure}
    \\
    \begin{subfigure}[b]{0.19\textwidth}
        \centering
        \includegraphics[width=\textwidth]{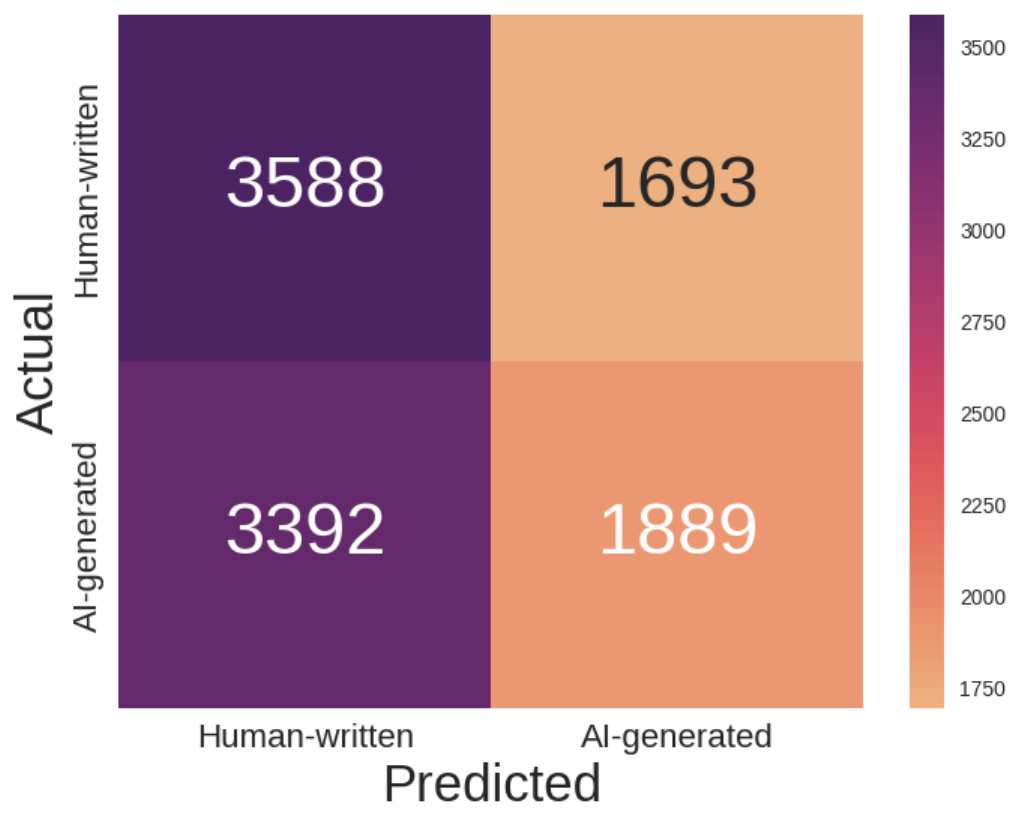}
        \caption{NDTV GPT-4}
        \label{fig:perp_model1}
    \end{subfigure}
    \hfill
    \begin{subfigure}[b]{0.19\textwidth}
        \centering
        \includegraphics[width=\textwidth]{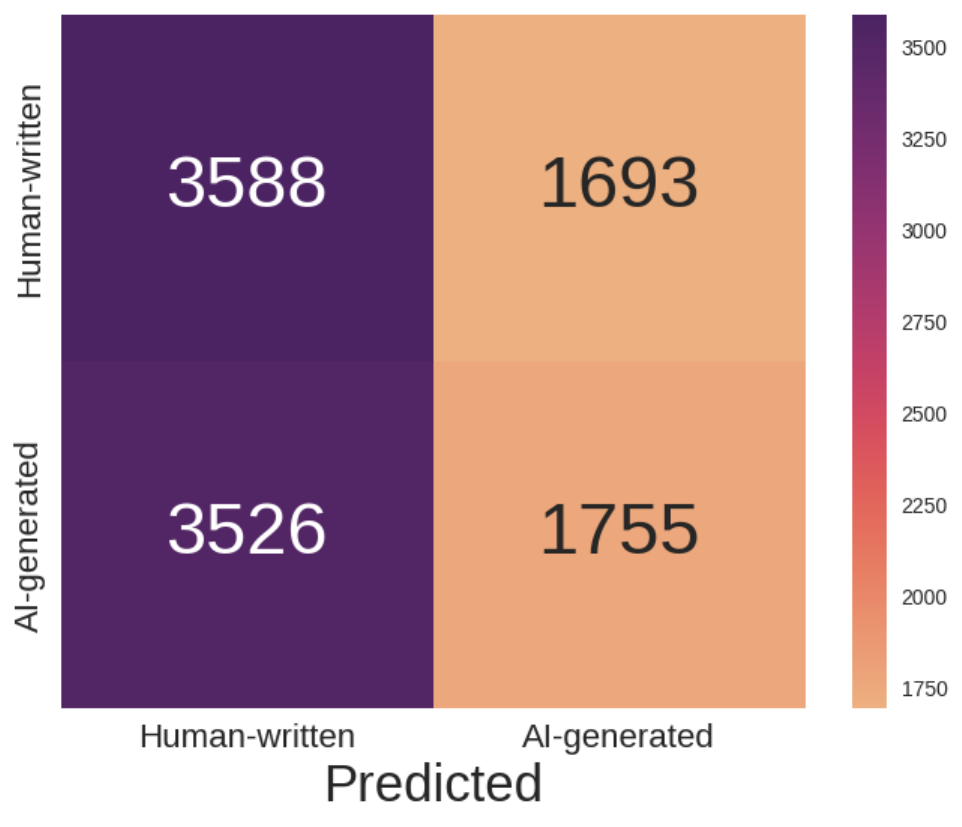}
        \caption{NDTV GPT-3.5}
        \label{fig:perp_model2}
    \end{subfigure}
    \hfill
    \begin{subfigure}[b]{0.19\textwidth}
        \centering
        \includegraphics[width=\textwidth]{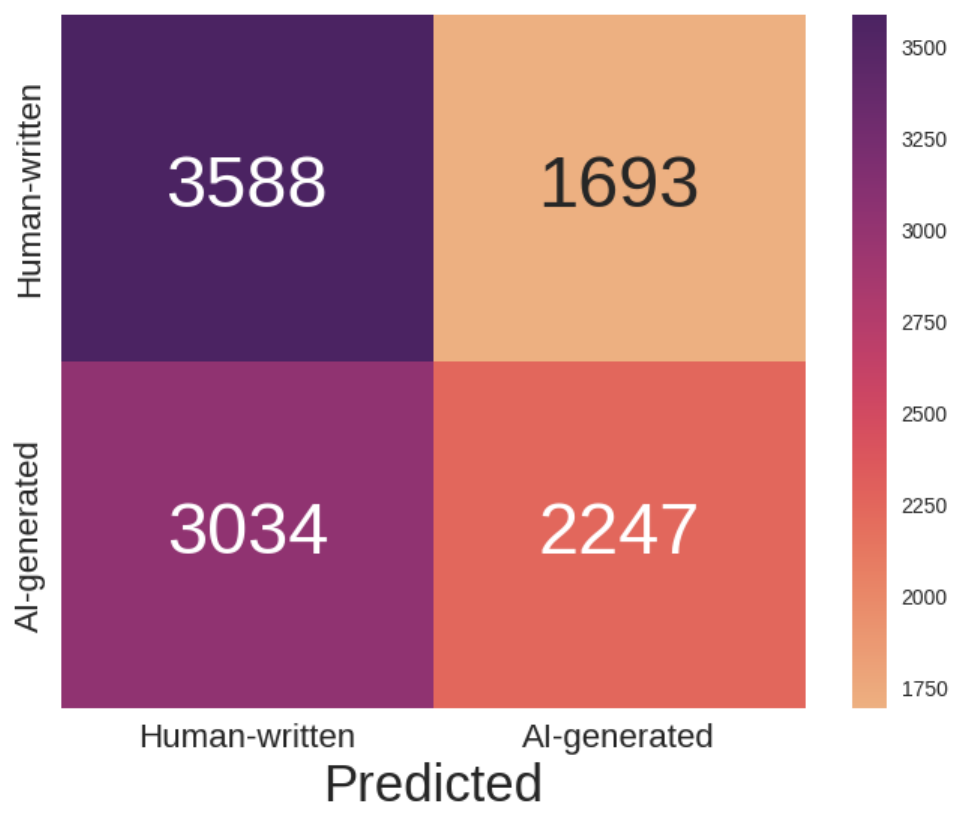}
        \caption{NDTV BARD}
        \label{fig:perp_model3}
    \end{subfigure}
    \hfill
    \begin{subfigure}[b]{0.20\textwidth}
        \centering
        \includegraphics[width=\textwidth]{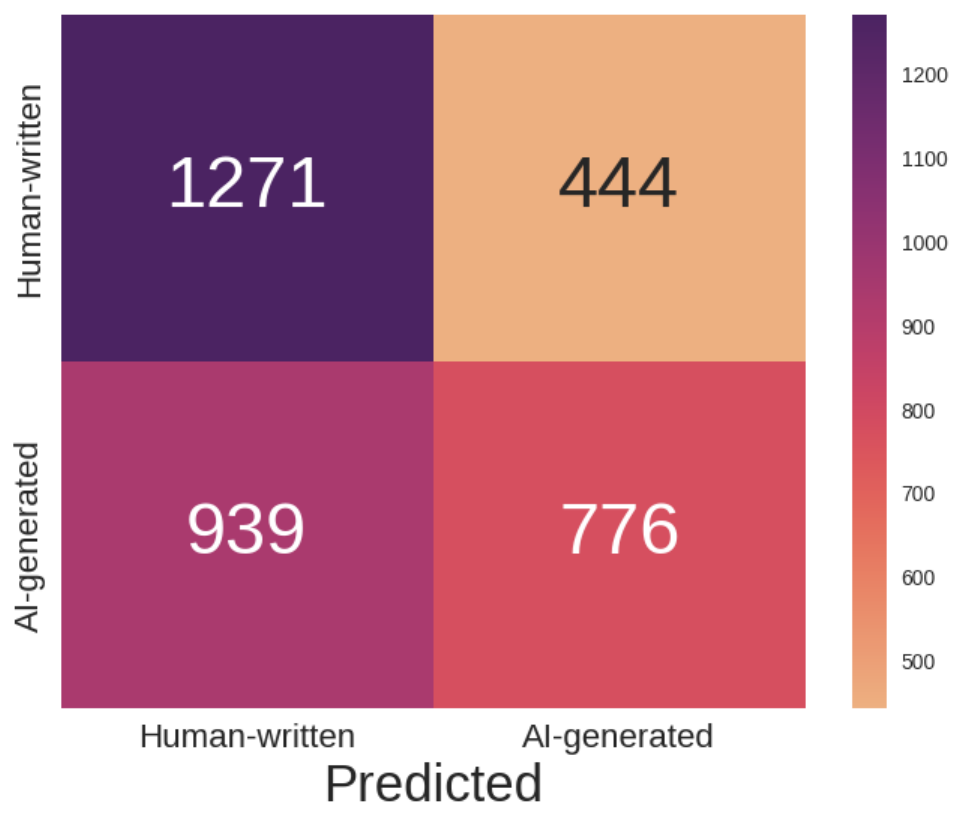}
        \caption{NDTV Gemma-2B}
        \label{fig:perp_model4}
    \end{subfigure}
    \hfill
    \begin{subfigure}[b]{0.20\textwidth}
        \centering
        \includegraphics[width=\textwidth]{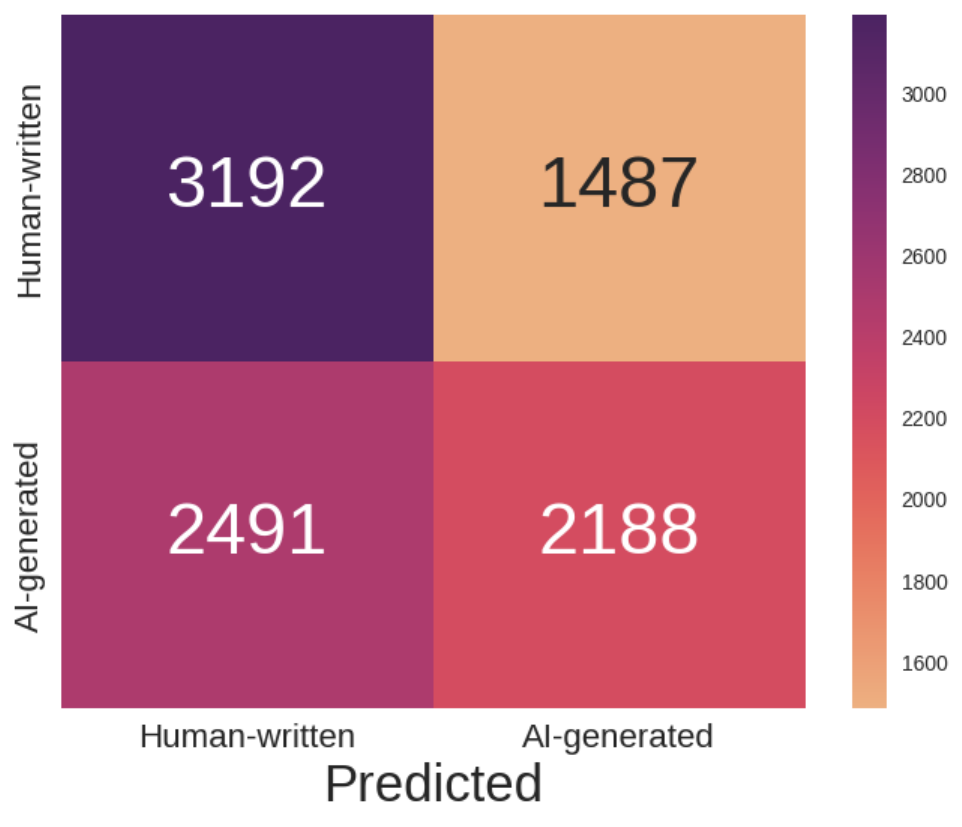}
        \caption{NDTV Gemma-7B}
        \label{fig:perp_model5}
    \end{subfigure}
    \caption{ConDA Confusion Matrices. The confusion matrices reveal various misclassifications of text. Notably, Gemma model responses are slightly more detectable compared to those generated by GPT models and BARD.}
    \label{fig:conda_cf}
\end{figure*}

The confusion matrices of all the methods can be found in Fig \ref{fig:raidar_cf}, Fig \ref{fig:radar_cf}, Fig \ref{fig:j_guard_cf} and Fig \ref{fig:conda_cf}.

\subsection{Results from Intrinsic Dimension Estimation} \label{sec:ID}

\begin{figure*}[!h]
    \small
    \centering
    \includegraphics[width=.9\linewidth]{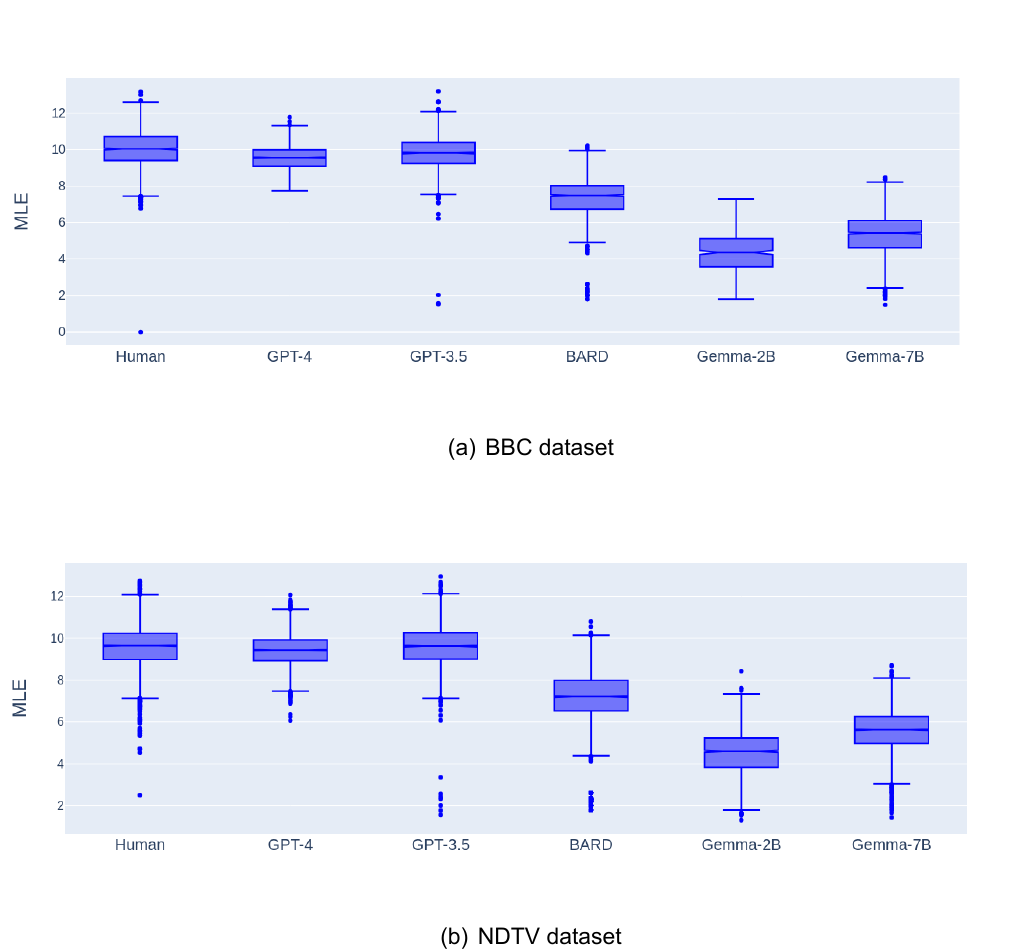}
    \caption{Maximum Likelihood estimation (MLE) of various models across datasets. MLE values of GPT models align closely with human MLE values, while BARD responses are slightly lower. In contrast, MLE values of Gemma models significantly differ from human values, facilitating easier identification of Gemma responses as AI-generated by MLE estimation.}
    \label{fig:IntrinsicDim-MLE}
\end{figure*}

\begin{figure*}[!h]
    \small
    \centering
    \includegraphics[width=.9\linewidth]{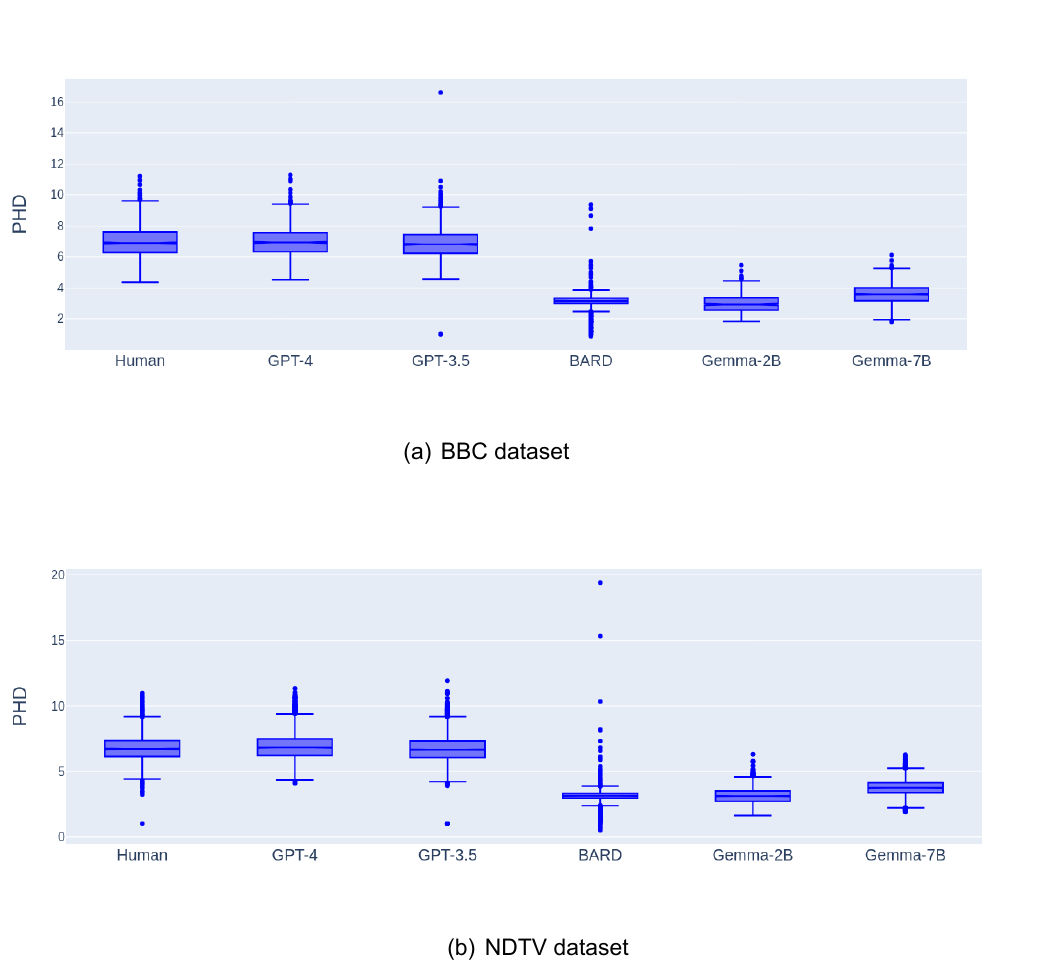}
    \caption{Persistence Homology Dimension (PHD) estimation  of various models across datasets. Two distinct clusters can be observed: one comprising GPT-4, GPT-3.5, and human-written PHD values, and the other comprising BARD, Gemma-2B, and Gemma-7B. The stark difference between these two groups, making the second group easily distinguishable from human text.}
    \label{fig:IntrinsicDim-PHD}
\end{figure*}

The results from intrinsic dimension estimation are presented as box plots in Fig \ref{fig:IntrinsicDim-MLE} and \ref{fig:IntrinsicDim-PHD}. We present the distribution of MLE and PHD estimations across datasets.

\subsection{Results from J-Guard} \label{sec:J-guard}

We present the cross-domain performance metrics like accuracy, precision, recall and F1 score for the J-Guard framework in Fig \ref{fig:jguard_acc}, Fig \ref{fig:jguard_prec}, Fig \ref{fig:jguard_recall} and Fig \ref{fig:jguard_f1} respectively. In this evaluation, the model undergoes training on a specific dataset and is subsequently tested on each distinct dataset. This method provides insights across various domains, exhibiting the model's ability to generalize to a dataset not encountered during the training phase.

\begin{table*}[h]
\centering
\resizebox{1\textwidth}{!}{
\begin{tabular}{ccccccccccccc}
\cline{4-13}
 &  &  & \multicolumn{10}{c}{\textbf{Testing Dataset}} \\ \cline{4-13} 
 &  &  & \multicolumn{5}{c}{\textbf{BBC Dataset}} & \multicolumn{5}{c}{\textbf{NDTV Dataset}} \\ \cline{4-13} 
 &  &  & GPT-4 & GPT-3.5 & BARD & Gemma-2B & Gemma-7B & GPT-4 & GPT-3.5 & BARD & Gemma-2B & Gemma-7B \\ \hline
\multirow{10}{*}{\textbf{\begin{tabular}[c]{@{}c@{}}Training \\ Dataset\end{tabular}}} & \multirow{5}{*}{\textbf{\begin{tabular}[c]{@{}c@{}}BBC \\ Dataset\end{tabular}}} & GPT-4 & 98.44 & 97.731 & 88.963 & 97.333 & 96.87 & 81.013 & 81.297 & 73.674 & 81.341 & 83.44 \\
 &  & GPT-3.5 & 99.149 & 99.291 & 98.44 & 96.8 & 97.023 & 70.403 & 79.64 & 76.752 & 84.475 & 87.42 \\
 &  & BARD & 99.574 & 98.156 & 99.007 & 98.667 & 98.168 & 81.013 & 80.919 & 79.072 & 85.714 & 86.699 \\
 &  & Gemma-2B & 97.73 & 96.312 & 97.73 & 99.467 & 99.237 & 83.807 & 84.706 & 90.009 & 88.12 & 91.587 \\
 &  & Gemma-7B & 96.879 & 95.603 & 98.014 & 99.733 & 99.237 & 83.617 & 85.369 & 93.466 & 88.557 & 88.916 \\ \cline{2-13} 
 & \multirow{5}{*}{\textbf{\begin{tabular}[c]{@{}c@{}}NDTV \\ Dataset\end{tabular}}} & GPT-4 & 99.574 & 99.433 & 99.574 & 97.6 & 98.55 & 99.242 & 98.438 & 99.006 & 91.327 & 90.598 \\
 &  & GPT-3.5 & 99.574 & 99.574 & 99.433 & 98.4 & 98.626 & 99.432 & 98.958 & 99.006 & 92.128 & 91.159 \\
 &  & BARD & 98.156 & 97.589 & 99.291 & 98.933 & 98.626 & 94.366 & 94.602 & 99.29 & 90.743 & 88.622 \\
 &  & Gemma-2B & 99.007 & 99.007 & 99.433 & 100 & 99.237 & 86.127 & 86.127 & 84.991 & 99.344 & 98.985 \\
 &  & Gemma-7B & 99.149 & 99.574 & 99.433 & 100 & 99.771 & 85.227 & 86.08 & 85.038 & 99.781 & 99.733 \\ \hline
\end{tabular}
}

\caption{ J-Guard Cross-model accuracy. J-Guard trained on GPT-3.5 data demonstrates high accuracies when testing on text generated by other models and slightly outperforms the J-Guard model trained on GPT-4 data.}
\label{fig:jguard_acc}
\end{table*}

\begin{table*}[h]
\centering
\resizebox{1\textwidth}{!}{
\begin{tabular}{ccccccccccccc}
\cline{4-13}
 &  &  & \multicolumn{10}{c}{\textbf{Testing Dataset}} \\ \cline{4-13} 
 &  &  & \multicolumn{5}{c}{\textbf{BBC Dataset}} & \multicolumn{5}{c}{\textbf{NDTV Dataset}} \\ \cline{4-13} 
 &  &  & GPT-4 & GPT-3.5 & BARD & Gemma-2B & Gemma-7B & GPT-4 & GPT-3.5 & BARD & Gemma-2B & Gemma-7B \\ \hline
 &  & GPT-4 & 99.718 & 99.696 & \cellcolor[HTML]{FFFFFF}100 & 100 & 98.594 & 74.557 & 75.099 & 69.726 & 76.847 & 78.413 \\
 &  & GPT-3.5 & 98.904 & 99.128 & 99.706 & 100 & 99.057 & 71.069 & 71.296 & 69.259 & 80.126 & 82.656 \\
 &  & BARD & 99.724 & 97.96 & 99.709 & 100 & 98.336 & 72.701 & 73.063 & 71.165 & 81.633 & 82.113 \\
 &  & Gemma-2B & 97.796 & 96.755 & 97.983 & 99.454 & 99.545 & 84.8 & 85.215 & 86.961 & 86.183 & 90.547 \\
 & \multirow{-5}{*}{\textbf{\begin{tabular}[c]{@{}c@{}}BBC \\ Dataset\end{tabular}}} & Gemma-7B & 97.23 & 96.988 & 97.994 & 100 & 99.246 & 88.616 & 88.949 & 92.293 & 90.347 & 94.548 \\ \cline{2-13} 
 &  & GPT-4 & 100 & 100 & 100 & 100 & 99.538 & 99.229 & 99.404 & 99.9 & 99.312 & 98.865 \\
 &  & GPT-3.5 & 100 & 100 & 100 & 100 & 99.692 & 99.613 & 99.606 & 99.8 & 98.829 & 98.759 \\
 &  & BARD & 100 & 99.695 & 100 & 100 & 99.087 & 99.462 & 99.566 & 99.505 & 97.32 & 97.552 \\
 &  & Gemma-2B & 98.901 & 99.123 & 99.712 & 100 & 99.545 & 78.37 & 78.96 & 77.283 & 98.996 & 98.745 \\
\multirow{-10}{*}{\textbf{\begin{tabular}[c]{@{}c@{}}Training \\ Dataset\end{tabular}}} & \multirow{-5}{*}{\textbf{\begin{tabular}[c]{@{}c@{}}NDTV \\ Dataset\end{tabular}}} & Gemma-7B & 98.904 & 99.133 & 99.145 & 100 & 100 & 76.969 & 78.318 & 76.677 & 99.568 & 99.529 \\ \hline
\end{tabular}
}

\caption{J-Guard Cross-model Precision. Models trained on the NDTV dataset and tested on the BBC dataset consistently demonstrate high precision, with many achieving 100\%. This indicates a low rate of misclassifying human text as AI-generated.}
\label{fig:jguard_prec}
\end{table*}

\begin{table*}[h]
\centering
\resizebox{1\textwidth}{!}{
\begin{tabular}{ccccccccccccc}
\cline{4-13}
 &  &  & \multicolumn{10}{c}{\textbf{Testing Dataset}} \\ \cline{4-13} 
 &  &  & \multicolumn{5}{c}{\textbf{BBC Dataset}} & \multicolumn{5}{c}{\textbf{NDTV Dataset}} \\ \cline{4-13} 
 &  &  & GPT-4 & GPT-3.5 & BARD & Gemma-2B & Gemma-7B & GPT-4 & GPT-3.5 & BARD & Gemma-2B & Gemma-7B \\ \hline
 &  & GPT-4 & 97.245 & 95.627 & \cellcolor[HTML]{FFFFFF}77.65 & 94.536 & 95.173 & 93.16 & 92.121 & 80.02 & 90.173 & 93.007 \\
 &  & GPT-3.5 & 99.449 & 99.417 & 97.135 & 93.443 & 95.023 & 97.977 & 97.374 & 92.913 & 92.052 & 95.216 \\
 &  & BARD & 99.449 & 98.251 & 98.281 & 97.268 & 98.039 & 98.266 & 96.304 & 94.98 & 92.486 & 94.374 \\
 &  & Gemma-2B & 97.796 & 95.627 & 97.421 & 99.454 & 98.944 & 81.696 & 82.977 & 93.209 & 91.04 & 93.165 \\
 & \multirow{-5}{*}{\textbf{\begin{tabular}[c]{@{}c@{}}BBC \\ Dataset\end{tabular}}} & Gemma-7B & 96.694 & 93.878 & 97.994 & 99.454 & 99.246 & 76.493 & 79.864 & 94.291 & 86.561 & 82.965 \\ \cline{2-13} 
 &  & GPT-4 & 99.174 & 98.834 & 99.14 & 95.082 & 97.587 & 99.229 & 97.374 & 98.032 & 83.382 & 82.44 \\
 &  & GPT-3.5 & 99.176 & 99.125 & 98.854 & 96.721 & 97.587 & 99.229 & 98.249 & 98.13 & 85.405 & 83.649 \\
 &  & BARD & 96.419 & 95.335 & 98.567 & 97.814 & 98.19 & 89.017 & 89.3 & 99.016 & 83.96 & 79.6 \\
 &  & Gemma-2B & 99.174 & 98.834 & 99.14 & 100 & 98.944 & 99.133 & 97.471 & 97.441 & 99.711 & 99.264 \\
\multirow{-10}{*}{\textbf{\begin{tabular}[c]{@{}c@{}}Training \\ Dataset\end{tabular}}} & \multirow{-5}{*}{\textbf{\begin{tabular}[c]{@{}c@{}}NDTV \\ Dataset\end{tabular}}} & Gemma-7B & 99.449 & 100 & 99.713 & 100 & 99.548 & 99.807 & 98.735 & 99.016 & 100 & 99.947 \\ \hline
\end{tabular}
}

\caption{J-Guard Cross-model Recall. We observe consistently high recall values across training and testing on both BBC and NDTV datasets, with only a few exceptions. This indicates that the model effectively minimizes missed detections of AI-generated text, irrespective of the dataset used. This robust performance suggests that the model's ability to accurately identify AI-generated text. }
\label{fig:jguard_recall}
\end{table*}

\begin{table*}[h]
\centering
\resizebox{1\textwidth}{!}{
\begin{tabular}{ccccccccccccc}
\cline{4-13}
 &  &  & \multicolumn{10}{c}{\textbf{Testing Dataset}} \\ \cline{4-13} 
 &  &  & \multicolumn{5}{c}{\textbf{BBC Dataset}} & \multicolumn{5}{c}{\textbf{NDTV Dataset}} \\ \cline{4-13} 
 &  &  & GPT-4 & GPT-3.5 & BARD & Gemma-2B & Gemma-7B & GPT-4 & GPT-3.5 & BARD & Gemma-2B & Gemma-7B \\ \hline
 &  & GPT-4 & 98.466 & 97.619 & \cellcolor[HTML]{FFFFFF}87.419 & 97.191 & 96.853 & 82.827 & 82.744 & 74.519 & 82.979 & 85.089 \\
 &  & GPT-3.5 & 99.176 & 99.272 & 98.403 & 96.61 & 96.998 & 82.382 & 82.319 & 79.361 & 85.676 & 88.493 \\
 &  & BARD & 99.586 & 98.108 & 98.99 & 98.615 & 98.187 & 83.572 & 83.089 & 81.366 & 86.721 & 87.818 \\
 &  & Gemma-2B & 97.796 & 96.188 & 97.701 & 99.454 & 99.244 & 83.219 & 84.081 & 89.976 & 88.545 & 91.837 \\
 & \multirow{-5}{*}{\textbf{\begin{tabular}[c]{@{}c@{}}BBC \\ Dataset\end{tabular}}} & Gemma-7B & 96.961 & 95.407 & 97.994 & 99.726 & 99.246 & 82.11 & 84.162 & 93.281 & 88.413 & 88.379 \\ \cline{2-13} 
 &  & GPT-4 & 99.585 & 99.413 & 99.568 & 97.479 & 98.553 & 99.299 & 98.378 & 98.957 & 90.652 & 89.908 \\
 &  & GPT-3.5 & 99.585 & 99.561 & 99.424 & 98.333 & 98.628 & 99.421 & 98.923 & 98.958 & 91.628 & 90.578 \\
 &  & BARD & 98.177 & 97.466 & 99.279 & 98.895 & 98.636 & 93.95 & 94.154 & 99.26 & 90.147 & 87.666 \\
 &  & Gemma-2B & 99.037 & 98.978 & 99.425 & 100 & 99.244 & 87.537 & 87.244 & 86.199 & 99.352 & 99.004 \\
\multirow{-10}{*}{\textbf{\begin{tabular}[c]{@{}c@{}}Training \\ Dataset\end{tabular}}} & \multirow{-5}{*}{\textbf{\begin{tabular}[c]{@{}c@{}}NDTV \\ Dataset\end{tabular}}} & Gemma-7B & 99.176 & 99.565 & 99.429 & 100 & 99.773 & 86.913 & 87.349 & 86.426 & 99.784 & 99.738 \\ \hline
\end{tabular}
}

\caption{J-Guard Cross-model F1 score. We observe that models trained on the BBC dataset and tested on the NDTV dataset exhibit lower performance compared to other combinations. Futhermore, models trained on Gemma responses struggle in detecting responses from GPT models and BARD.}
\label{fig:jguard_f1}
\end{table*}

\subsection{Results from RAIDAR}\label{sec:RAIDAR_appendix}
We experiment with the prompts used by \cite{mao2024raidar} for rewriting the text samples from the dataset. We employ Gemma-2B for this purpose. To rewrite the articles in Hindi itself, we modify the prompts accordingly. We observe that the prompts used significantly affect the output language of the rewritten text. We add an instruction along with the text while prompting it to the LLM. The following instructions were effective in generating rewritten articles in Hindi:
\begin{enumerate}[leftmargin=25pt]
    \item Concise this for me in Hindi only and keep all the information.
    \item Help me polish this in Hindi only.
    \item Make this fluent in Hindi only while making minimal changes.
    \item Refine the following paragraph for me in Hindi only.
    \item Revise this in Hindi only with your best efforts.
    \item Rewrite this in Hindi only.
\end{enumerate}

\section{AI Detectability Index for Hindi  ($ADI_{hi}$)}
\subsection{Probability Distribution Generation} \label{sec:distribution}
In this section, we outline the process to calculate the probability distributions essential to calculate the JSD. 
\\
\noindent For each word \( w_i \) in \( V \), we extract sentences \( S_h \) and \( S_{ai} \) from the human-written and AI-generated texts, respectively, containing \( w_i \). To evaluate the contextual similarity of these words, we create \( V_{\text{comb}} \) by taking the union of the word sets from \( S_h \) and \( S_{ai} \), which helps assess the surrounding context of \( w_i \) in both types of texts. We then generate co-occurrence vectors \( C_h \) and \( C_{ai} \), which record the frequency of words from \( V_{\text{comb}} \) in their respective sentences. These vectors are normalized to form probability distributions \( P_h \) and \( P_{ai} \), and we calculate the Jensen-Shannon Divergence (JSD) between the two distributions. The detailed algorithm for calculating $ADI_hi$ can be found in Algorithm \ref{alg:adi_calc}.
\begin{equation}
V_{\text{comb}} = V(S_h) \cup V(S_{ai})
\end{equation}
\begin{equation}
C_h(w) = \text{frequency of } w \text{ in } S_h \quad \forall w \in V_{\text{comb}}
\end{equation}
\begin{equation}
C_{ai}(w) = \text{frequency of } w \text{ in } S_{ai} \quad \forall w \in V_{\text{comb}}
\end{equation}
\begin{equation}
P_h(w) = \frac{C_h(w)}{\sum_{w' \in V_{\text{comb}}} C_h(w')} \quad \forall w \in V_{\text{comb}}
\end{equation}
\begin{equation}
P_{ai}(w) = \frac{C_{ai}(w)}{\sum_{w' \in V_{\text{comb}}} C_{ai}(w')} \quad \forall w \in V_{\text{comb}}
\end{equation}


\begin{algorithm}
\caption{$ADI_{hi}$ Calculation}
\label{alg:adi_calc}
\begin{algorithmic}[1]
\Require Human-written text $T_h$, AI-generated text $T_{ai}$
\Ensure ADI\(_{hi}\) spectrum
\State Extract vocabularies $V_h$ from $T_h$ and $V_{ai}$ from $T_{ai}$
\State Compute word intersection $V = V_h \cap V_{ai}$

    \State Extract sentences $S_h$ from $T_h$ and $S_{ai}$ from $T_{ai}$ containing $w_i$
    \State Compute $V_{comb} = V(S_h) \cup V(S_{ai})$
    
    \State Create co-occurrence vector $C_h$ for $S_h$ and $C_{ai}$ for $S_{ai}$ with respect to $V_{comb}$

    \State Normalize $C_h$ and $C_{ai}$ to create probability distributions $P_h$ and $P_{ai}$
    
    \State Compute the Jensen-Shannon Divergence: 
    \[
    \text{JSD}(P_h, P_{ai}) = \frac{1}{2} D_{\text{KL}}(P_h \| M) + \frac{1}{2} D_{\text{KL}}(P_{ai} \| M)
    \]
    where $M = \frac{1}{2}(P_h + P_{ai})$
\State Repeat steps 3-7 for each word $w_i$ in $V$ 

\State Compute average JSD across all words in $V$: 
\State Apply Yeo-Johnson power transformation to ensure normal distribution
\State Perform min-max normalization to scale values between 0 and 100
\State Output the ADI\(_{hi}\) spectrum

\end{algorithmic}
\end{algorithm}


\begin{figure*}[!h]
    \centering
    \small
    \includegraphics[width=1\textwidth]{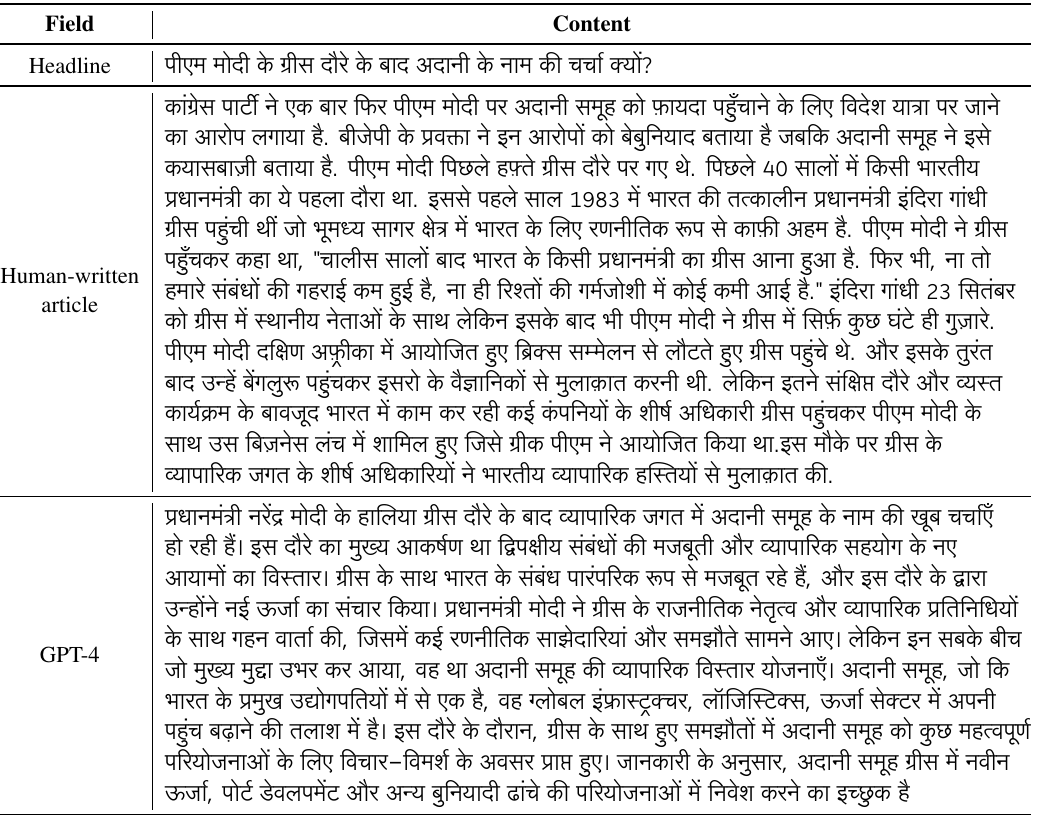}
    
    \caption{Samples of articles generated by accepted models (GPT-4, GPT-3.5, BARD, Gemma-2B and Gemma-7B).}
    \label{fig:Hindi_examples_1}
\end{figure*}

\end{document}